\def\year{2020}\relax
\documentclass[letterpaper]{article} 
\usepackage{aaai20}  
\usepackage{times}  
\usepackage{helvet} 
\usepackage{courier}  
\usepackage[hyphens]{url}  
\usepackage{graphicx} 
\usepackage{graphicx}  
\newcommand{\citet}[1]{\citeauthor{#1} \shortcite{#1}}
\newcommand{\citep}{\cite}

\usepackage[utf8]{inputenc} 
\usepackage[T1]{fontenc}    
\usepackage{booktabs}       
\usepackage{amsfonts}       
\usepackage{nicefrac}       
\usepackage{microtype}      

\usepackage{footnote}
\usepackage{tablefootnote}

\usepackage[para,online,flushleft]{threeparttable}

\usepackage{amsfonts,bm,color}
\usepackage{booktabs}       
\usepackage{amsopn}
\usepackage{amsmath}
\usepackage{amssymb}
\usepackage{empheq}
\usepackage{float}
\usepackage{framed}
\usepackage{multirow}

\usepackage{adjustbox}
\usepackage{algorithm,algorithmicx}
\usepackage{algpseudocode}
\usepackage[small]{caption}

\usepackage{subcaption}
\usepackage{todonotes}
\usepackage[normalem]{ulem}
\usepackage{multicol,lipsum}

\makeatletter
\newcommand{\leqnomode}{\tagsleft@true\let\veqno\@@leqno}
\newcommand{\reqnomode}{\tagsleft@false\let\veqno\@@eqno}
\makeatother

\newtheorem{theorem}{\bf{Theorem}}
\newtheorem{remark}{\bf{Remark}}

\newcommand{\argmin}{\operatornamewithlimits{arg\,min}}
\newcommand{\argmax}{\operatornamewithlimits{arg\,max}}

\DeclareMathOperator*{\minimize}{\text{minimize}}
\DeclareMathOperator*{\maximize}{\text{maximize}}
\DeclareMathOperator*{\st}{\text{subject to}}
\DeclareMathAlphabet\mathbfcal{OMS}{cmsy}{b}{n}
\newcommand{\Def}[0]{\mathrel{\mathop:}=}
\newcommand{\bz}{\mathbf z}
\newcommand{\bu}{\mathbf u}
\newcommand{\btheta}{\boldsymbol{\theta}}
\newcommand{\tbtheta}{\widetilde{\boldsymbol{\theta}}}
\newcommand{\bdelta}{\boldsymbol{\delta}}
\newcommand{\blambda}{\boldsymbol{\lambda}}
\newcommand{\iter}[2]{#1{}^{(#2)}}

\newcommand{\sijia}[1]{\todo[color=yellow]{Sijia: #1}}
\definecolor{Sijia_color}{rgb}{0., 0., 0.}
\definecolor{Horst_color}{rgb}{0., 0., 0.}
\definecolor{PRC}{rgb}{0., 0., 0.}
\definecolor{DV}{rgb}{0., 0., 0.}
\newcommand{\prnote}{\textcolor{PRC}}
\newcommand{\dvnote}{\textcolor{DV}}

\definecolor{aquamarine}{rgb}{0.5, 1.0, 0.83}
\definecolor{orange-red}{rgb}{1.0, 0.27, 0.0}
\definecolor{gray}{rgb}{0.5, 0.5, 0.5}
\definecolor{blue}{rgb}{0.0, 0.0, 1.0}
\definecolor{fuchsia}{rgb}{1.0, 0.0, 1.0}
\definecolor{black}{rgb}{0.0, 0.0, 0.0}
\definecolor{coral}{rgb}{1.0, 0.25, 0.25}

\title{An ADMM Based Framework for AutoML Pipeline Configuration}
\author{
{\Large{Sijia Liu,\textsuperscript{*}
Parikshit Ram,\textsuperscript{*}
Deepak Vijaykeerthy, 
Djallel Bouneffouf, 
Gregory Bramble,}}\\
\textbf{\Large{Horst Samulowitz,
Dakuo Wang,
Andrew Conn, 
Alexander Gray}}\\
\Large{IBM Research AI}\\
\textsuperscript{*} Equal contributions
}
\begin{document}
\maketitle
\begin{abstract}
We study the AutoML problem of automatically configuring machine learning pipelines by jointly selecting algorithms and their appropriate hyper-parameters for all steps in supervised learning pipelines. This \textit{black-box} (gradient-free) optimization with \textit{mixed} integer \& continuous variables is a   challenging problem. We propose a novel
AutoML scheme by leveraging the alternating direction method of multipliers (ADMM). 
The proposed framework is able to (i) {decompose the optimization problem into easier sub-problems that have a reduced number of variables and circumvent the challenge of mixed variable categories,}
and (ii) incorporate black-box constraints alongside the black-box optimization objective. We empirically evaluate the flexibility (in utilizing existing AutoML techniques), effectiveness (against open source AutoML toolkits), and unique capability (of executing AutoML with practically motivated black-box constraints) of our proposed scheme \dvnote{on a collection of binary classification data sets from UCI ML \& OpenML repositories. We observe that on an average our framework provides significant gains in comparison to other AutoML frameworks (Auto-sklearn \& TPOT)}, highlighting the practical advantages of this framework.
\end{abstract}
\section{Introduction} \label{sec:intro}
%
%
%
%
%
%
%
Automated machine learning (AutoML) research has received increasing attention. The focus has shifted from hyper-parameter optimization (HPO)  for the best configuration of a {\em single} machine learning (ML) algorithm  \citep{snoek2012practical},
to configuring multiple stages of a ML pipeline (e.g., transformations, feature selection, predictive modeling) \citep{feurer2015efficient}. 
%
Among the wide-range of research challenges offered by AutoML,
we focus on the automatic pipeline configuration problem (that is, joint algorithm selection and HPO), and tackle it from the perspective of mixed continuous-integer black-box nonlinear programming. This problem has two main challenges: (i) the tight coupling between the ML algorithm selection \& HPO; and (ii) the black-box nature of optimization objective lacking any explicit functional form and gradients -- optimization feedback is only available in the form of function evaluations.
We propose a new AutoML framework to address these challenges by leveraging the alternating direction method of multipliers (ADMM). ADMM offers a  {\em two-block} alternating optimization procedure that splits an involved problem (with multiple variables \& constraints) into simpler sub-problems
\citep{boyd2011distributed}

\noindent
\textbf{Contributions.}
Starting with a combinatorially large set of algorithm candidates and their collective set of hyper-parameters, we utilize ADMM to decompose the AutoML problem into three problems: (i) HPO with a  {\em small} set of {\em only continuous} variables \& constraints, (ii) a closed-form Euclidean projection onto an integer set, and (iii) a combinatorial problem of algorithm selection. 
Moreover, we exploit the ADMM framework to handle {\em any black-box constraints alongside the black-box objective (loss) function} -- the above decomposition seamlessly incorporates such constraints while retaining almost the same sub-problems.

Our contributions are:
(i) We explicitly model the coupling between hyper-parameters and available algorithms, and exploit the hidden structure in the AutoML problem (Section 3).
(ii) We employ ADMM to decompose the problem into a sequence of sub-problems (Section 4.1), which decouple the difficulties in AutoML and can each be solved more efficiently and effectively,
\prnote{
demonstrating over $10\times$ speedup and $10\%$ improvement in many cases.%
}
(iii) We present the first AutoML framework that explicitly handles general black-box constraints (Section 4.2).
(iv) We demonstrate the flexibility and effectiveness of the ADMM-based scheme empirically against popular AutoML toolkits Auto-sklearn \citep{feurer2015efficient} \& TPOT \citep{olson2016tpot} (Section 5),
\prnote{
performing best on $50\%$ of the datasets; Auto-sklearn performed best on $27\%$ and TPOT on $20\%$.%
}
%
%

%
%
%
%
\section{Related work} \label{sec:litreview}
%
%
%
%
%
%
%
%
%
%

\textbf{Black-box optimization in AutoML.}
Beyond grid-search for HPO, random search is a very competitive baseline because of its simplicity and parallelizability \citep{bergstra2012random}. Sequential model-based optimization (SMBO) \citep{smac} is a common technique with different `models' such as Gaussian processes \citep{snoek2012practical}, random forests \citep{smac} and tree-parzen estimators \citep{bergstra2011algorithms}. However, black-box optimization is a time consuming process because the expensive black-box function evaluation involves model training and scoring (on a held-out set). Efficient {\em multi-fidelity} approximations of the black-box function based on some budget (training samples/epochs) combined with bandit learning can skip unpromising candidates early via successive halving \citep{jamieson2016non,sabharwal2016selecting} and  HyperBand \citep{li2018hyperband}. However, these schemes essentially perform an efficient random search and are well suited for search over discrete spaces or discretized continuous spaces. BOHB \citep{bohb} combines SMBO (with TPE) and HyperBand for improved optimization. Meta-learning \citep{vanschoren2018meta} leverages past experiences in the optimization with search space refinements and promising starting points.
{The collaborative filtering based methods \citep{yang2019oboe} are examples of meta-learning, where information from past evaluation on other datasets is utilized to pick pipelines for any new datasets.
Compared to the recent works on iterative pipeline construction using tree search \citep{mohr2018ml,rakotoarison2019automated}, we provide a natural yet formal primal-dual decomposition of autoML pipeline configuration problems.   
}


%

\noindent
\textbf{Toolkits.}
Auto-WEKA~\citep{autoweka1,autoweka2} and Auto-sklearn \citep{feurer2015efficient} are the main representatives of SBMO-based AutoML. Both apply the general purpose framework SMAC \dvnote{(Sequential Model-based Algorithm Configuration)}~\citep{smac} to find optimal ML pipelines. Both consider a fixed shape of the pipeline with functional modules (preprocessing, transforming, modeling) and automatically select a ML algorithm and its hyper-parameters for each module. Auto-sklearn improves upon Auto-WEKA with two innovations: 
(i) a meta-learning based preprocessing step that uses `meta-features' of the dataset to determine good initial pipeline candidates based on past experience to {\em warm start} the optimization, 
(ii) an greedy forward-selection ensembling \citep{caruana2004ensemble} of the pipeline configurations found during the optimization as an independent post-processing step. Hyperopt-sklearn \citep{komer2014hyperopt} utilizes TPE as the SMBO. TPOT \citep{olson2016tpot} and ML-Plan \citep{mohr2018ml} use genetic algorithm and hierarchical task networks planning respectively to optimize over the pipeline shape and the algorithm choices, but require discretization of the hyper-parameter space (\dvnote{which can be inefficient in practice as it leads performance degradation}).
{AlphaD3M \citep{drori2018alphad3m} integrates reinforcement learning with Monte-Carlo tree search (MCTS) for solving AutoML problems but without imposing efficient decomposition over hyperparameters and model selection. 
    AutoStacker \citep{chen2018autostacker} focuses on ensembling and cascading to generate complex pipelines and the actual algorithm selection and hyper-parameter optimization happens via random search.
}

\section{An Optimization Perspective to AutoML}\label{sec:prob-setting}
%
%
We focus on the \textit{joint} algorithm selection and HPO for a {\em fixed pipeline} -- a ML pipeline with a fixed sequence of functional modules (preprocessing $\to$ missing/categorical handling $\to$ transformations $\to$ feature selection $\to$ modeling) with a set of algorithm choices in each module -- termed asthe CASH ({\bf c}ombined {\bf a}lgorithm {\bf s}election and {\bf H}PO) problem \citep{autoweka1,feurer2015efficient} and solved with toolkits such as Auto-WEKA and Auto-sklearn. We extend this formulation by explicitly expressing the combinatorial nature of the algorithm selection with {Boolean} variables and constraints. We will also briefly discuss how this formulation facilities
extension to other flexible pipelines.

\textbf{{Problem statement.}}
For $N$ \textit{functional modules} 
(e.g.,  preprocessor,  transformer, estimator)
with a choice of $K_i$ \textit{algorithms} in each, 
let $\bz_i \in \{0, 1\}^{K_i}$ denote the algorithm choice in module $i$, with the constraint $\mathbf{1}^\top \bz_i = \sum_{j=1}^{K_i} z_{ij} = 1$ ensuring that only a single algorithm is chosen from each module. Let $\bz = \left\{ \bz_1, \ldots, \bz_N \right\}$. 
Assuming that categorical {hyper-parameters} can be encoded as integers (using standard techniques), 
%
%
let $\btheta_{ij}$ be the \textit{hyper-parameters} of algorithm $j$ in module $i$, with $\btheta_{ij}^c \in \mathcal{C}_{ij} \subset \mathbb{R}^{m_{ij}^c}$ as the \textit{continuous} hyper-parameters (constrained to the set $\mathcal{C}_{ij}$) and $\btheta_{ij}^d \in \mathcal{D}_{ij} \subset \mathbb{Z}^{m_{ij}^d} $ as the \textit{integer} hyper-parameters (constrained to $\mathcal{D}_{ij}$). 
{Conditional hyper-parameters} can be handled with additional constraints $\btheta_{ij} \in \mathcal{E}_{ij}$ or by ``flattening'' the hyper-parameter tree and considering each leaf as a different algorithm. For simplicity of exposition, we assume that the conditional hyper-parameters are flattened into additional algorithm choices. 
Let $\btheta = \left\{ \btheta_{ij}, \forall i \in [N], j \in [K_i] \right\}$, where 
 $[n] = \{1, \ldots, n \}$ for  $n \in \mathbb{N}$. Let $f\left(\bz, \btheta; \mathcal{A} \right)$ represent some notion of loss of a ML pipeline corresponding to the algorithm choices as per $\bz$ with the hyper-parameters $\btheta$ on a learning task with data $\mathcal{A}$ (such as the $k$-fold cross-validation or holdout validation loss). The optimization problem of automatic pipeline configuration is stated as:
{\small\begin{equation}\label{eq: prob0}
\begin{array}{l}
   \displaystyle
   \min_{\bz, \btheta} f(\bz,  \btheta; \mathcal{A})   \\
     \st
\left\{
\begin{array}{l}
\bz_i \in \{0,1 \}^{K_i}, \mathbf 1^{\top} \bz_i = 1, \forall i \in [N], \\
\btheta_{ij}^c \in \mathcal{C}_{ij}, \btheta_{ij}^d \in \mathcal{D}_{ij},
\forall i \in [ N ], j \in [ K_i ].
\end{array}
\right.
\end{array}
\end{equation}}%

We highlight 2 key differences of problem \eqref{eq: prob0} from the conventional CASH formulation: (i) we use explicit Boolean variables $\bz$ to encode the algorithm selection, (ii) {we differentiate continuous variables/constraints from discrete ones for a possible efficient decomposition between  continuous optimization and integer programming.
%
%
}
These features better characterize the properties of the problem and thus enable more effective joint optimization. For any given $(\bz, \btheta)$ and data $\mathcal{A}$, the objective (loss) function $f(\bz, \btheta; \mathcal{A})$ is a black-box function -- it does not have an analytic form with respect to $(\bz, \btheta)$ (hence no derivatives). The actual evaluation of $f$ usually involves training, testing and scoring the ML pipeline corresponding to $(\bz, \btheta)$ on some split of the data $\mathcal{A}$.
%

\textbf{AutoML with black-box constraints.}
With the increasing adoption of AutoML, the   formulation \eqref{eq: prob0} {may not be} sufficient. AutoML may need to find ML pipelines with high predictive performance (low loss) that also {\em explicitly satisfy} application specific {\em constraints}. Deployment constraints may require the pipeline to have prediction latency or size in memory below some threshold (latency $\leq 10 \mu\mbox{s}$, memory $\leq 100$MB). Business specific constraints may desire pipelines with low overall classification error and an explicit upper bound on the false positive rate -- in a loan default risk application, false positives leads to loan denials to eligible candidates, which may violate regulatory requirements. In the quest for {\em fair} AI, regulators may explicitly require the ML pipeline to be above some predefined fairness threshold \citep{friedler2019comparative}. Furthermore, many applications have very domain specific metric(s) with corresponding constraints -- custom metrics are common in Kaggle competitions. We incorporate such requirements by extending AutoML formulation \eqref{eq: prob0} to include $M$ {\em black-box constraints}: 
%
{\small\begin{equation}
    \label{eq: prob0-csts}
    g_i\left(\bz, \btheta; \mathcal{A} \right) \leq \epsilon_i, i \in [M].
 \end{equation}}%
These functions have no analytic form with respect to $(\bz, \btheta)$, in constrast to the analytic constraints in problem \eqref{eq: prob0}. 
One approach is to incorporate these constraints into the black-box objective with a penalty function $p$, where the new objective becomes $f + \sum_i p(g_i, \epsilon_i)$ or $f \cdot \prod_i p(g_i, \epsilon_i)$. However, these schemes are very sensitive to the choice of the penalty function and do not guarantee feasible solutions.
%

\textbf{Generalization for more flexible pipelines.}
We can extend the problem formulation \eqref{eq: prob0} to enable optimization over the ordering of the functional modules. For example, we can choose between `preprocessor $\to$ transformer $\to$ feature selector' OR `feature selector $\to$ preprocessor $\to$ transformer'. 
\prnote{
The ordering of $T \leq N$ modules can be optimized by introducing $T^2$ Boolean variables $\mathbf{o} = \{ o_{ik} \colon i,k \in [T] \}$, where $o_{ik} = 1$ indicates that module $i$ is placed at position $k$.
}
The following constraints are then needed: (i) $\sum_{k \in [T]} o_{ik} = 1,  \forall i \in [T]$ indicates that module $i$ is placed at a single position, and (ii) $\sum_{i \in [T]} o_{ik} = 1 \forall k \in [T]$ enforces that only one module is placed at position $k$. These variables
can be added to $\bz$ in problem \eqref{eq: prob0} ($\bz = \{ \bz_1, \ldots, \bz_N, \mathbf{o} \}$). \textcolor{Sijia_color}{The resulting formulation still obeys the generic form of \eqref{eq: prob0}, which as will be evident later, can be efficiently solved by an operator splitting framework like ADMM \citep{boyd2011distributed}.}

\section{ADMM-Based Joint Optimizer} \label{sec: ADMM}

%
\prnote{%
ADMM provides a general effective optimization framework to solve complex problems with mixed variables and multiple constraints \citep{boyd2011distributed,liu2018zeroth}. We utilize this framework to decompose  problem \eqref{eq: prob0} without and with black-box constraints \eqref{eq: prob0-csts} into easier sub-problems.%
}

\subsection{4.1~ Efficient operator splitting  for AutoML
}
In what follows, we focus on  solving problem \eqref{eq: prob0} with  analytic constraints. The handling of black-box constraints will be elaborated on in the next section. Denoting $\btheta^c = \{ \btheta_{ij}^c, \forall i \in [N], j \in [K_i] \}$ as all the continuous hyper-parameters and $\btheta^d$ (defined correspondingly) as all the integer hyper-parameters, we re-write problem \eqref{eq: prob0} as:
{\small
\begin{equation}\label{eq: prob0-rw}
\begin{array}{l}
     \displaystyle    \min_{\bz, \btheta = \{ \btheta^c, \btheta^d \} }
f\left(\bz,  \left\{\btheta^c, \btheta^d \right\}; \mathcal{A} \right)  \\
     \st
\left\{
\begin{array}{l}
\bz_i \in \{0,1 \}^{K_i}, \mathbf 1^{\top} \bz_i = 1, \forall i \in [N], \\
\btheta_{ij}^c \in \mathcal{C}_{ij}, \btheta_{ij}^d \in \mathcal{D}_{ij},
\forall i \in [ N ], j \in [ K_i ].
\end{array}
\right.
\end{array}
\end{equation}}%
%
%

\textbf{{Introduction of continuous surrogate loss.}}
With $\widetilde{\mathcal{D}}_{ij}$ as the continuous relaxation of the integer space $\mathcal{D}_{ij}$ (if $\mathcal{D}_{ij}$ includes integers ranging from $\{l, \ldots, u \}\subset \mathbb{Z}$, then $\widetilde{\mathcal{D}}_{ij} = [ l, u ] \subset \mathbb{R}$), and $\tbtheta^d$ as the continuous surrogates for $\btheta^d$ with $\tbtheta_{ij} \in \widetilde{\mathcal{D}}_{ij}$ (corresponding to $\btheta_{ij} \in \mathcal{D}_{ij}$), we utilize a surrogate loss function $\widetilde{f}$ for problem \eqref{eq: prob0-rw} defined solely over the continuous domain with respect to $\btheta$:
{\small
\begin{equation}
    \label{eq:bbloss-continuous-surrogate}
    \widetilde{f} \left(\bz,  \left\{\btheta^c, \tbtheta^d \right\}; \mathcal{A} \right) \Def f\left(\bz,  \left\{\btheta^c, \mathcal{P}_{\mathcal{D}} \left( \tbtheta^d \right) \right\}; \mathcal{A} \right),
\end{equation}}%
where
$\mathcal{P}_{\mathcal{D}} ( \tbtheta^d ) = \{ \mathcal{P}_{\mathcal{D}_{ij}} ( \tbtheta_{ij}^d ), \forall i \in [N], j \in [K_i] \}$ is the projection of the continuous surrogates onto the integer set. 
This projection is {\bf necessary} since the black-box function is defined (hence {\em can only be evaluated}) on the integer sets $\mathcal{D}_{ij}$s. 
%
%
\prnote{%
Ergo, problem \eqref{eq: prob0-rw} can be {\em equivalently} posed as%
}
{\small 
\begin{equation}\label{eq: prob0-rr}
\begin{array}{l}
     \displaystyle \min_{\bz, \btheta^c, \tbtheta^d, \bdelta}
\widetilde{f} \left(\bz,  \left\{\btheta^c, \tbtheta^d \right\}; \mathcal{A} \right)   \\
     \st
\left\{
\begin{array}{l}
\bz_i \in \{0,1 \}^{K_i}, \mathbf 1^{\top} \bz_i = 1, \forall i \in [N] \\
\btheta_{ij}^c \in \mathcal{C}_{ij}, \tbtheta_{ij}^d \in \widetilde{\mathcal{D}}_{ij}, \forall i \in [ N ], j \in [ K_i ] \\
\bdelta_{ij} \in \mathcal{D}_{ij}, \forall i \in [ N ], j \in [ K_i ] \\
\tbtheta_{ij}^d  = \bdelta_{ij}, \forall i \in [ N ], j \in [ K_i ],
\end{array}
\right. 
\end{array}
\end{equation}}%
where the {equivalence between problems \eqref{eq: prob0-rw} \& \eqref{eq: prob0-rr}} is established by the equality constraint $\tbtheta_{ij}^d = \bdelta_{ij} \in \mathcal{D}_{ij}$, implying $\mathcal{P}_{\mathcal{D}_{ij}} ( \tbtheta_{ij}^d ) = \tbtheta_{ij}^d \in \mathcal{D}_{ij} $ 
and $\widetilde{f} (\bz,  \{\btheta^c, \tbtheta^d \}; \mathcal{A} ) = f(\bz,  \{\btheta^c, \tbtheta^d \}; \mathcal{A} )$.
%
The continuous surrogate loss \eqref{eq:bbloss-continuous-surrogate} is key in being able to perform theoretically grounded operator splitting (via ADMM) over mixed continuous/integer variables in the AutoML problem \eqref{eq: prob0-rw}.

\textbf{{Operator splitting from ADMM.}}
Using the notation that $I_\mathcal{X}(\mathbf{x}) = 0$ if $\mathbf{x} \in \mathcal{X}$ else  $+\infty$, and defining the sets 
$\mathcal{Z}  = \{ \bz \colon \bz = \{\bz_i \}, \bz_i \in \{0,1 \}^{K_i}, \mathbf 1^{\top} \bz_i = 1, \forall i \in [N]
\}$,
$ \mathcal{C}  = \{\btheta^c \colon \btheta^c = \{ \btheta_{ij}^c\},  \btheta_{ij}^c \in \mathcal{C}_{ij}, \forall i \in [N], j \in [K_i]  \}$, 
$ \mathcal{D}  = \{\bdelta \colon \bdelta = \{ \bdelta_{ij} \},  \bdelta_{ij} \in \mathcal{D}_{ij}, \forall i \in [N], j \in [K_i]  \}$ and 
$ \widetilde{\mathcal{D}} = \{\tbtheta^d \colon \tbtheta^d = \{ \tbtheta_{ij}^d \}, \tbtheta_{ij}^d \in \widetilde{\mathcal{D}}_{ij}, \forall i \in [N], j \in [K_i]  \}$,
we can re-write problem \eqref{eq: prob0-rr} as
{\small
\begin{align}
    \label{eq:prob1}
   \displaystyle   \min_{\bz, \btheta^c, \tbtheta^d, \bdelta}~
  &  \widetilde{f} \left(\bz,  \left\{\btheta^c, \tbtheta^d \right\}; \mathcal{A} \right)
    + I_\mathcal{Z}(\bz) + I_\mathcal{C}(\btheta^c)
    + I_{\widetilde{\mathcal{D}}}(\tbtheta^d) \nonumber  \\
    & + I_\mathcal{D}(\bdelta); ~
    \st\ \  \tbtheta^d = \bdelta.
\end{align}}%
with the corresponding augmented Lagrangian function 
{\small \begin{align}
    \label{eq:prob1-al}
  & \mathcal{L}( \bz, \btheta^c, \tbtheta^d, \bdelta, \blambda ) \Def  \widetilde{f} \left(\bz,  \left\{\btheta^c, \tbtheta^d \right\}; \mathcal{A} \right)
    + I_\mathcal{Z}(\bz) + I_\mathcal{C}(\btheta^c) \nonumber \\
  &  \quad    + I_{\widetilde{\mathcal{D}}}(\tbtheta^d) 
  + I_\mathcal{D}(\bdelta) 
    + \blambda^\top \left( \tbtheta^d - \bdelta \right) 
    + \frac{\rho}{2} \left\| \tbtheta^d - \bdelta \right\|_2^2,
\end{align}}%
where $\blambda$ is the Lagrangian multiplier, and $\rho > 0$ is a penalty parameter for the augmented term. 

ADMM \citep{boyd2011distributed} 
alternatively minimizes the augmented Lagrangian function \eqref{eq:prob1-al} over \textit{two} blocks of variables, leading to an efficient operator splitting framework for nonlinear programs with \textit{nonsmooth} objective function and \textit{equality} constraints. 
%
Specifically, 
ADMM solves problem \eqref{eq: prob0} by alternatively minimizing \eqref{eq:prob1-al} over  variables  $\{{\btheta^c}, {\tbtheta^d} \} $,  and $\{  \boldsymbol{\delta}, \mathbf z \} $. This can be equivalently converted into   3 sub-problems over variables $\{{\btheta^c}, {\tbtheta^d} \} $, $\boldsymbol{\delta}$ and $\mathbf z$, respectively. We refer readers to Algorithm\,1 for simplified sub-problems and Appendix \ref{app:thr_admm}\footnote{Appendices are  at \url{https://arxiv.org/pdf/1905.00424.pdf}} for detailed derivation.

The rationale behind the advantage of ADMM  is that it decomposes the AutoML problem into sub-problems with smaller number of variables: This is  crucial in black-box optimization where convergence is  strongly dependent on the number of variables. For example, the number of black-box evaluations needed for critical point convergence
is typically $O(n  \sim n^3)$ for $n$ variables \citep{larson2019derivative}. 
In what follows, we show that the easier sub-problems in Algorithm\,1 yield great interpretation of \prnote{the AutoML problem \eqref{eq: prob0}} and suggest efficient solvers  in terms of continuous hyper-parameter optimization, integer projection operation, and combinatorial algorithm selection.  

\begin{algorithm*}[htb]
\caption{\textbf{Operator splitting from ADMM to solve problem \eqref{eq: prob0-rr}}}
\label{alg:ADMM}
\begin{algorithmic} 
\State 
{\small\begin{align}
\label{eq: theta_min}
 \left\{ \iter{\btheta^c}{t+1}, \iter{\tbtheta^d}{t+1} \right\} 
   & = \argmin_{\btheta^c, \tbtheta^d} 
    \widetilde{f} \left(\iter{\bz}{t},  \left\{\btheta^c, \tbtheta^d \right\}; \mathcal{A} \right)
    + I_\mathcal{C}(\btheta^c)
    + I_{\widetilde{\mathcal{D}}}(\tbtheta^d) 
    + (\rho/2) \left\| \tbtheta^d - \mathbf{b} \right\|_2^2, \quad
  \mathbf{b} \Def  \iter{\bdelta}{t} - \frac{1}{\rho} \iter{\blambda}{t}, \tag{$\boldsymbol{\theta}$-min}\\
    \label{eq: delta_min}
    \iter{\bdelta}{t+1} 
    & = \argmin_{\bdelta} 
    I_\mathcal{D}(\bdelta) 
    + (\rho/2) \left\| \mathbf{a} - \bdelta \right\|_2^2, \quad  \mathbf{a} \Def \iter{\tbtheta^d}{t+1} + (1 / \rho) \iter{\blambda}{t}, \tag{$\boldsymbol{\delta}$-min}
    \\
    \label{eq: z_step}
    \iter{\bz}{t+1}
    & = \argmin_\bz
    \widetilde{f} \left(\bz,  \left\{\iter{\btheta^c}{t+1}, \iter{\tbtheta^d}{t+1} \right\}; \mathcal{A} \right)
    + I_\mathcal{Z}(\bz), \tag{$\mathbf z$-min}
\end{align}}%
where  $(t)$ represents the  iteration index, and
  the Lagrangian multipliers $\boldsymbol{\lambda}$ are updated as  $\iter{\blambda}{t+1}
     = \iter{\blambda}{t} + \rho ( \iter{\tbtheta^d}{t+1} - \iter{\bdelta}{t+1} )$.
\end{algorithmic}
\end{algorithm*}
\textbf{Solving \ref{eq: theta_min}.} 
Problem \eqref{eq: theta_min} can be rewritten as
{\small 
\begin{equation}
    \label{eq:cont-bb-2-rw}
    \begin{array}{l}
     \displaystyle      \min_{\btheta^c, \tbtheta^d} 
    \widetilde{f} \left(\iter{\bz}{t},  \left\{\btheta^c, \tbtheta^d \right\}; \mathcal{A} \right)
    + \frac{\rho}{2} \left\| \tbtheta^d - \mathbf{b} \right\|_2^2  \\
        \st 
    \left\{
    \begin{array}{l}
    \btheta_{ij}^c \in \mathcal{C}_{ij}\\
    \tbtheta_{ij}^d \in \widetilde{\mathcal{D}}_{ij} ,
    \end{array}
    \right.
    \forall i \in [N], j \in [K_i],
    \end{array}
\end{equation}}%
where both $\btheta^c$ and $\tbtheta^d$ are continuous optimization variables.
Since the algorithm selection scheme $\iter{\bz}{t}$ is fixed for this problem, $\widetilde{f}$ in problem \eqref{eq:cont-bb-2-rw} only depends on the hyper-parameters of the chosen algorithms -- the {\em active set} of continuous variables $(\btheta_{ij}^c, \tbtheta_{ij}^d)$ where $\iter{z_{ij}}{t} = 1$. This splits problem \eqref{eq:cont-bb-2-rw} even further into two problems. The \textit{inactive} set problem reduces to the following for all $i \in [N], j \in [K_i]$ such that $z_{ij} = 0$:
{\small\begin{equation}
    \label{eq:cbb-inactive}
    \min_{\tbtheta_{ij}^d}
    \frac{\rho}{2} \| \tbtheta_{ij}^d - \mathbf{b}_{ij} \|_2^2 
    \quad \st  \quad
    \tbtheta_{ij}^d \in \widetilde{\mathcal{D}}_{ij},
\end{equation}}%
which is solved by a Euclidean projection of $\mathbf{b}_{ij}$ onto $\widetilde{\mathcal{D}}_{ij}$. 
%

For the \textit{active} set of variables $S = \{ (\btheta_{ij}^c, \tbtheta_{ij}^d) \colon \btheta_{ij}^c \in \mathcal{C}_{ij}, \tbtheta_{ij} \in \widetilde{\mathcal{D}}_{ij}, z_{ij} = 1,  \forall i \in [N], j \in [K_i] \}$, problem \eqref{eq:cont-bb-2-rw} reduces to the following black-box optimization with only the {\em small} active set of {\em  continuous} variables\footnotemark
\footnotetext{%
\prnote{%
For the AutoML problems we consider in our empirical evalutations, $ | \btheta | = | \btheta_{ij}^c | + | \tbtheta_{ij}^d | \approx 100 $ while the largest possible active set $S$ is less than $15$ and typically less than $10$.%
}%
}
{\small
\begin{equation}
    \label{eq:cbb-active}
    \min_{(\btheta^c, \tbtheta^d) \in S} 
    \widetilde{f} \left(\iter{\bz}{t},  \left\{\btheta^c, \tbtheta^d \right\}; \mathcal{A} \right)
    + \frac{\rho}{2} \left\| \tbtheta^d - \mathbf{b} \right\|_2^2.
\end{equation}}%
The above problem can be solved using Bayesian optimization \citep{shahriari2016taking}, direct search \citep{larson2019derivative},
or   trust-region based derivative-free optimization \citep{conn2009introduction}.
%


\textbf{Solving \ref{eq: delta_min}.} 
According to the definition of $\mathcal D$, problem \eqref{eq: delta_min} can be rewritten as 
{\small\begin{equation}
    \label{eq:int-bb-2d-rw}
    \min_{  \bdelta  } \frac{\rho}{2} \| \bdelta -  {\mathbf a} \|_2^2
   \quad \st   \bdelta_{ij} \in \mathcal D_{ij}, \forall i \in [N], j \in [K_i],
\end{equation}}%
and solved in closed form by projecting $\mathbf{a}$ onto $\widetilde{\mathcal{D}}$ and then rounding to the nearest integer in $\mathcal{D}$.
%
%

\textbf{Solving \ref{eq: z_step}.} 
\label{sssec:algsel}
Problem \eqref{eq: z_step}  rewritten as
{\small\begin{equation}
    \label{eq:zbb-rw}
    \begin{array}{l}
\displaystyle      \min_\bz
    \widetilde{f} \left(\bz,  \left\{\iter{\btheta^c}{t+1}, \iter{\tbtheta^d}{t+1} \right\}; \mathcal{A} \right)    \\
         \st ~
    \bz_i \in \{0,1 \}^{K_i}, \mathbf 1^{\top} \bz_i = 1, \forall i \in [N]
    \end{array}
\end{equation}}%
is a black-box integer program solved exactly with $\prod_{i=1}^N K_i$ evaluations of $\widetilde{f}$. 
However, this is generally not feasible.
Beyond random sampling, there are a few ways to leverage existing AutoML schemes:
%
%
(i) \textit{Combinatorial multi-armed bandits.} -- Problem \eqref{eq:zbb-rw} can be interpreted through combinatorial bandits as the selection of the optimal $N$ arms (in this case, algorithms) from $\sum_{i=1}^N K_i$ arms based on bandit feedback and can be efficiently solved with Thompson sampling \citep{durand2014thompson} 
%
(ii) \textit{Multi-fidelity approximation of black-box evaluations} -- Techniques such as successive halving \citep{jamieson2016non,li2018hyperband} or incremental data allocation \citep{sabharwal2016selecting} can efficiently 
search over a discrete set of $\prod_{i=1}^N K_i$ candidates.
%
(iii) \textit{Genetic algorithms} -- Genetic programming can perform this discrete black-box optimization 
starting from a randomly generated population and building the next generation based on the `fitness' of the pipelines and random `mutations' and `crossovers'.
%

\begin{algorithm*}[!htb]
\caption{\textbf{Operator splitting from ADMM to solve problem \eqref{eq:bb-csts-rr} (with black-box constraints)}}
\label{alg:ADMM-csts}
\begin{algorithmic} 
\State 
{\small 
\begin{align}
    \left\{ \iter{\btheta^c}{t+1}, \iter{\tbtheta^d}{t+1},
    \iter{\bu}{t+1}\right\} 
    & =  \argmin_{\btheta^c, \tbtheta^d, \bu}
    \widetilde{f} 
    + \frac{\rho}{2} 
    \left\| \tbtheta^d - \mathbf{b} \right\|_2^2  
    + I_\mathcal{C}(\btheta^c)
    + I_{\widetilde{\mathcal{D}}}(\tbtheta^d)   + I_{\mathcal{U}}(\bu) 
    + \frac{\rho}{2} \sum_{i=1}^{M} \left[ 
    \widetilde{g}_i
    + u_i  -  \epsilon_i  + \frac{\iter{\mu_i}{t}}{\rho}
    \right]^2, \nonumber \\
    \iter{\bdelta}{t+1} &=  \argmin_{  \bdelta  } \quad \frac{\rho}{2} \| \bdelta -  {\mathbf a} \|_2^2 + I_{\mathcal{D}}(\bdelta), 
    \nonumber \\
    \iter{\bz}{t+1} & =  \argmin_\bz 
   \widetilde{f} 
    + I_{\mathcal{Z}}(\bz) 
   + \frac{\rho}{2}
    \sum_{i=1}^{M} \left[ 
    \widetilde{g}_i 
    -  \epsilon_i + \iter{u_i}{t+1}  + \frac{1}{\rho} \iter{\mu_i}{t}
    \right]^2,  \nonumber 
\end{align}}%
where  the arguments of $\tilde f$ and $\tilde g_i$ are omitted for brevity, 
$\mathbf a$  and $\mathbf{b}$ have been defined in Algorithm\,1, $\mathcal{U} = \{ \mathbf{u} \colon \mathbf{u} = \{ u_i \}$,
and   $\mu_i$ is the Lagrangian multiplier  corresponding to the equality constraint $\tilde g_i -\epsilon_i + u_i = 0$  in \eqref{eq:bb-csts-rr}  and updated as $ \iter{\mu_i}{t+1}
     = \iter{\mu_i}{t} + \rho ( \widetilde{g}_i ( \iter{\bz}{t+1},  \{\iter{\btheta^c}{t+1}, \iter{\tbtheta^d}{t+1} \}; \mathcal{A} ) -  \epsilon_i + \iter{u_i}{t+1} )$ for $\forall i \in [M]$.
\end{algorithmic}
\end{algorithm*}

\subsection{4.2~ ADMM with  black-box constraints}
We next consider problem \eqref{eq: prob0-rw}
in the presence of black-box constraints  \eqref{eq: prob0-csts}. 
Without loss of generality, we assume that   $\epsilon_i \geq 0$ for $i \in [M]$.
By introducing scalars $u_i \in [0, \epsilon_i]$, we can reformulate the inequality constraint  \eqref{eq: prob0-csts} as the equality constraint together with a box constraint 
{\small\begin{equation}
    \label{eq: prob0-csts1}
    g_i\left(\bz, \left  \{ \btheta^c, \btheta^d \right \}; \mathcal{A} \right)- \epsilon_i + u_i,~ u_i \in [0, \epsilon_i],~ i \in [M].
 \end{equation}}%
We then introduce a continuous surrogate black-box functions $\widetilde{g}_i$ for $g_i,  \forall i \in [M]$ in a similar manner to $\widetilde{f}$ given by \eqref{eq:bbloss-continuous-surrogate}.
Following the reformulation of \eqref{eq: prob0-rw} that lends itself to the application of ADMM, the version with  black-box constraints  \eqref{eq: prob0-csts1} can be equivalently transformed into
{\small
\begin{equation}
\hspace*{-0.2in}\begin{array}{l}
   \displaystyle \min_{\bz, \btheta^c, \tbtheta^d, \bdelta}
\widetilde{f} \left(\bz,  \left\{\btheta^c, \tbtheta^d \right\}; \mathcal{A} \right)  \\
    \st
\left\{
\begin{array}{l}
\bz_i \in \{0,1 \}^{K_i}, \mathbf 1^{\top} \bz_i = 1, \forall i \in [N] \\
\btheta_{ij}^c \in \mathcal{C}_{ij}, \tbtheta_{ij}^d \in \widetilde{\mathcal{D}}_{ij}, \forall i \in [ N ], j \in [ K_i ] \\
\bdelta_{ij} \in \mathcal{D}_{ij}, \forall i \in [ N ], j \in [ K_i ] \\
\tbtheta_{ij}^d  = \bdelta_{ij}, \forall i \in [ N ], j \in [ K_i ] \\
u_i \in [0, \epsilon_i], \forall i \in [M] \\
\widetilde{g}_i \left(\bz,  \left\{\btheta^c, \tbtheta^d \right\}; \mathcal{A} \right)
-  \epsilon_i + u_i = 0, \forall i \in [M].
\end{array}
\right.
\end{array}
\hspace*{-0.3in}
\label{eq:bb-csts-rr}
\end{equation}}
Compared to problem \eqref{eq: prob0-rr},
the introduction of auxiliary variables $\{ u_i\}$ 
enables ADMM to incorporate \textit{black-box equality} constraints as well as elementary \textit{white-box}  constraints. Similar to Algorithm\,1, the ADMM solution to problem \eqref{eq:bb-csts-rr} can be achieved by solving three sub-problems \prnote{of similar nature}, summarized in Algorithm\,2 and derived in Appendix \ref{app: ADMM_blk}.

{We remark that the integration of ADMM and gradient-free operations was also  considered in \cite{liu2018zeroth} and \cite{aricoll17,ariafar2019admmbo}, where the former used randomized gradient estimator when  optimizing a black-box smooth objective function, and the latter used  Bayesian optimization (BO) as the internal solver to solve  black-box optimization problems with black-box constraints. However, the aforementioned works  cannot directly be applied to tackling our considered AutoML problem, which requires a more involved splitting over hypeparameters and model selection variables. Moreover, different from \cite{ariafar2019admmbo}, we handle the black-box inequality constraint  through the reformulated   equality constraint \eqref{eq: prob0-csts1}.  By contrast, the work \cite{ariafar2019admmbo} introduced an indicator function for a black-box constraint and further handled  it by modelling  as a Bernoulli
random  variable.}

\subsection{4.3~ Implementation and convergence}
We highlight that our ADMM based scheme is not a single AutoML algorithm but rather a framework that can be used to mix and match different existing  black-box solvers. \dvnote {This is especially useful since this enables the end-user to plug-in efficient solvers tailored for the sub-problems (HPO \& algorithm selection in our case)}.
\dvnote{In addition to the above}, the ADMM decomposition allows us to solve simpler sub-problems with a smaller number of optimization variables \dvnote{(a significantly reduced search space since \eqref{eq: theta_min} only requires optimization over the active set of continuous variables).}
Unless specified otherwise, we adopt Bayesian optimization (BO) to solve the HPO \eqref{eq: theta_min}, e.g.,  \eqref{eq:cbb-active}. We use customized Thompson sampling to solve the
combinatorial multi-armed bandit problem, namely, the \eqref{eq: z_step} for algorithm selection. 
We refer readers to Appendix\,\ref{asec:gpopt} and \ref{asec:cmab} for more   derivation and implementation details.
\prnote{%
In Appendix\,\ref{sec:expts:flex}, we demonstrate
the generalizability of ADMM to different solvers   for \eqref{eq: theta_min} and \eqref{eq: z_step}.%
}%
The theoretical convergence guarantees of ADMM have been established under certain assumptions, e.g., convexity or smoothness  \citep{boyd2011distributed,hong2017linear,liu2018zeroth}. 
Unfortunately, the  AutoML problem    violates  these restricted assumptions. Even for non-ADMM  based AutoML pipeline search, 
 there is no  theoretical  convergence   established in the existing baselines to the best of our knowledge.  
\prnote{%
Empirically, we will demonstrate the improved convergence of the proposed scheme against baselines in the following section.%
}
\section{Empirical Evaluations} \label{sec:expts}
In this evaluation of our proposed framework, we demonstrate three important characteristics: 
\textcolor{Sijia_color}{
(i) the empirical performance against existing AutoML toolkits, {\em highlighting the empirical competitiveness of the theoretical formalism},}
\prnote{%
(ii) the systematic capability to handle black-box constraints, {\em enabling AutoML to address real-world ML tasks, and}%
}
\textcolor{Sijia_color}{
(iii) the flexibility to incorporate various learning procedures and solvers for the sub-problems, {\em highlighting that our proposed scheme is not a single algorithm but a complete framework} for AutoML pipeline configuration. 
}
%

\noindent
\textbf{Data and black-box objective function.} We consider $30$ binary classification\footnotemark datasets from the UCI ML \citep{asuncion2007uci} \& 
\footnotetext{
Our scheme applies to multiclass classification \& regression. 
}
OpenML repositories \citep{bischl2017openml}, and Kaggle.
We consider a subset of OpenML100 limited to binary classification and small enough to allow for meaningful amount of optimization for {\em all baselines} in the allotted 1 hour to ensure that we are evaluating the optimizers and not the initialization heuristics.
Dataset details are in Appendix \ref{asec:data}. We consider $(1 - \mbox{AUROC})$ ({\em {\bf a}rea {\bf u}nder the {\bf ROC} curve}) as the black-box objective and evaluate it on a 80-20\% train-validation split for all baselines. We consider AUROC since it is a meaningful predictive performance metric regardless of the class imbalance (as opposed to classification error).
\noindent\textbf{Comparing ADMM to AutoML baselines.} 
%
Here we evaluate the proposed ADMM framework against widely used AutoML systems Auto-sklearn \citep{feurer2015efficient} and TPOT \citep{olson2016tpot}. This comparison is limited to black-box optimization with {\em analytic constraints only} given by  \eqref{eq: prob0} since existing AutoML toolkits {\em cannot handle black-box constraints explicitly}. 
\prnote{
We consider SMAC based vanilla Auto-sklearn ASKL\footnotemark (disabling ensembles and meta-learning),
}
%
\footnotetext{
\prnote{%
Meta-learning and ensembling in ASKL are preprocessing and postprocessing steps respectively to the actual black-box optimization and can be applied to any optimizer. We demonstrate this for ADMM in Appendix \ref{asec:ensembles}. So we skip these aspects of ASKL here.%
}
}%
\prnote{%
random search RND, and TPOT with a population of $50$ (instead of the default $100$) to ensure that TPOT is able to process multiple generations of the genetic algorithm in the allotted time on all data sets.%
}
For ADMM, we utilize 
\prnote{BO}
for \eqref{eq: theta_min} 
and CMAB for \eqref{eq: z_step} -- ADMM(BO,Ba)\footnotemark. 
%
\footnotetext{
In this setup, ADMM has 2 parameters: (i) the penalty $\rho$ on the augmented term, (ii) the loss upper-bound $\hat{f}$ in the CMAB algorithm 
(Appendix \ref{asec:cmab}). 
We evaluate the sensitivity of ADMM on these parameters in Appendix \ref{asec: para_sensitivity}. The results indicate that 
ADMM 
is fairly robust to these parameters, and hence set $\rho=1$ and $\hat{f} = 0.7$ throughout. We start the ADMM optimization with $\boldsymbol{\lambda}^{(0)} = \mathbf 0$. 
}
%
%
%
%
\begin{figure}[htb]
    \centering
    \begin{subfigure}{0.3\textwidth}
    \includegraphics[width=\textwidth]{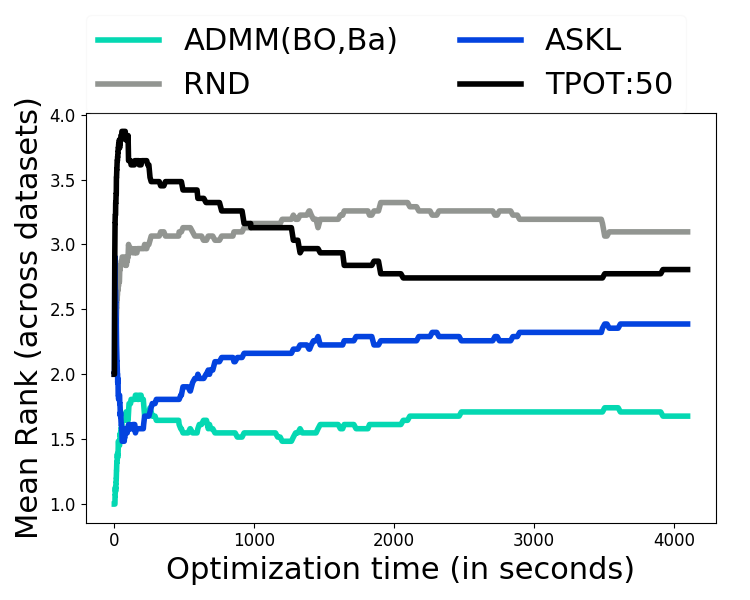}
    \caption{All methods}
    \end{subfigure}
    \\
    \begin{subfigure}{0.17\textwidth}
    \includegraphics[width=\textwidth]{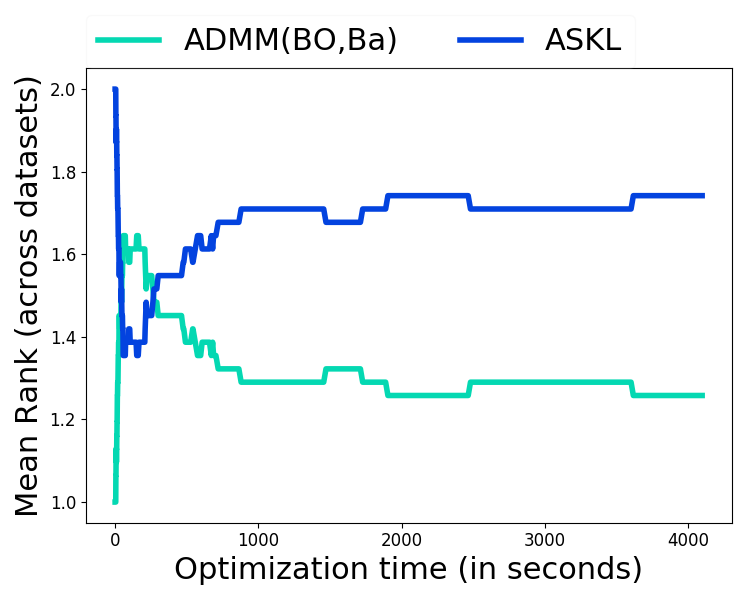}
    \caption{ASKL vs. ADMM}
    \label{fig:admm-askl}
    \end{subfigure}
    \hspace{0.2in}
    \begin{subfigure}{0.17\textwidth}
    \includegraphics[width=\textwidth]{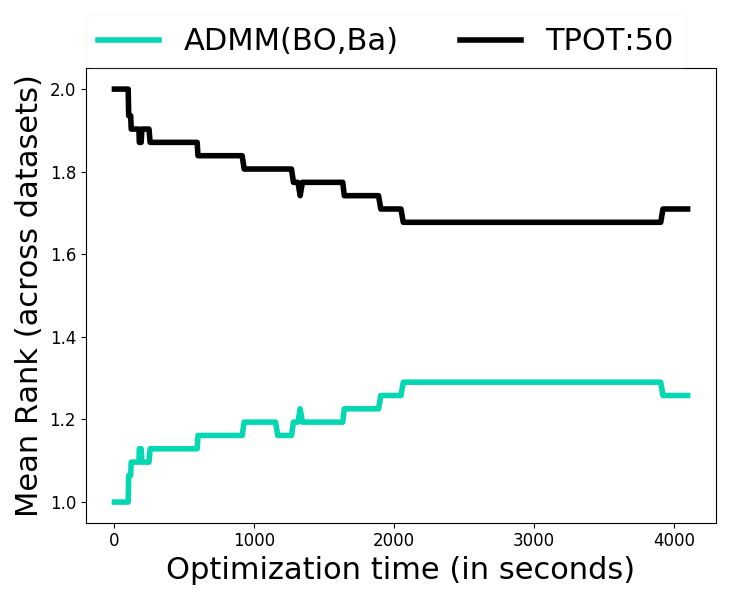}
    \caption{TPOT50 vs. ADMM}
    \label{fig:admm-tpot}
    \end{subfigure}
    \caption{Average rank (across 30 datasets) of mean performance across 10 trials -- {\em lower rank is better}. 
    }
    \label{fig:avg-ranks-mean-obj}
\end{figure}
%

\prnote{%
For all optimizers, we use \texttt{scikit-learn} algorithms \citep{scikit_learn}. The functional modules and the algorithms (with their hyper-parameters) are presented in Table \ref{table:search-space-automl} in Appendix \ref{asec:baselines-search-space}. We maintain parity\footnotemark across the various AutoML baselines by searching over the same set of algorithms
\footnotetext{ASKL and ADMM search over the same search space of {\em fixed pipeline shape \& order}. TPOT also searches over different pipeline shapes \& orderings because of the nature of its genetic algorithm.}
(see Appendix \ref{asec:baselines-search-space}). For each scheme, the algorithm hyper-parameter ranges are set using Auto-sklearn as the reference\footnotemark.%
}
\footnotetext{{\scriptsize \url{github.com/automl/auto-sklearn/tree/master/autosklearn/pipeline/components}}}
We optimize for 1 hour \& generate time vs. incumbent black-box objective curves aggregated over 10 trials. Details on the complete  setup are in Appendix \ref{asec:expt-details}. The optimization convergence for all 30 datasets are in Appendix \ref{asec:cvgplot}. At completion, ASKL achieves the lowest mean objective (across trials) in 6/30 datasets, TPOT50 in 8/30, RND in 3/30 and ADMM(BO,Ba) in 15/30, showcasing ADMM's effectiveness.
%
%
%
%
%
%

\textcolor{Sijia_color}{
 Figure \ref{fig:avg-ranks-mean-obj} presents the overall   performance of the different AutoML schemes versus optimization time. Here we consider the relative rank of each scheme (with respect to the mean objective over $10$ trials) for every timestamp, and average this rank across $30$ data sets 
similar to the comparison in \citet{feurer2015efficient}.}
\prnote{%
With enough time, all schemes outperform random search RND. TPOT50 performs worst in the beginning because of the initial start-up time involved in the genetic algorithm. ASKL and ADMM(BO,Ba) have comparable performance initially. As the optimization continues, ADMM(BO,Ba) significantly outperforms all other baselines. We present the pairwise performance of ADMM with ASKL (figure \ref{fig:admm-askl}) \& TPOT50 (figure \ref{fig:admm-tpot}).%
}

%
\noindent \textbf{AutoML with black-box constraints.} 
%
To demonstrate the capability of the ADMM framework to incorporate real-world black-box constraints, 
we consider the recent Home Credit Default Risk Kaggle challenge\footnotemark 
%
\footnotetext{{\scriptsize \url{www.kaggle.com/c/home-credit-default-risk}}}
with the black-box objective of $(1 - \mbox{AUROC})$, and 2 black-box constraints: (i) ({\bf deployment}) Prediction latency $t_p$ enforcing real-time predictions, (ii) ({\bf fairness}) Maximum pairwise disparate impact $d_I$ \citep{calders2010three} across all loan applicant age groups enforcing fairness across  groups (see Appendix \ref{asec:group-disparity}).
%
\begin{figure}[tb]
    \centering
    \begin{subfigure}{0.225\textwidth}
    \includegraphics[width=\textwidth]{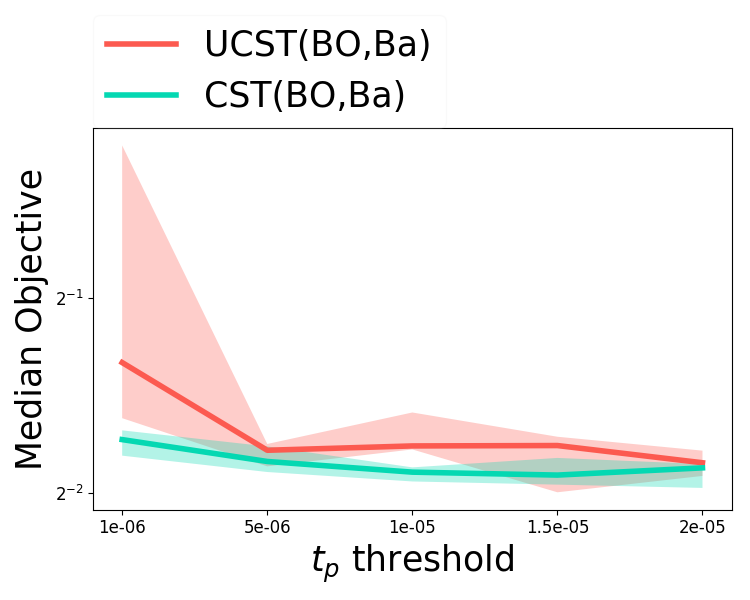}
    \caption{Varying $t_p$, $d_I = 0.07$}
    \label{fig:obj-t_p}
    \end{subfigure}
    ~
    \begin{subfigure}{0.225\textwidth}
    \includegraphics[width=\textwidth]{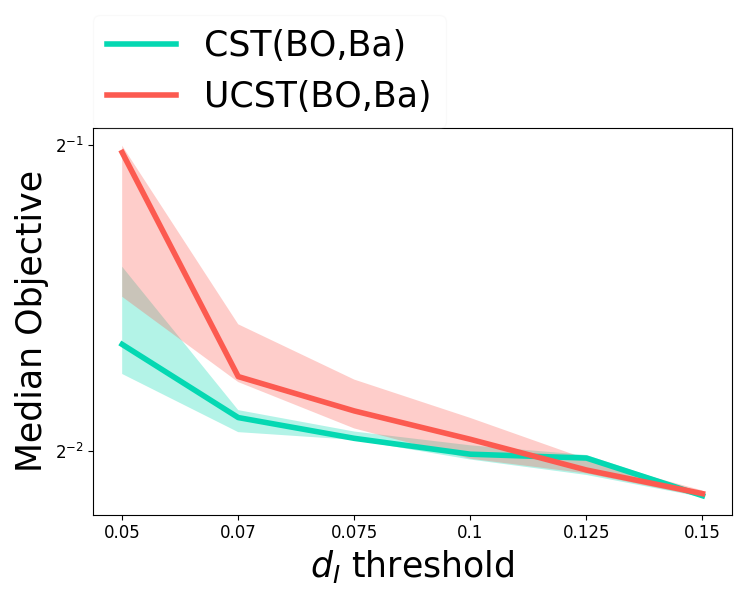}
    \caption{Varying $d_I$, $t_p = 10\mu s$}
    \label{fig:obj-d_i}
    \end{subfigure}
    \caption{Best objective achieved by any constraint satisfying pipeline from running the optimization for 1 hour seconds with varying thresholds for the two constraints -- {\em lower is better} ({\em please view in color}). Note the log-scale on the vertical axis.}
    \label{fig:bbo-bbc-obj}
\end{figure}

We run a set of experiments for each of the constraints: (i) fixing $d_I = 0.7$, we optimize for each of the thresholds $t_p = \{1, 5, 10, 15, 20\}$ (in $\mu s$), and (ii) fixing $t_p = 10\mu s$ and we optimize for each of $d_I = \{ 0.05, 0.075, 0.1, 0.125, 0.15 \}$.
Note that the constraints get less restrictive as the thresholds increase. We apply ADMM to the unconstrained problem (UCST) and post-hoc filter constraint satisfying pipelines to demonstrate that these constraints are not trivially satisfied. 
Then we execute ADMM with these constraints (CST). 
Using BO for \eqref{eq: theta_min} \& CMAB for \eqref{eq: z_step}, we get two variants -- UCST(BO,Ba) \& CST(BO,Ba). 
This results in $(5 + 5)\times 2 = 20$ ADMM executions, each repeated $10 \times$.
\begin{figure}[b]
    \centering
    \begin{subfigure}{0.225\textwidth}
    \includegraphics[width=\textwidth]{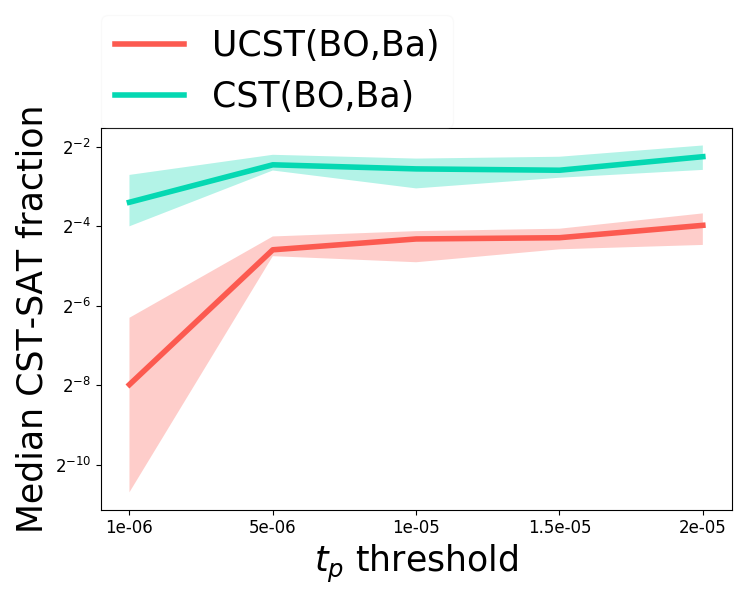}
    \caption{Varying $t_p$, $d_I = 0.07$}
    \label{fig:csat-t_p}
    \end{subfigure}
    ~
    \begin{subfigure}{0.225\textwidth}
    \includegraphics[width=\textwidth]{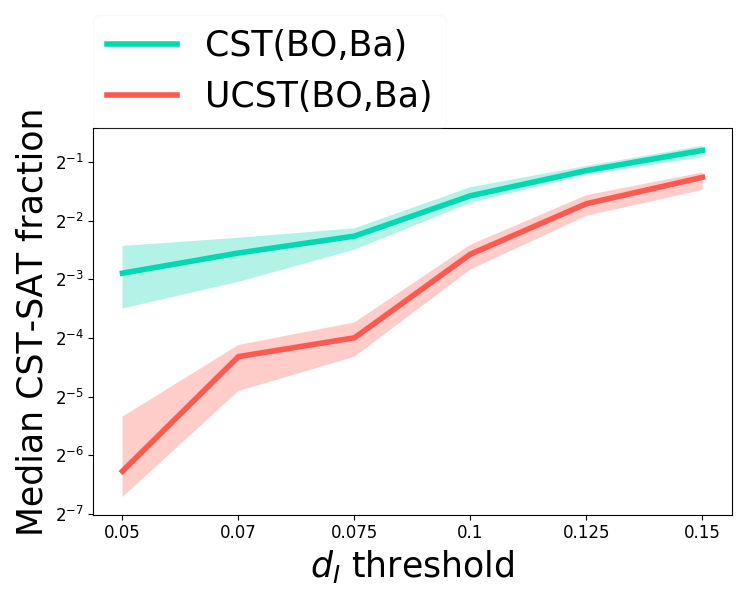}
    \caption{Varying $d_I$, $t_p = 10\mu s$}
    \label{fig:csat-d_i}
    \end{subfigure}
    \caption{Fraction of pipelines found satisfying constraints  with optimization for 1 hour with varying thresholds for the 2 constraints -- {\em higher is better}. 
    Note the log-scale on the vertical axis.}
    \label{fig:bbo-bbc-csat}
\end{figure}

Figure \ref{fig:bbo-bbc-obj} presents the objective achieved by the optimizer when limited only to constraint satisfying pipelines. Figure \ref{fig:obj-t_p} presents the effect of relaxing the constraint on $t_p$ while Figure \ref{fig:obj-d_i} presents the same for the constraint on $d_I$. As expected, the objective improves as the constraints relax. In both cases, CST outperforms UCST, with UCST approaching CST as the constraints relax. Figure \ref{fig:bbo-bbc-csat} presents the constraint satisfying capability of the optimizer by considering the fraction of constraint-satisfying pipelines found (Figure \ref{fig:csat-t_p} \& \ref{fig:csat-d_i} for varying $t_p$ \& $d_I$ respectively). CST again significantly outperforms UCST, indicating that the constraints are non-trivial to satisfy, and that ADMM is able to effectively incorporate the constraints for improved performance.

\noindent
\textbf{\textcolor{Sijia_color}{Flexibility \& benefits from ADMM operator splitting.}}
It is common in ADMM to solve the sub-problems to higher approximation in the initial iterations and to an increasingly lower approximation as ADMM progresses (instead of the same approximation throughout) \citep{boyd2011distributed}. 
We demonstrate (empirically) that this {\em adaptive ADMM} produces expected gains in the AutoML problem.
Moreover, we show the empirical gains of ADMM from (i) splitting the AutoML problem \eqref{eq: prob0} into smaller sub-problems which are solved in an alternating fashion, \& (ii) using different solvers for the differently structured \eqref{eq: theta_min} and \eqref{eq: z_step}.

First we use BO for both \eqref{eq: theta_min} and \eqref{eq: z_step}. For ADMM with a fixed approximation level ({\em fixed ADMM}), we solve the sub-problems with BO to a fixed number $I = 16, 32, 64, 128$ iterations, denoted by ADMM$I$(BO,BO) (e.g., ADMM16(BO,BO)). For adaptive ADMM, we start with $16$ BO iterations for the sub-problems and progressively increase it with an additive factor $F = 8$ \& $16$ with every ADMM iteration until $128$ denoted by AdADMM-F8(BO,BO) \& AdADMM-F16(BO,BO) respectively. We optimize for 1 hour and aggregate over 10 trials. 
%
%
%
\begin{figure}[t]
    \centering
    \includegraphics[width=0.32\textwidth]{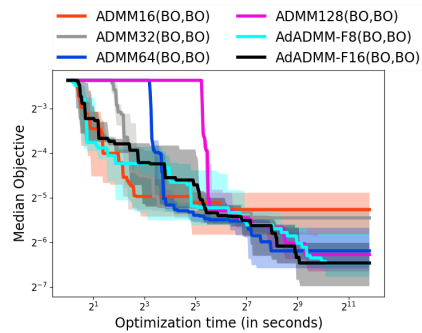}
    \caption{Optimization time (in seconds) vs. median validation performance with the inter-quartile range over 10 trials on fri-c2 dataset -- lower is better ({\em please view in color}). Note the log scale on both axes. See Appendix \ref{asec:adadmm} for additional results.} 
    \label{fig:time-v-valperf-iqr-adadmm}
\end{figure}

\textcolor{Sijia_color}{
Figure \ref{fig:time-v-valperf-iqr-adadmm} presents optimization convergence for 1 dataset (fri-c2).}
We see the expected behavior -- fixed ADMM with small $I$ dominate for small time scales but saturate soon; large $I$ require significant start-up time but dominate for larger time scales. Adaptive ADMM ($F=8$ \& $16$) is able to match the performance of the best fixed ADMM at every time scale.%
Please refer to Appendix \ref{asec:adadmm} for additional results.

\textcolor{Sijia_color}{Next, we illustrate the advantage of ADMM on  operator splitting. We consider 2 variants,  
AdADMM-F16(BO,BO) and AdADMM-F16(BO,Ba), where the latter uses CMAB for \eqref{eq: z_step}. For comparison, we  solve the complete {\em joint problem} \eqref{eq: prob0} with BO, leading to a Gaussian Process with a large number of variables, denoted as JOPT(BO).%
}
%
%
\begin{figure}[tb]
  \centering
  \includegraphics[width=0.32\textwidth]{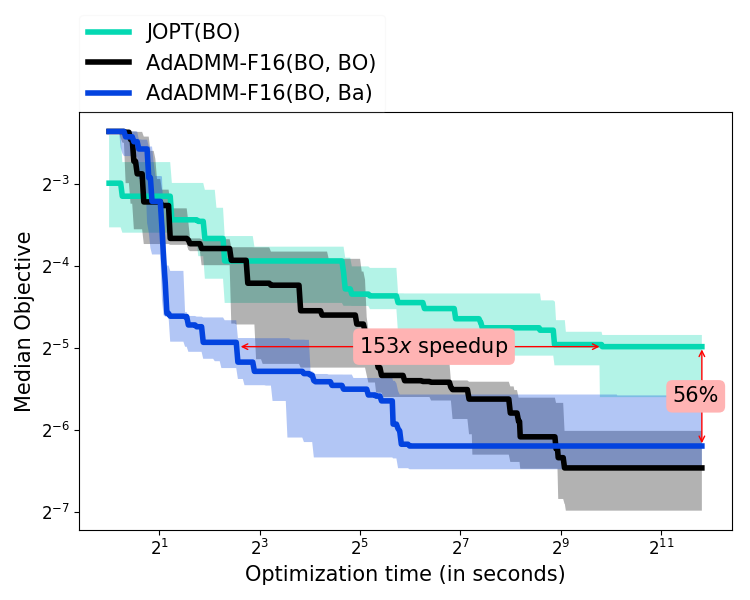}
  \caption{Optimization time vs. median validation performance with the inter-quartile range over 10 trials on fri-c2 dataset -- lower is better ({\em please view in color}). Note the log scale on both axes. See Appendix \ref{asec:opsplit} for additional results.}
  \label{fig:time-v-valperf-iqr-opsplit}
\end{figure}

Figure \ref{fig:time-v-valperf-iqr-opsplit} shows the optimization convergence for 1 dataset (fri-c2). 
The results indicate that the operator splitting in ADMM provides significant improvements over JOPT(BO), with ADMM reaching the final objective achieved by JOPT with significant speedup, and then further improving upon that final objective significantly. These improvements of ADMM over JPOT on 8 datasets are summarized in Table \ref{tab:opsplit-gains}, indicating significant speedup (over $10\times$ in most cases) and further improvement (over $10\%$ in many cases).

Let us use $\mathbf{S}_{\mbox{Ba}}$ and $\mathbf{S}_{\mbox{BO}}$ to represent the temporal speedup achieved by AdADMM(BO,Ba) and AdADMM(BO,BO) (eliding ``-F16'') respectively to reach the best objective of JOPT, and similarly use $\mathbf{I}_{\mbox{Ba}}$ and $\mathbf{I}_{\mbox{BO}}$ to represent the objective improvement at the final converged point. Table \ref{tab:opsplit-gains}    shows that between AdADMM(BO,BO) and AdADMM(BO,Ba), the latter provides {\em significantly higher speedups}, but the former provides higher additional improvement in the final objective. 
This demonstrates ADMM's flexibility, for example, allowing choice between faster or more improved solution.
%
%
%
%
%
\begin{table}[t]
    \centering
    \begin{adjustbox}{max width=0.3\textwidth }
    \begin{tabular}{|l|r|r|r|r|}
        \hline 
        Dataset & $\mathbf{S}_{\mbox{Ba}}$  & $\mathbf{S}_{\mbox{BO}}$ & $\mathbf{I}_{\mbox{Ba}}$ & $\mathbf{I}_{\mbox{BO}}$ \\
        \hline\hline
        Bank8FM    & $10 \times$ & $2 \times$ & $0 \%$ & $5 \%$\\
        CPU small  & $4  \times$ & $5 \times$ & $0 \%$ & $5 \%$\\
        fri-c2     & $153\times$ & $25\times$ & $56\%$ & $64\%$\\
        PC4        & $42 \times$ & $5 \times$ & $8 \%$ & $13\%$\\
        Pollen     & $25 \times$ & $7 \times$ & $4 \%$ & $3 \%$\\
        Puma8NH    & $11 \times$ & $4 \times$ & $1 \%$ & $1 \%$\\
        Sylvine    & $9  \times$ & $2 \times$ & $9 \%$ & $26\%$\\
        Wind       & $40 \times$ & $5 \times$ & $0 \%$ & $5 \%$\\
        \hline
    \end{tabular}
    \end{adjustbox}
    \caption{Comparing ADMM schemes to JOPT(BO), we list the speedup $\mathbf{S}_{\mbox{Ba}}$ \& $\mathbf{S}_{\mbox{BO}}$ achieved by AdADMM(BO,Ba) \& AdADMM(BO,BO) respectively to reach the best objective of JOPT, and the final objective improvement $\mathbf{I}_{\mbox{Ba}}$ \& $\mathbf{I}_{\mbox{BO}}$ (respectively) over the JOPT objective.
    These numbers are generated using the aggregate performance of JOPT and AdADMM over 10 trials.%
    }
    \label{tab:opsplit-gains}
\end{table}
%

%
\section{Conclusions}\label{sec:conclusions}
Posing the problem of joint algorithm selection and HPO for automatic pipeline configuration in AutoML as a {\em formal mixed continuous-integer nonlinear program}, we leverage the ADMM optimization framework to {\em decompose} this problem into {\em 2 easier sub-problems}: (i) black-box optimization with {\em a small set of continuous variables}, and (ii) a combinatorial optimization problem involving {\em only Boolean variables}. These sub-problems can be effectively addressed by existing AutoML techniques, allowing ADMM to solve the overall problem effectively. This scheme also {\em seamlessly incorporates black-box constraints} alongside the black-box {\em objective}. We empirically demonstrate the flexibility of the proposed ADMM framework to leverage existing AutoML techniques and its effectiveness against open-source baselines. 

\bibliographystyle{aaai}
\bibliography{ref}
%
\clearpage
\newpage
\setcounter{section}{0}
\setcounter{figure}{0}

\onecolumn
\section*{Appendices of `An ADMM Based Framework for AutoML Pipeline Configurations'}

\makeatletter 
\renewcommand{\thefigure}{A\@arabic\c@figure}
\makeatother
\setcounter{table}{0}
\renewcommand{\thetable}{A\arabic{table}}
\setcounter{algorithm}{0}
\renewcommand{\thealgorithm}{A\arabic{algorithm}}
\renewcommand{\theequation}{A\arabic{equation}}
\renewcommand{\thesubfigure}{\roman{subfigure}}

\section{Derivation of ADMM sub-problems in Table\,1} 
\label{app:thr_admm}

ADMM decomposes the optimization variables into two blocks and alternatively minimizes the augmented Lagrangian function \eqref{eq:prob1-al} in the following manner at any ADMM iteration $t$
\begin{align}
    \label{eq:cont-bb} \left\{ \iter{\btheta^c}{t+1}, \iter{\tbtheta^d}{t+1} \right\} 
    & = \argmin_{\btheta^c, \tbtheta^d} 
    \mathcal{L}\left( \iter{\bz}{t}, \btheta^c, \tbtheta^d, \iter{\bdelta}{t}, \iter{\blambda}{t} \right) \\
    \label{eq:int-bb} \left\{ \iter{\bdelta}{t+1}, \iter{\bz}{t+1} \right\} 
    & = \argmin_{\bdelta, \bz} 
    \mathcal{L}\left( \bz, \iter{\btheta^c}{t+1}, \iter{\tbtheta^d}{t+1}, \bdelta, \iter{\blambda}{t} \right) \\
    \label{eq:lambda-update} \iter{\blambda}{t+1}
    & = \iter{\blambda}{t} + \rho \left( \iter{\tbtheta^d}{t+1} - \iter{\bdelta}{t+1} \right).
\end{align}
Problem \eqref{eq:cont-bb} can be simplified by removing constant terms to get
\begin{align}
    \label{eq:cont-bb-1} \left\{ \iter{\btheta^c}{t+1}, \iter{\tbtheta^d}{t+1} \right\} 
    & = \argmin_{\btheta^c, \tbtheta^d}
    \widetilde{f} \left(\iter{\bz}{t},  \left\{\btheta^c, \tbtheta^d \right\}; \mathcal{A} \right)
    + I_\mathcal{C}(\btheta^c)
    + I_{\widetilde{\mathcal{D}}}(\tbtheta^d) \\
    \nonumber & \quad \quad \quad \quad \quad
    + \iter{\blambda}{t}{}^\top \left( \tbtheta^d - \iter{\bdelta}{t} \right) 
    + \frac{\rho}{2} \left\| \tbtheta^d - \iter{\bdelta}{t} \right\|_2^2, \\
    \label{eq:cont-bb-2} & = \argmin_{\btheta^c, \tbtheta^d} 
    \widetilde{f} \left(\iter{\bz}{t},  \left\{\btheta^c, \tbtheta^d \right\}; \mathcal{A} \right)
    + I_\mathcal{C}(\btheta^c)
    + I_{\widetilde{\mathcal{D}}}(\tbtheta^d) 
    + \frac{\rho}{2} \left\| \tbtheta^d - \mathbf{b} \right\|_2^2 \\
    \nonumber &  \quad \quad \quad \quad \quad
    \mbox{ where } \mathbf{b} = \iter{\bdelta}{t} - \frac{1}{\rho} \iter{\blambda}{t}.
\end{align}
A similar treatment to problem \eqref{eq:int-bb} gives us
\begin{align}
    \label{eq:int-bb-1} \left\{ \iter{\bdelta}{t+1}, \iter{\bz}{t+1} \right\} 
    & = \argmin_{\bdelta, \bz} 
    \widetilde{f} \left(\bz,  \left\{\iter{\btheta^c}{t+1}, \iter{\tbtheta^d}{t+1} \right\}; \mathcal{A} \right)
    + I_\mathcal{Z}(\bz) \\
    \nonumber & \quad \quad \quad \quad \quad
    + I_\mathcal{D}(\bdelta) 
    + \iter{\blambda}{t}{}^\top \left( \iter{\tbtheta^d}{t+1} - \bdelta \right) 
    + \frac{\rho}{2} \left\| \iter{\tbtheta^d}{t+1} - \bdelta \right\|_2^2, \\
    \label{eq:int-bb-2} & = \argmin_{\bdelta, \bz} 
    \widetilde{f} \left(\bz,  \left\{\iter{\btheta^c}{t+1}, \iter{\tbtheta^d}{t+1} \right\}; \mathcal{A} \right)
    + I_\mathcal{Z}(\bz) \\
    \nonumber & \quad \quad \quad \quad \quad
    + I_\mathcal{D}(\bdelta) 
    + \frac{\rho}{2} \left\| \mathbf{a} - \bdelta \right\|_2^2 \mbox{ where } \mathbf{a} = \iter{\tbtheta^d}{t+1} + \frac{1}{\rho} \iter{\blambda}{t}.\\
\end{align}
This simplification exposes the independence between $\bz$ and $\bdelta$, allowing us to solve problem \eqref{eq:int-bb} independently for $\bz$ and $\bdelta$ as:
\begin{align}
    \label{eq:int-bb-2d} \iter{\bdelta}{t+1} 
    & = \argmin_{\bdelta} 
    I_\mathcal{D}(\bdelta) 
    + \frac{\rho}{2} \left\| \mathbf{a} - \bdelta \right\|_2^2 \mbox{ where } \mathbf{a} = \iter{\tbtheta^d}{t+1} + \frac{1}{\rho} \iter{\blambda}{t}, \\
    \label{eq:int-bb-2z} \iter{\bz}{t+1}
    & = \argmin_\bz
    \widetilde{f} \left(\bz,  \left\{\iter{\btheta^c}{t+1}, \iter{\tbtheta^d}{t+1} \right\}; \mathcal{A} \right)
    + I_\mathcal{Z}(\bz).
\end{align}
So we are able to decompose problem (3) into problems \eqref{eq:cont-bb-2}, \eqref{eq:int-bb-2d} and \eqref{eq:int-bb-2z} which can be solved iteratively along with the $\blambda^{(t)}$ updates (see Table 1).
\hfill $\square$

%

\section{Derivation of ADMM sub-problems in Table\,2} 
\label{app: ADMM_blk}
Defining $\mathcal{U} = \{ \mathbf{u} \colon \mathbf{u} = \{ u_i \in [0, \epsilon_i] \forall i \in [M] \} \}$, we can go through the mechanics of ADMM to get the augmented Lagrangian with $\blambda$ and $\mu_i \forall i \in [M]$ as the Lagrangian multipliers and $\rho > 0$ as the penalty parameter as follows:
\begin{equation}
    \begin{split}
    \label{eq:bbc-al}
    \mathcal{L}\left( \bz, \btheta^c, \tbtheta^d, \bdelta, \bu, \blambda, \boldsymbol{\mu} \right) =
    & \quad
    \widetilde{f} \left(\bz,  \left\{\btheta^c, \tbtheta^d \right\}; \mathcal{A} \right)
    + I_\mathcal{Z}(\bz) + I_\mathcal{C}(\btheta^c)
    + I_{\widetilde{\mathcal{D}}}(\tbtheta^d) 
    + I_\mathcal{D}(\bdelta) \\
    & \quad
    + \blambda^\top \left( \tbtheta^d - \bdelta \right) 
    + \frac{\rho}{2} \left\| \tbtheta^d - \bdelta \right\|_2^2 \\
    & \quad I_\mathcal{U}(\bu)
    + \sum_{i=1}^{M} \mu_i \left( \widetilde{g}_i \left(\bz,  \left\{\btheta^c, \tbtheta^d \right\}; \mathcal{A} \right) -  \epsilon_i + u_i \right) \\
    & \quad
    + \frac{\rho}{2} \sum_{i=1}^{M} \left( 
    \widetilde{g}_i \left(\bz,  \left\{\btheta^c, \tbtheta^d \right\}; \mathcal{A} \right) -  \epsilon_i + u_i 
    \right)^2.
    \end{split}
\end{equation}
ADMM decomposes the optimization variables into two blocks for alternate minimization of the augmented Lagrangian in the following manner at any ADMM iteration $t$
\begin{align}
    \label{eq:cbbc} \left\{ \iter{\btheta^c}{t+1}, \iter{\tbtheta^d}{t+1},
    \iter{\bu}{t+1}\right\} 
    & = \argmin_{\btheta^c, \tbtheta^d, \bu} 
    \mathcal{L}\left( \iter{\bz}{t}, \btheta^c, \tbtheta^d, \iter{\bdelta}{t}, \bu, \iter{\blambda}{t}, \iter{\boldsymbol{\mu}}{t} \right) \\
    \label{eq:ibbc} \left\{ \iter{\bdelta}{t+1}, \iter{\bz}{t+1} \right\} 
    & = \argmin_{\bdelta, \bz} 
    \mathcal{L}\left( \bz, \iter{\btheta^c}{t+1}, \iter{\tbtheta^d}{t+1}, \bdelta, \iter{\bu}{t+1}, \iter{\blambda}{t}, \iter{\boldsymbol{\mu}}{t} \right) \\
    \label{eq:lup} \iter{\blambda}{t+1}
    & = \iter{\blambda}{t} + \rho \left( \iter{\tbtheta^d}{t+1} - \iter{\bdelta}{t+1} \right) \\
    \label{eq:mup} \forall i \in [M],\ \  \iter{\mu_i}{t+1}
    & = \iter{\mu_i}{t} + \rho \left( \widetilde{g}_i ( \iter{\bz}{t+1},  \{\iter{\btheta^c}{t+1}, \iter{\tbtheta^d}{t+1} \}; \mathcal{A} ) -  \epsilon_i + \iter{u_i}{t+1} \right).
\end{align}
Note that, unlike the unconstrained case, the update of the augmented Lagrangian multiplier $\mu_i$ requires the evaluation of the black-box function for the constraint $g_i$.

Simplifying problem \eqref{eq:cbbc} gives us
\begin{equation}
\begin{split}
    \label{eq:cbbc-s-0}
    \min_{\btheta^c, \tbtheta^d, \bu} & \quad
    \widetilde{f} \left( \iter{\bz}{t},  \left\{\btheta^c, \tbtheta^d \right \}; \mathcal{A} \right) \\
    & \quad \quad \quad
    + \frac{\rho}{2} \left[ 
    \left\| \tbtheta^d - \mathbf{b} \right\|_2^2 
    + \sum_{i=1}^{M} \left[ 
    \widetilde{g}_i \left( \iter{\bz}{t},  \left\{\btheta^c, \tbtheta^d \right \}; \mathcal{A} \right) 
    -  \epsilon_i + u_i  + \frac{1}{\rho}\iter{\mu_i}{t}
    \right]^2
    \right]
    \\ 
    & \st 
    \left\{
    \begin{array}{l}
    \btheta_{ij}^c \in \mathcal{C}_{ij} \forall i \in [N], j \in [K_i],\\
    \tbtheta_{ij}^d \in \widetilde{\mathcal{D}}_{ij} \forall i \in [N], j \in [K_i], \\
    u_i \in [0, \epsilon_i],
    \end{array}
    \right.
    \mbox{ where } \mathbf{b} = \iter{\bdelta}{t} - \frac{1}{\rho} \iter{\blambda}{t},
\end{split}
\end{equation}
which can be further split into active and inactive set of continuous variables based on the $\iter{\bz}{t}$ as in the solution of problem \eqref{eq:cont-bb-2} (the $\btheta$-min problem). The main difference from the unconstrained case in problem \eqref{eq:cont-bb-2} (the $\btheta$-min problem) to note here is that the black-box optimization with continuous variables now has $M$ new variables $u_i$ ($M$ is the total number of black-box constraints) which are active in every ADMM iteration. This problem \eqref{eq:cbbc-s-0} can be solved in the same manner as problem \eqref{eq:cont-bb-2} ($\btheta$-min) using SMBO or TR-DFO techniques.

Simplifying and utilizing the independence of $\bz$ and $\bdelta$, we can split problem \eqref{eq:ibbc} into the following problem for $\bdelta$
\begin{equation}
    \label{eq:ibbc-sd-0}
    \min_{  \bdelta  } \quad \frac{\rho}{2} \| \bdelta -  {\mathbf a} \|_2^2  
    \quad \st \bdelta_{ij} \in \mathcal D_{ij} \forall i \in [N], j \in [K_i] 
    \mbox{ where } \mathbf{a} = \iter{\tbtheta^d}{t+1} + \frac{1}{\rho} \iter{\blambda}{t},
\end{equation}
which remains the same as problem \eqref{eq:int-bb-2d} (the $\bdelta$-min problem) in the unconstrained case, while the problem for $\bz$ becomes
\begin{equation}
\begin{split}
    \label{eq:ibbc-sz-0}
    \min_\bz & \quad
    \widetilde{f} (\bz,  \{\iter{\btheta^c}{t+1}, \iter{\tbtheta^d}{t+1} \}; \mathcal{A} ) \\
    & \quad \quad \quad 
    + \frac{\rho}{2}
    \sum_{i=1}^{M} \left[ 
    \widetilde{g}_i ( \bz, \{\iter{\btheta^c}{t+1}, \iter{\tbtheta^d}{t+1} \}; \mathcal{A}) 
    -  \epsilon_i + \iter{u_i}{t+1}  + \frac{1}{\rho} \iter{\mu_i}{t}
    \right]^2
    \\
    & \st 
    \bz_i \in \{0,1 \}^{K_i}, \mathbf 1^{\top} \bz_i = 1, \forall i \in [N].
\end{split}
\end{equation}
The problem for $\bz$ is still a black-box integer programming problem, but now with an updated black-box function and can be handled with techniques proposed for the combinatorial problem  \eqref{eq:int-bb-2z} in the absence of black-box constraints (the $\bz$-min problem).
\hfill $\square$

\clearpage\pagebreak

\section{Bayesian Optimization for solving the \eqref{eq: theta_min} problem on the active set} 
\label{asec:gpopt}
Problem \eqref{eq:cbb-active} (\eqref{eq: theta_min} on the active set) is a HPO problem. This can be solved with Bayesian optimization (BO) \citep{shahriari2016taking}. BO has become a core component of various AutoML systems \citep{snoek2012practical}. For any black-box objective function $f(\btheta)$ defined on continuous variables $\btheta \in \mathcal{C}$, BO assumes a statistical model, usually a Gaussian process (GP), for $f$. Based on the observed function values  $\mathbf y =   [ f(\btheta^{(0)}),\ldots, f(\btheta^{(t)}) ]^{\top}$, BO updates the GP and determines the next query point $\btheta^{(t+1)}$ by maximizing the expected improvement (EI) over the posterior GP model. Specifically the objective $f(\btheta)$ is modeled as a GP with a \textit{prior} distribution $f(\cdot) \sim \mathcal N(\mu(\cdot), \kappa(\cdot, \cdot))$, where $\kappa(\cdot, \cdot)$ is a positive definite kernel. Given the observed function values $\mathbf y$, the posterior probability of a new function evaluation $f(\btheta)$ at {iteration $t+1$} is modeled as a Gaussian distribution with mean $\mu(\btheta)$ and variance $\sigma^2(\btheta)$ \citep[Sec.\,III-A]{shahriari2016taking}, where
\begin{equation}
    \label{eq: post_f}
      \mu(\hat {\boldsymbol{\theta}}) = \boldsymbol \kappa^{\top} [\boldsymbol{\Gamma} + \sigma_n^2 \mathbf I]^{-1} \mathbf y 
      \quad \mbox{ and } \quad
      \sigma^2(\hat {\boldsymbol{\theta}})  = \kappa(\hat {\boldsymbol{\theta}} , \hat {\boldsymbol{\theta}}) - \boldsymbol \kappa^{\top} [\boldsymbol{\Gamma} + \sigma_n^2 \mathbf I]^{-1}\boldsymbol \kappa,
\end{equation}
where $\boldsymbol{\kappa}$  is a vector of covariance terms between $\btheta$ and  $\{ \btheta^{(i)} \}_{i=0}^t$, and $\boldsymbol{\Gamma}$ denotes the covariance of  $\{ \btheta^{(i)} \}_{i=0}^t$, namely, $\Gamma_{ij} = \kappa(\btheta^{(i)}, \btheta^{(j)})$, and $\sigma_n^2$ is a small positive number to model the variance of the observation noise.

\begin{remark} \label{remark:ZOSBGD}
To determine
the GP model \eqref{eq: post_f},
we choose the kernel function $\kappa(\cdot,\cdot)$  as the ARD Mat\'ern $5/2$ kernel \citep{snoek2012practical,shahriari2016taking},
\begin{equation}\label{eq: kernel_paras}
    \kappa(\mathbf x, \mathbf x^\prime) = \tau_0^2 \mathrm{exp}(-\sqrt{5} r) 
    ( 1 +  \sqrt{5}r + \frac{5}{3} r^2) 
\end{equation}
for two vectors $\mathbf x, \mathbf x^\prime$, where  $r^2 = \sum_{i=1}^d (x_i - x_i^\prime)^2/\tau_i^2$, and
 $  \{ \tau_i \}_{i=0}^d$ are kernel parameters. We determine the GP hyper-parameters $\boldsymbol{\psi} = \{  \{ \tau_i \}_{i=0}^d,  \sigma_n^2\} $
by minimizing the negative log marginal likelihood  $\log p(\mathbf y | \boldsymbol{\psi})$ \citep{shahriari2016taking},
\begin{equation}\label{eq: learn_hyper}
\minimize_{\boldsymbol{\psi}} \  \  \log\mathrm{det} ( \boldsymbol{\Gamma} + \sigma_n^2\mathbf{I} ) + \mathbf{y}^\top\left( \boldsymbol{\Gamma} + \sigma_n^2\mathbf{I}\right)^{-1}\mathbf{y}.
\end{equation}%
\end{remark}
With the posterior model \eqref{eq: post_f}, the desired next query point $\btheta^{(t+1)}$ maximizes the EI acquisition function 
\begin{align}
\label{eq:EI0} \btheta^{(t+1)} &=  \argmax_{ \{ \btheta \in \mathcal C \} } \mathrm{EI}( \btheta) \Def \left ( y^+ - f(\btheta) \right ) \mathcal I (f(\btheta) \leq y^+) \\
\label{eq: EI} &= \argmax_{ \{ \btheta \in \mathcal C \} }  ~  (y^+ - \mu) \Phi \left ( \frac{y^+ - \mu}{\sigma} \right ) + \sigma\phi\left ( \frac{y^+ - \mu}{\sigma} \right ) , 
\end{align}
where $y^+ = \min_{i \in [t]} f(\btheta^{(i)})$, namely, the minimum observed value, 
$\mathcal I (f(\btheta) \leq y^+) = 1$ if $f(\btheta) \leq y^+$, and $0$ otherwise (indicating that the desired next query point $\btheta$ should yield a smaller loss than the observed minimum loss), and $\mu$ \& $\sigma^2$ are defined in \eqref{eq: post_f}, $\Phi$ denotes the cumulative distribution function (CDF) of the standard normal distribution, and $\phi$ is its probability distribution function (PDF). This is true because substituting \eqref{eq: post_f} into \eqref{eq:EI0} allows us to simplify the EI acquisition function as follows:
\begin{align*}
 \mathrm{EI}( \btheta)  & \overset{f^\prime = 
    \frac{f( \btheta) - \mu}{\sigma} }{=} \mathbb E_{f^\prime}
    \left [
  ( y^+ - f^\prime \sigma - \mu ) \mathcal I \left (
  f^\prime \leq \frac{y^+ - \mu}{\sigma}
  \right )
    \right ] \\
    & = (y^+ - \mu) \Phi \left ( \frac{y^+ - \mu}{\sigma} \right ) - \sigma \mathbb E_{f^\prime}
    \left [ 
    f^\prime \mathcal I \left (
  f^\prime \leq \frac{y^+ - \mu}{\sigma}
  \right )
    \right ]  \\
    & =  (y^+ - \mu) \Phi \left ( \frac{y^+ - \mu}{\sigma} \right ) - \sigma \int_{-\infty}^{\frac{y^+ - \mu}{\sigma}} f^\prime \phi (f^\prime) d f^\prime \\
    & =  (y^+ - \mu) \Phi \left ( \frac{y^+ - \mu}{\sigma} \right ) + \sigma\phi\left (
    \frac{y^+ - \mu}{\sigma}
    \right ), 
\end{align*}
where the last equality holds since $\int x \phi(x) d x = - \phi (x) + C$ for some constant $C$. Here we omitted the constant $C$ since it does not affect the solution to the EI maximization problem \eqref{eq: EI}. With the aid of \eqref{eq: EI}, EI can be maximized via projected gradient ascent. In practice, a customized bound-constrained L-BFGS-B solver \citep{zhu1997algorithm} is often adopted.

\clearpage\pagebreak

\section{Combinatorial Multi-Armed Bandit (CMAB) for \eqref{eq: z_step} (problem \eqref{eq:int-bb-2z})} \label{asec:cmab}
\begin{algorithm}[bh]
\caption{Thompson Sampling for CMAB with probabilistic rewards}
\label{alg:CMAB0}
\begin{algorithmic}[1]
\State {\bf Input:} Beta distribution priors $\alpha_0$ and $\delta_0$, maximum iterations $L$, upper bound    $\hat{f}$ of loss $f$.
\State {\bf Set:} $n_j(k)$ and $r_j(k)$ as the cumulative counts and rewards respectively of arm $j$ pulls at bandit iteration $k$.
\For{$k \gets 1,2,\ldots,L$}
\For{all arms $j \in [K]$}
\State $\alpha_j(k) \gets \alpha_0 + r_j(k)$, $\delta_j(k) \gets \delta_0 + n_j(k) - r_j(k)$.
\State Sample $\omega_j \sim \mathrm{Beta}(\alpha_j(k), \delta_j(k))$.
\EndFor
\State Determine the arm selection scheme $\mathbf z(k)$ by solving
\begin{equation}\label{eq: bandit_sel0}
\maximize_{\bz} \sum_{i=1}^N (\bz_i)^{\top} \boldsymbol{\omega}^i 
\st \bz_i \in \{ 0, 1\}^{K_i}, \mathbf 1^{\top} \bz_i = 1, i \in [N],  
\end{equation}
\hspace*{0.2in}where $\boldsymbol{\omega} = [(\boldsymbol{\omega}^1)^{\top}, \ldots, (\boldsymbol{\omega}^N)^{\top}   ]^{\top} $ is the vector of $\{ \omega_j \}$, and $\boldsymbol{\omega}^i $ is its subvector limited to module $i$.
\State Apply strategy $\mathbf z(k)$ and observe   continuous reward $\widetilde{r}$
\begin{equation}\label{eq: reward0}
\widetilde{r} = 1 - \min \left  \{ 
\max \left \{  \frac{f(k+1)}{ \hat{f}} , 0 \right \}, 1 
\right \}
\end{equation}
    \hspace*{0.2in}where $f(k+1)$ is the loss value after applying $\mathbf z(k)$. 
\State Observe binary reward $r \sim \mathrm{Bernoulli}(\widetilde{r})$.
\For{all arms $j \in [K]$}
  \State Update $n_j(k+1) \gets n_j(k) + z_j(k) $.
  \State Update $r_j(k+1) \gets r_j(k) + z_j(k) r$.
\EndFor
\EndFor
\end{algorithmic}
\end{algorithm}
As mentioned earlier, problem \eqref{eq:int-bb-2z} can be solved as an integer program, but has two issues: (i) $\prod_{i=1}^N K_i$ black-box function queries would be needed in each ADMM iteration, and 
(ii) integer programming is difficult with the equality constraints $\sum_{j=1}^{K_i} z_{ij} = 1 \forall i \in [N]$.

We propose a customized combinatorial multi-armed bandit (CMAB) algorithm as a query-efficient alternative by interpreting problem \eqref{eq:int-bb-2d} through combinatorial bandits:
We are considering bandits due to the stochasticity in the algorithm selection arising from the fact that we train the algorithm in a subset of pipelines and not the complete combinatorially large set of all pipelines -- the basic idea is to project an optimistic upper bound on the accuracy of the full set of pipelines using Thompson sampling.
We wish to  select the optimal $N$ algorithms (arms) from $K = \sum_{i=1}^N K_i$ algorithms based on bandit feedback (`reward') $r$  inversely proportional to the loss $f$. CMAB problems can be efficiently solved with Thompson sampling (TS) \citep{durand2014thompson}. However, the conventional algorithm utilizes binary rewards, and hence is not  directly applicable to our case of continuous rewards  (with $r \propto 1 - f$ where the loss $f\in [0, 1]$ denotes the black-box objective). We address this issue by using ``probabilistic rewards'' \citep{agrawal2012analysis}.

We present the customized CMAB algorithm  in Algorithm \ref{alg:CMAB0}. The \textit{closed-form} solution of problem \eqref{eq: bandit_sel0} is given by $z_j^i = 1$ for $j = \argmax_{j \in [K_i]} \omega_j^i$, and $z_j^i = 0$ otherwise.
Step 9 of Algorithm  \ref{alg:CMAB0} normalizes the continuous loss $f$ with respect to its  upper bound $\hat f$ (assuming the lower bound is $0$), and  maps it to the continuous reward $\widetilde{r}$ within $[0,1]$.
Step 10 of Algorithm  \ref{alg:CMAB0}  converts a probabilistic reward to a binary reward. Lastly, steps 11-12 of Algorithm \ref{alg:CMAB0} update the priors of TS for combinatorial bandits  \citep{durand2014thompson}. For our experiments, we set $\alpha_0 = \delta_0 = 10$. We study the effect of $\hat{f}$ on the solution of the \eqref{eq: z_step} problem in Appendix \ref{sec:expts:flex}.

\clearpage\pagebreak

\section{ADMM with different solvers for the sub-problems} \label{sec:expts:flex}
%
%
We wish to demonstrate that our ADMM based scheme is not a single AutoML algorithm but rather a framework that can be used to mix and match different existing (and future new) black-box solvers. First we demonstrate the ability to plug in different solvers for the continuous black-box optimization involved in problem (10) (\ref{eq: theta_min} on the active set). We consider a search space containing $39$ \texttt{scikit-learn} \citep{scikit_learn} ML algorithms allowing for over $6000$ algorithm combinations. The 4 different modules and the algorithms (along with their number and types of hyper-parameters) in each of those modules is listed in Table \ref{table:search-space-hpo-subopts} in section \ref{asec:hposubopt-search-space} of the supplement. For the solvers, we consider random search (\texttt{RND}), an off-the-shelf Gaussian process based Bayesian optimization \citep{williams2006gaussian} using \texttt{scikit-optimize} (\texttt{BO}), our implementation of a Gaussian process based Bayesian optimization (\texttt{BO*})(see section \ref{asec:gpopt} in the supplement for details), and \texttt{RBFOpt} \citep{costa2018rbfopt}. We use a randomized algorithm selection scheme \eqref{eq: z_step} -- from each functional module, we randomly select an algorithm from the set of choices, and return the best combination found. The penalty parameter $\rho$ for the augmented Lagrangian term in ADMM is set $1.0$ throughout this evaluation.
%
%
%
%
\begin{figure}[htb]
    \centering
    \begin{subfigure}{0.18\textwidth}
    \includegraphics[width=\textwidth]{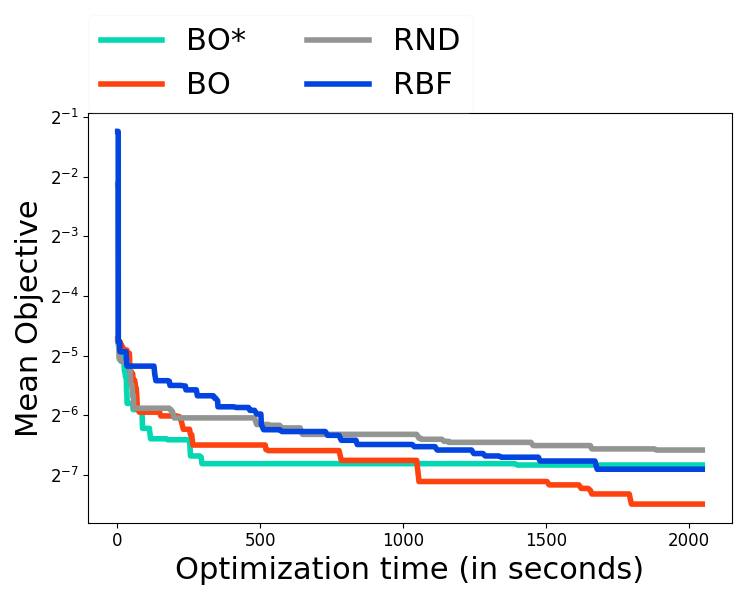}
    \caption{Oil spill}
    \end{subfigure}
    ~
    \begin{subfigure}{0.18\textwidth}
    \includegraphics[width=\textwidth]{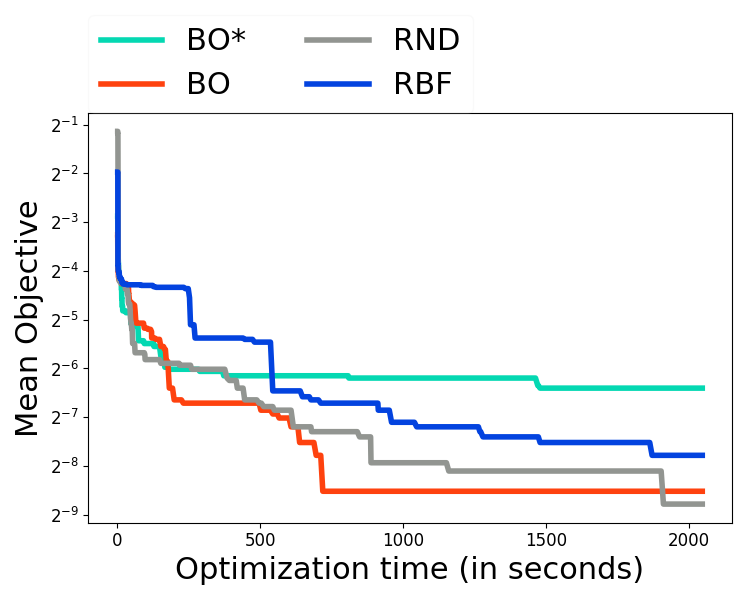}
    \caption{Sonar}
    \end{subfigure}
    ~
    \begin{subfigure}{0.18\textwidth}
    \includegraphics[width=\textwidth]{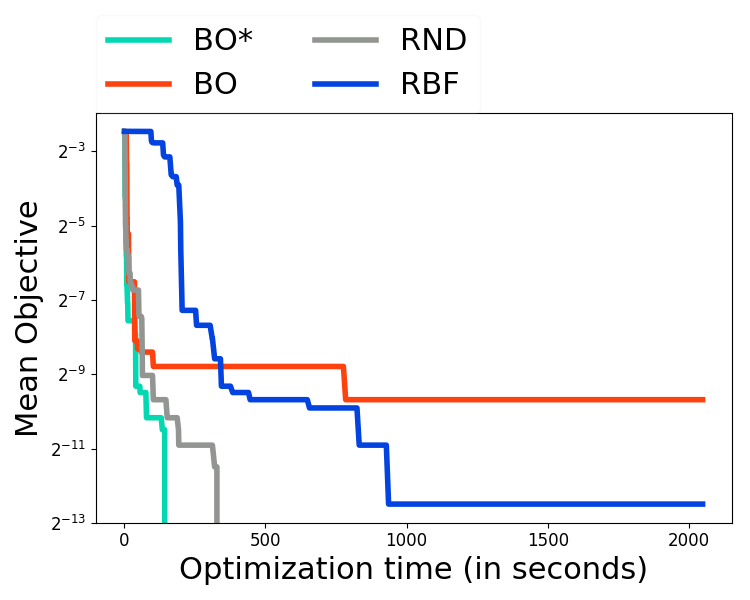}
    \caption{Ionosphere}
    \end{subfigure}
    ~ 
    \begin{subfigure}{0.18\textwidth}
    \includegraphics[width=\textwidth]{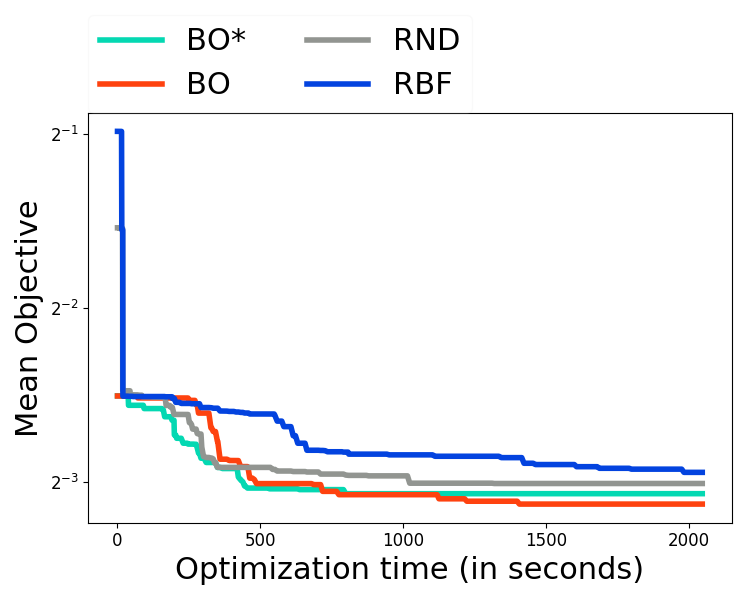}
    \caption{PC3}
    \end{subfigure}
    ~
    \begin{subfigure}{0.18\textwidth}
    \includegraphics[width=\textwidth]{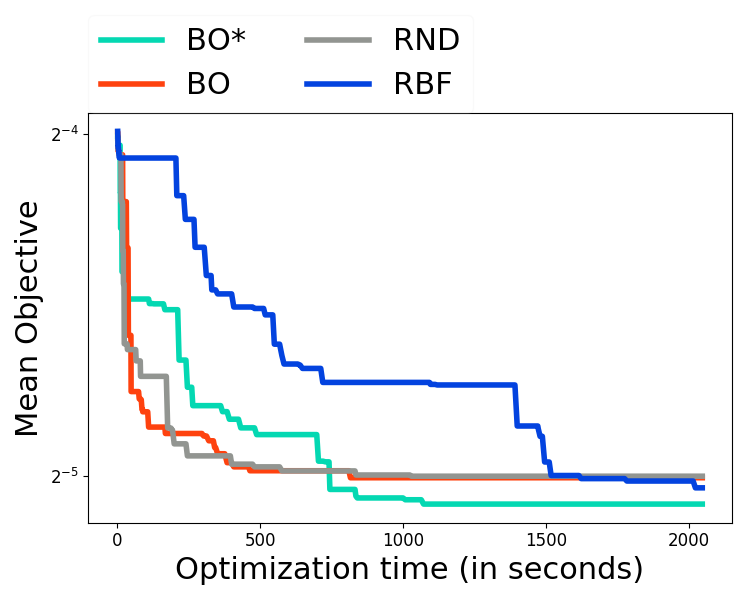}
    \caption{PC4}
    \end{subfigure}
    \caption{Average performance (across 10 runs) of different solvers for the ADMM sub-problem \eqref{eq:cont-bb-2} ({\em Please view in color}).}
    \label{fig:hposubopts}
\end{figure}
\begin{figure}[htb]
    \centering
    \begin{subfigure}{0.18\textwidth}
    \includegraphics[width=\textwidth]{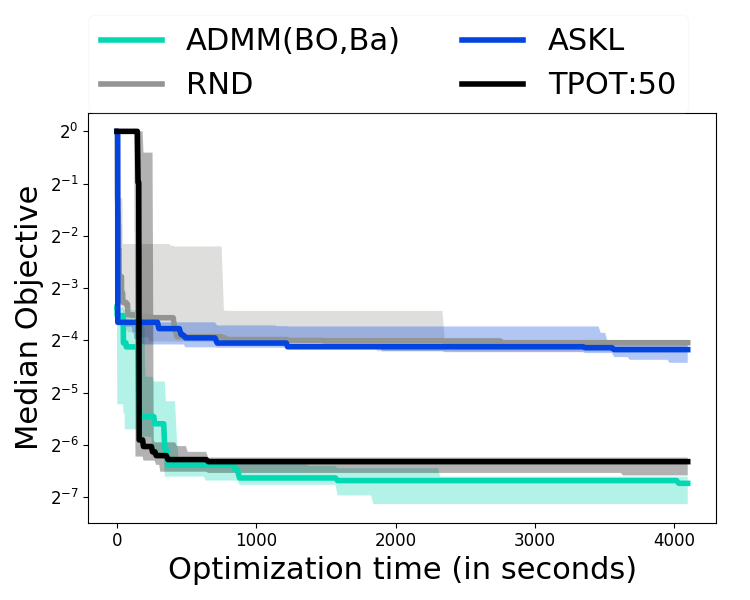}
    \caption{Oil spill}
    \end{subfigure}
    ~
    \begin{subfigure}{0.18\textwidth}
    \includegraphics[width=\textwidth]{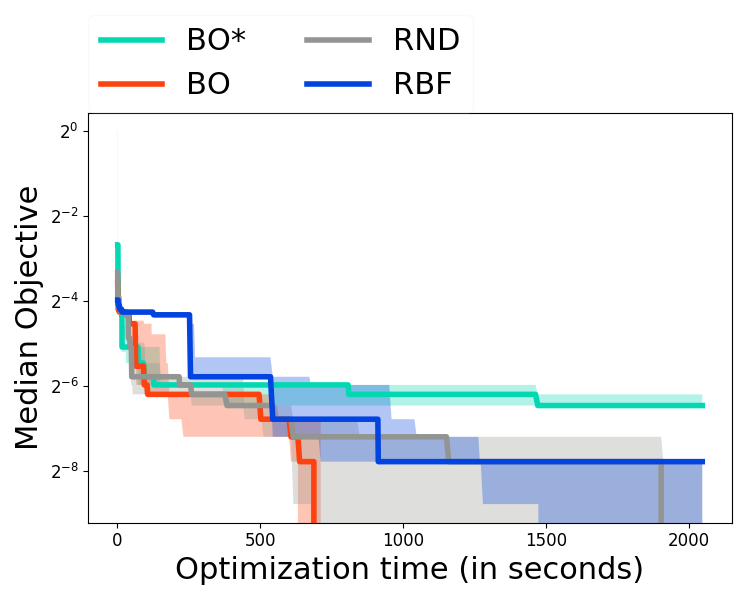}
    \caption{Sonar}
    \end{subfigure}
    ~
    \begin{subfigure}{0.18\textwidth}
    \includegraphics[width=\textwidth]{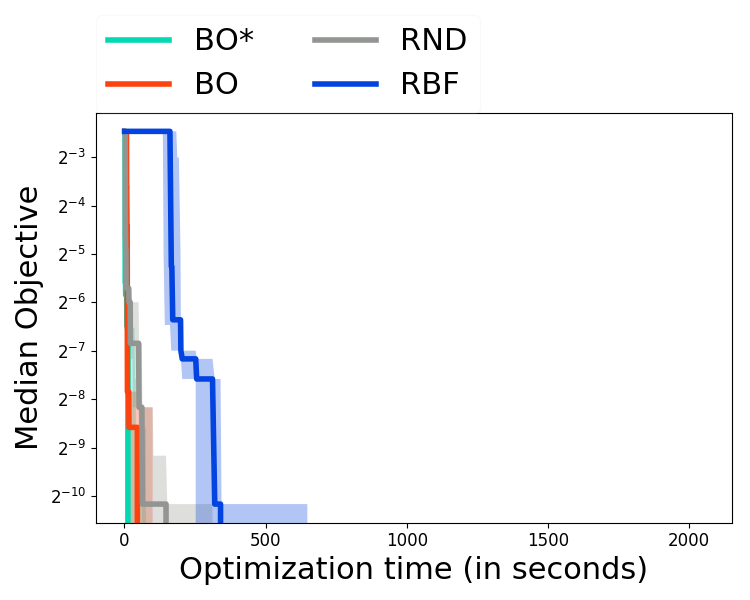}
    \caption{Ionosphere}
    \end{subfigure}
    ~ 
    \begin{subfigure}{0.18\textwidth}
    \includegraphics[width=\textwidth]{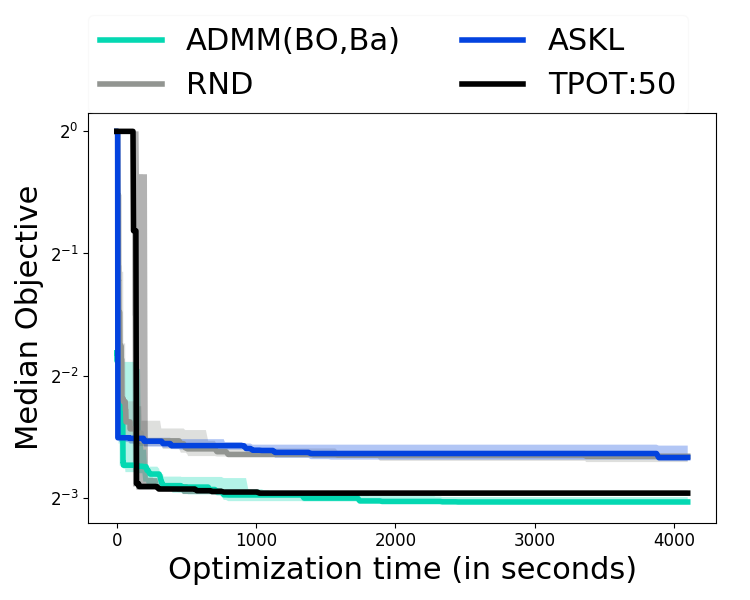}
    \caption{PC3}
    \end{subfigure}
    ~
    \begin{subfigure}{0.18\textwidth}
    \includegraphics[width=\textwidth]{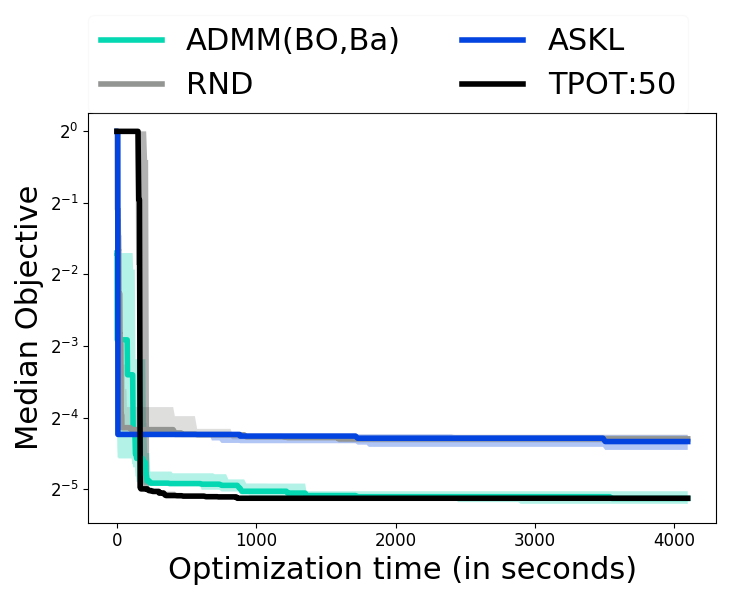}
    \caption{PC4}
    \end{subfigure}
    \caption{Performance inter-quartile range of different solvers for the ADMM sub-problem \eqref{eq:cont-bb-2} ({\em Please view in color}).}
    \label{fig:hposubopts-iqr}
\end{figure}

We present results for 5 of the datasets in the form of convergence plots showing the incumbent objective (the best objective value found till now) against the wall clock time. Here $t_{\max} = 2048, n = 128, R = 10.$ The results are presented in figures \ref{fig:hposubopts} \& \ref{fig:hposubopts-iqr}. The results indicate that the relative performance of the black-box solvers vary between data sets. However, our goal here is not to say which is best, but rather to demonstrate that our proposed ADMM based scheme is capable of utilizing any solver for the \eqref{eq: theta_min} sub-problem to search over a large space pipeline configurations.

For the algorithm selection combinatorial problem \eqref{eq: z_step}, we compare random search to a Thompson sampling \citep{durand2014thompson} based combinatorial multi-armed bandit (CMAB) algorithm. We developed a customized Thompson sampling scheme with probabilistic rewards. We detail this CMAB scheme in Appendix \ref{asec:cmab} (Algorithm \ref{alg:CMAB0}) and believe that this might be of independent interest. Our proposed CMAB scheme has two parameters: (i) the beta distribution priors $\alpha_0, \delta_0$ (set to $10$), and (ii) the loss upper bound $\hat{f}$ (which we vary as $0.3, 0.5, 0.7$).
\begin{figure}[htb]
    \centering
    \begin{subfigure}{0.18\textwidth}
    \includegraphics[width=\textwidth]{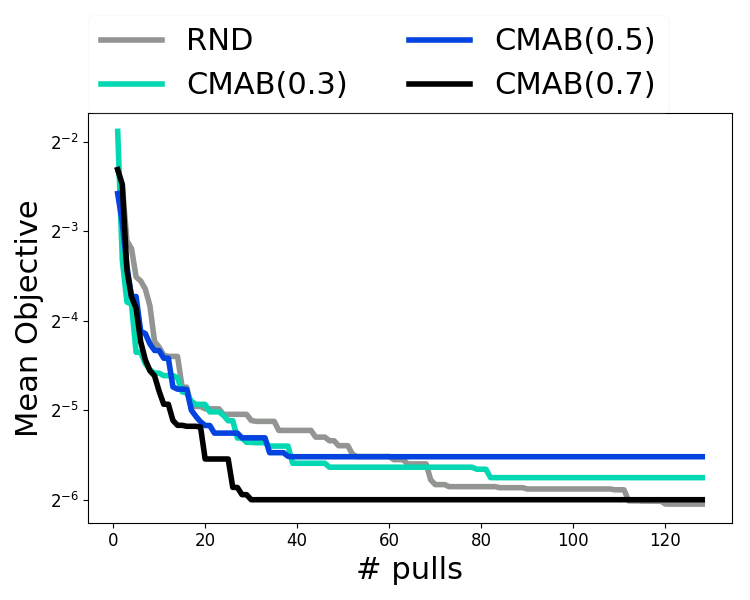}
    \caption{Oil spill}
    \end{subfigure}
    ~
    \begin{subfigure}{0.18\textwidth}
    \includegraphics[width=\textwidth]{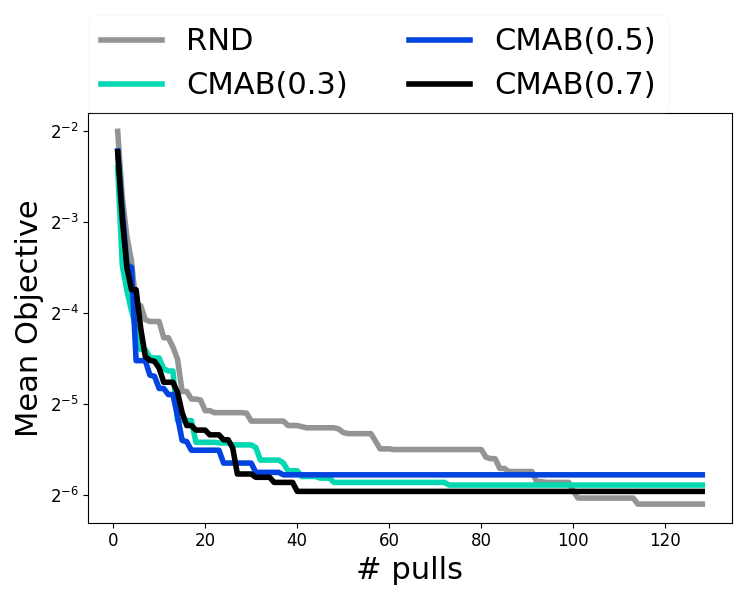}
    \caption{Sonar}
    \end{subfigure}
    ~
    \begin{subfigure}{0.18\textwidth}
    \includegraphics[width=\textwidth]{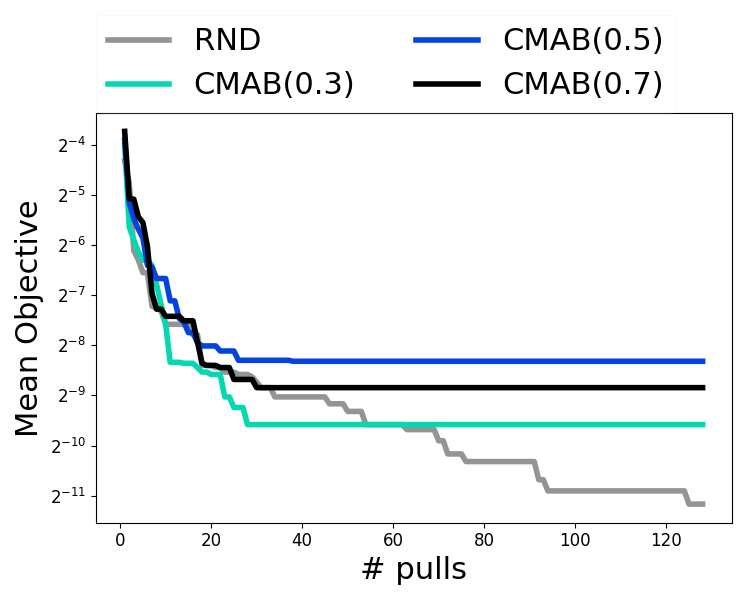}
    \caption{Ionosphere}
    \end{subfigure}
    ~ 
    \begin{subfigure}{0.18\textwidth}
    \includegraphics[width=\textwidth]{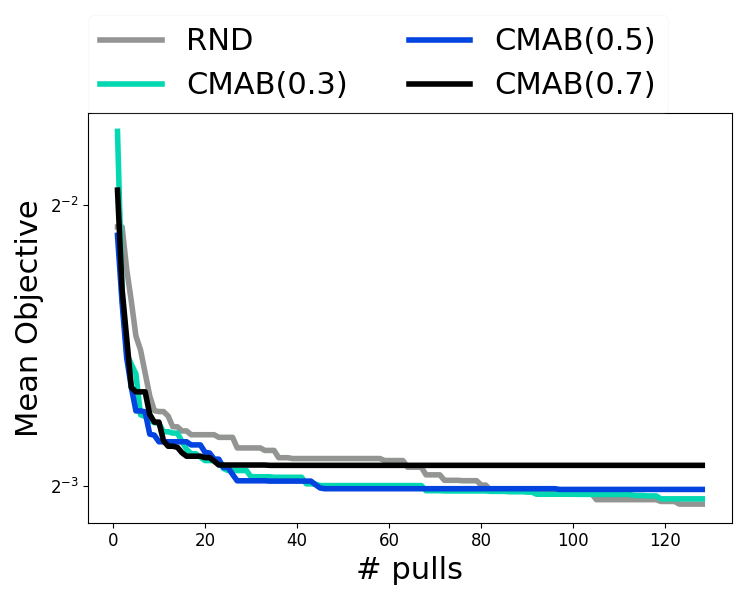}
    \caption{PC3}
    \end{subfigure}
    ~
    \begin{subfigure}{0.18\textwidth}
    \includegraphics[width=\textwidth]{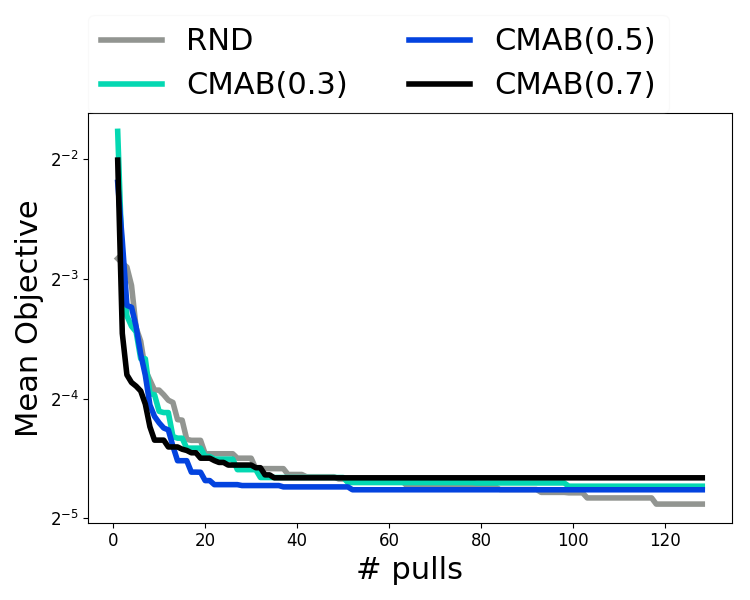}
    \caption{PC4}
    \end{subfigure}
    \caption{Average performance (across 10 runs) of different solvers for the ADMM sub-problem \eqref{eq:int-bb-2z} ({\em please view in color}).}
    \label{fig:algselsubopts}
\end{figure}
%
\begin{figure}[htb]
    \centering
    \begin{subfigure}{0.18\textwidth}
    \includegraphics[width=\textwidth]{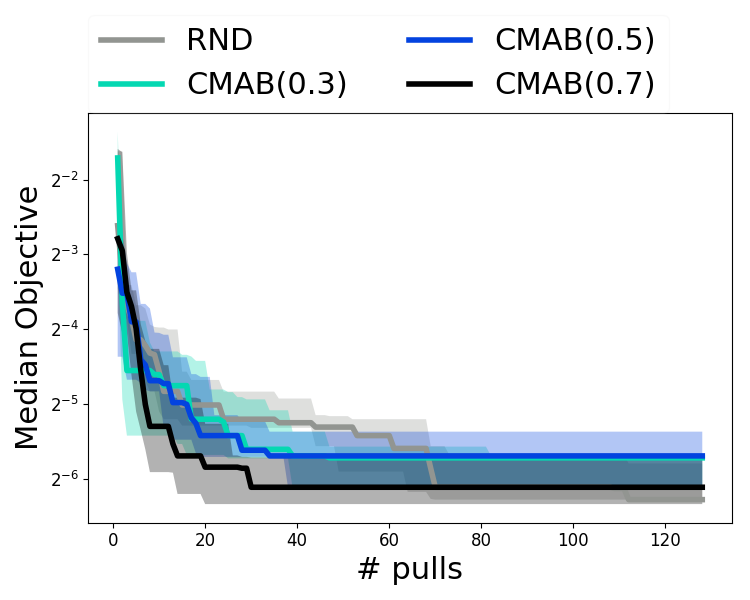}
    \caption{Oil spill}
    \end{subfigure}
    ~
    \begin{subfigure}{0.18\textwidth}
    \includegraphics[width=\textwidth]{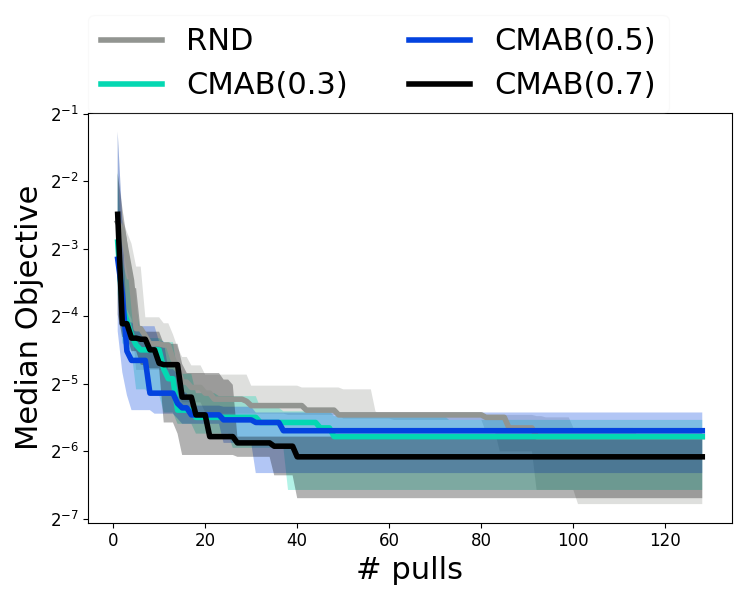}
    \caption{Sonar}
    \end{subfigure}
    ~
    \begin{subfigure}{0.18\textwidth}
    \includegraphics[width=\textwidth]{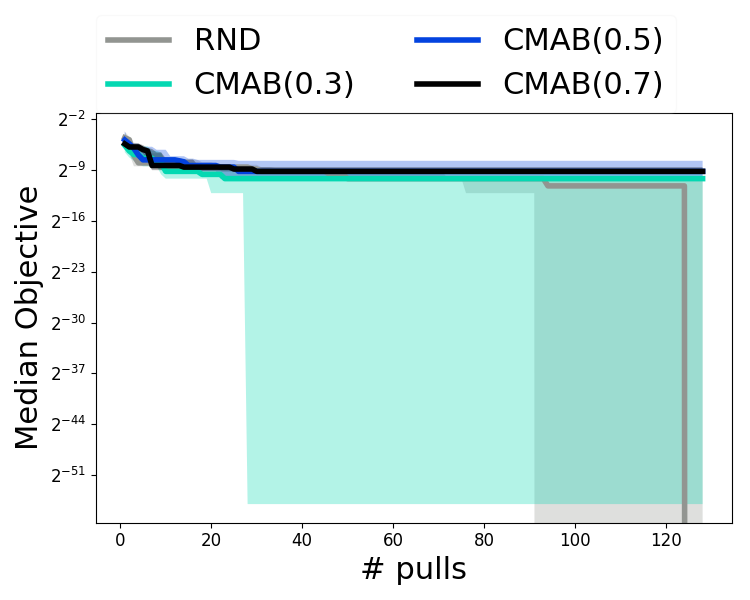}
    \caption{Ionosphere}
    \end{subfigure}
    ~ 
    \begin{subfigure}{0.18\textwidth}
    \includegraphics[width=\textwidth]{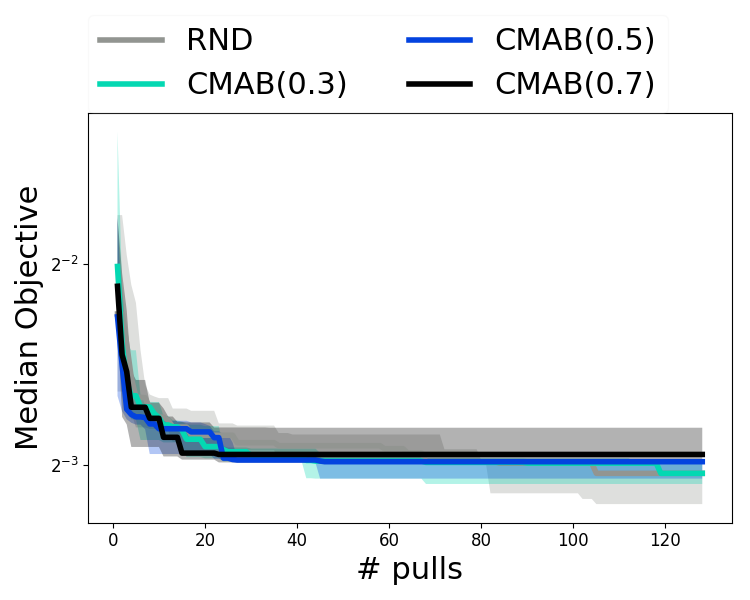}
    \caption{PC3}
    \end{subfigure}
    ~
    \begin{subfigure}{0.18\textwidth}
    \includegraphics[width=\textwidth]{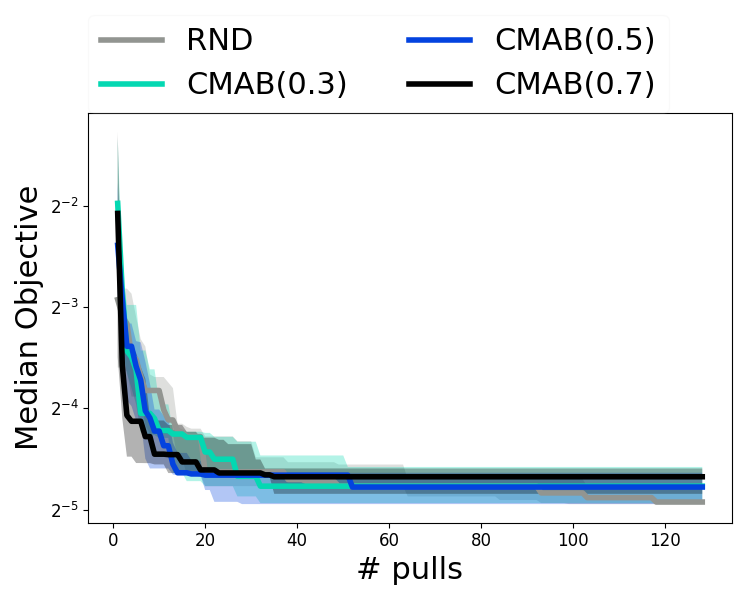}
    \caption{PC4}
    \end{subfigure}
    \caption{Performance inter-quartile range of different solvers for the ADMM sub-problem \eqref{eq:int-bb-2z} ({\em Please view in color}).}
    \label{fig:algselsubopts-iqr}
\end{figure}

We again consider results in the form of convergence plots showing the incumbent objective (the best objective value found till now) against the number of pipeline combinations tried (number of ``arms pulled'') in figures \ref{fig:algselsubopts} \& \ref{fig:algselsubopts-iqr}. The results indicate for large number of pulls, all schemes perform the same. However, on $2/5$ datasets, CMAB(0.7) (and other settings) outperforms random search for small number of pulls by a significant margin. Random search significantly outperforms CMAB on the Ionosphere dataset. The results indicate that no one method is best for all data sets, but ADMM is not tied to a single solver, and is able to leverage different solvers for the \eqref{eq: z_step} step. 

\clearpage\pagebreak

\section{Details on the data} \label{asec:data}
We consider data sets corresponding to the binary classification task from the UCI machine learning repository \citep{asuncion2007uci}, OpenML and Kaggle. The names, sizes and sources of the data sets are presented in Table \ref{tab:datasets}. There are couple of points we would like to explicitly mention here:
\begin{itemize}
    \item While we are focusing on binary classification, the proposed ADMM based scheme is applicable to any problem (such as multiclass \& multi-label classification, regression) since it is a black-box optimization scheme and can operate on any problem specific objective.
    \item We consider a subset of OpenML100 limited to binary classification and small enough to allow for meaningful amount of optimization for {\em all baselines} in the allotted 1 hour to ensure that we are evaluating the optimizers and not the initialization heuristics.
\end{itemize}
The HCDR data set from Kaggle is a subset of the data presented in the recent Home Credit Default Risk competition (\url{https://www.kaggle.com/c/home-credit-default-risk}). We selected the subset of 10000 rows and 24 features using the following steps:
\begin{itemize}
    \item We only considered the public training set since that is only set with labels available
    \item We kept all columns with keyword matches to the terms ``HOME'', ``CREDIT'', ``DEFAULT'', ``RISK'', ``AGE'', ``INCOME'', ``DAYS'', ``AMT''.
    \item In addition, we selected the top 10 columns most correlated to the labels column.
    \item For this set of features, we randomly selected 10000 rows with $\leq 4$ missing values in each rows while maintaining the original class ratio in the dataset.
\end{itemize}
\begin{table}[!!htb]
    \centering
    \caption{Details of the data sets used for the empirical evaluations. The `Class ratios' column corresponds to the ratio of the two classes in the data set, quantifying the class imbalance in the data.}
    \label{tab:datasets}
    \begin{tabular}{lccll}
        \hline
        Data & \# rows & \# columns & Source & Class ratio \\
        \hline
         Sonar & 208 & 61 & UCI & $1:0.87$ \\
         Heart statlog & 270 & 14 & UCI & $1:0.8$\\
         Ionosphere & 351 & 35 & UCI & $1:1.79$ \\
         Oil spill & 937 & 50 & OpenML & $1:0.05$ \\
         fri-c2 & 1000 & 11 & OpenML & $1:0.72$ \\
         PC3 & 1563 & 38 & OpenML & $1:0.11$ \\
         PC4 & 1458 & 38 & OpenML & $1:0.14$ \\
         Space-GA & 3107 & 7 & OpenML & $1:0.98$ \\
         Pollen & 3848 & 6 & OpenML  & $1:1$ \\
         Ada-agnostic & 4562 & 48 & OpenML & $ 1:0.33$ \\
         Sylvine & 5124 & 21 & OpenML & $1:1$\\
         Page-blocks & 5473 & 11 & OpenML & $1:8.77$ \\
         Optdigits & 5620 & 64 & UCI & $ 1 : 0.11$ \\
         Wind & 6574 & 15 & OpenML & $1:1.14$ \\
         Delta-Ailerons & 7129 & 6 & OpenML & $1:1.13$ \\
         Ringnorm & 7400 & 21 & OpenML & $1:1.02$\\
         Twonorm & 7400 & 21 & OpenML & $1:1$\\
         Bank8FM & 8192 & 9 & OpenML & $1:1.48$\\
         Puma8NH & 8192 & 9 & OpenML & $1:1.01$ \\ 
         CPU small & 8192 & 13 & OpenML & $1:0.43$ \\
         Delta-Elevators & 9517 & 7 & OpenML & $1:0.99$ \\
         Japanese Vowels & 9961 & 13 & OpenML & $1:0.19$ \\
         HCDR & 10000 & 24 & Kaggle  & $1:0.07$ \\
         Phishing websites & 11055 & 31 & UCI & $1:1.26$ \\
         Mammography & 11183 & 7 & OpenML & $1:0.02$ \\
         EEG-eye-state & 14980 & 15 & OpenML & $1:0.81$ \\
         Elevators & 16598 & 19 & OpenML & $1:2.24$ \\
         Cal housing & 20640 & 9 & OpenML & $1:1.46$ \\
         MLSS 2017 CH\#2 & 39948 & 12 & OpenML & $1:0.2$ \\
         2D planes & 40768 & 11 & OpenML & $1:1$ \\
         Electricity & 45312 & 9 & OpenML & $1:0.74$\\
        \hline
    \end{tabular}
\end{table}

\clearpage\pagebreak

\section{Search space: Algorithm choices and hyper-parameters}\label{asec:search-space}
In this section, we list the different search spaces we consider for the different empirical evaluations in section \ref{sec:expts} of the paper. 
\subsection{Larger search space} \label{asec:hposubopt-search-space}
For the empirical evaluation of black-box constraints (section \ref{sec:expts} (ii)), ADMM flexibity (section \ref{sec:expts} (iii)) and Appendix \ref{sec:expts:flex}, 
%
we consider $5$ functional modules -- feature preprocessors, feature scalers, feature transformers, feature selectors, and finally estimators. The missing handling and the categorical handling is always applied if needed. For the rest of the modules, there are $8$, $11$, $7$ and $11$ algorithm choices respectively, allowing for $6776$ possible pipeline combinations. We consider a total of $92$ hyperparamters across all algorithms. The algorithm hyper-parameter ranges are set using Auto-sklearn as the reference ( see \url{https://github.com/automl/auto-sklearn/tree/master/autosklearn/pipeline/components}).
\begin{table}[!hhhhh]
\centering
\caption{Overview of the scikit-learn feature preprocessors, feature transformers, feature selectors and estimators used in our empirical evaluation. The preprocessing is always applied so there is no choice there. Barring that, we are searching over a total of $8 \times 11 \times 7 \times 11 = 6776$ possible pipeline compositions.}
\label{table:search-space-hpo-subopts}
\begin{adjustbox}{max width=0.7\textwidth }
\begin{threeparttable}
\begin{tabular}{c|c|c}
\hline
\begin{tabular}[c]{@{}c@{}}Module\end{tabular}     & \begin{tabular}[c]{@{}c@{}}Algorithm  \end{tabular}      & \begin{tabular}[c]{@{}c@{}}\# parameters\end{tabular}      
\\ \hline  
{Preprocessors} 
  & \begin{tabular}{c} Imputer \\ OneHotEncoder \end{tabular} 
  &  \begin{tabular}{c}  1d \\  none  \end{tabular} 
\\ \hline
{Scalers $\times 8$} 
  & \begin{tabular}{c} 
  None$^*$  \\ Normalizer  \\  QuantileTransformer \\ MinMaxScaler \\
  StandardScaler \\ RobustScaler \\ Binarizer \\ KBinsDiscretizer
  \end{tabular} 
  &  \begin{tabular}{c}
  none \\   none \\ 2d$^\dag$ \\ none \\ none \\ 2c$^\dag$, 2d \\ 2d
  \end{tabular}
\\ \hline
{Transformer $\times 11$}   
  &  \begin{tabular}{c} 
  None \\ SparseRandomProjection \\ GaussianRandomProjection \\ RBFSampler \\
  Nystroem \\ TruncatedSVD \\ KernelPCA \\ FastICA \\ FactorAnalysis \\ PCA \\
  PolynomialFeatures
  \end{tabular} 
  & \begin{tabular}{c} 
  none \\ 1c, 1d \\ 1d \\ 1c, 1d \\ 2c, 3d \\ 2d \\ 2c, 4d \\ 5d \\ 3d \\ 1c, 1d \\ 3d
  \end{tabular}
\\ \hline
{Selector $\times 7$}   
  &  \begin{tabular}{c} 
  None \\ SelectPercentile \\ SelectFpr \\ SelectFdr \\ SelectFwe \\ VarianceThreshold 
  \end{tabular} 
  & \begin{tabular}{c} 
  none \\ 1d \\ 1c \\ 1c \\ 1c \\ 1c 
  \end{tabular}
\\ \hline
{Estimator $\times 11$}  
  & \begin{tabular}{c}
  GaussianNB  \\  QuadraticDiscriminantAnalysis \\ GradientBoostingClassifier \\ 
  KNeighborsClassifier \\ RandomForestClassifier \\ ExtraTreesClassifier \\
  AdaBoostClassifier \\ DecisionTreeClassifier \\ GaussianProcessClassifier \\
  LogisticRegression \\ MLPClassifier 
  \end{tabular} 
  &  \begin{tabular}{c} 
  none  \\ 1c \\ 3c, 6d \\  3d \\ 1c, 5d \\ 1c, 5d \\ 1c, 2d \\ 3c, 3d \\ 2d \\ 
  2c, 3d \\ 2c, 5d
\end{tabular}
\\ \hline
\end{tabular}
\begin{tablenotes}
{\small  $^*$None means no algorithm is selected and corresponds to a empty set of hyper-parameters. $^\dag$ `d' and `c' represents discrete and continuous variables, respectively.}%
\end{tablenotes}
\end{threeparttable}
\end{adjustbox}
\end{table}
\subsection{Smaller search space for comparing to AutoML baselines} \label{asec:baselines-search-space}
%
We choose a relatively smaller search space in order to keep an efficient fair comparison across all baselines, auto-sklearn, TPOT and ADMM, {\bf with the same set of operators}, including all imputation and rescaling. However, there is a technical issue  -- many of the operators in Auto-sklearn are custom preprocessors and estimators (kitchen sinks, extra trees classifier preprocessor, linear svc preprocessors, fastICA, KernelPCA, etc) or have some custom handling in there (see \url{https://github.com/automl/auto-sklearn/tree/master/autosklearn/pipeline/components}). Inclusion of these operators makes it infeasible to have a fair comparison across all methods.
Hence, we consider a reduced search space, detailed in Table \ref{table:search-space-automl}. It represents $4$ functional modules with a choice of $6 \times 3 \times 6 = 108$ possible method combinations (contrast to Table \ref{table:search-space-hpo-subopts}). 
For each scheme, the algorithm hyper-parameter ranges are set using Auto-sklearn as the reference (see \url{https://github.com/automl/auto-sklearn/tree/master/autosklearn/pipeline/components}).
\begin{table}[!hhhhh]
\centering
\caption{Overview of the scikit-learn preprocessors, transformers, and estimators used in our empirical evaluation comparing ADMM, auto-sklearn, TPOT. We consider a choice of $6 \times 3 \times 6 = 108$ possible method combinations (see text for further details).}
\label{table:search-space-automl}
\begin{adjustbox}{max width=0.7\textwidth }
\begin{threeparttable}
\begin{tabular}{c|c|c}
\hline
\begin{tabular}[c]{@{}c@{}}Module\end{tabular}     & \begin{tabular}[c]{@{}c@{}}Algorithm  \end{tabular}      & \begin{tabular}[c]{@{}c@{}}\# parameters\end{tabular}      
 
\\ \hline  
{Preprocessors} 
  & \begin{tabular}{c} Imputer \\ OneHotEncoder \end{tabular} 
  &  \begin{tabular}{c}  1d \\  none  \end{tabular} 
\\ \hline
{Scalers $\times 6$} 
  & \begin{tabular}{c} 
  None$^*$  \\ Normalizer  \\  QuantileTransformer \\ MinMaxScaler \\
  StandardScaler \\ RobustScaler 
  \end{tabular} 
  &  \begin{tabular}{c}
  none \\   none \\ 2d$^\dag$ \\ none \\ none \\ 2c$^\dag$, 2d \\ 2d
  \end{tabular}
\\ \hline
{Transformer $\times 3$}   &  \begin{tabular}{c} 
None \\ 
PCA \\ 
PolynomialFeatures \\
\end{tabular} & \begin{tabular}{c} 
none \\ 
1c, 1d\\  
1c, 2d \\
  \end{tabular}
\\ \hline
{Estimator $\times 6$}  & \begin{tabular}{c}
GaussianNB  \\  
QuadraticDiscriminantAnalysis \\
GradientBoostingClassifier \\ 
KNeighborsClassifier \\
RandomForestClassifier \\
ExtraTreesClassifier
\end{tabular} &  \begin{tabular}{c} 
none  \\
1c \\
3c, 6d\\   
3d \\ 
1c, 5d \\
1c, 5d
\end{tabular}
\\ \hline
\end{tabular}
\begin{tablenotes}
{\small  $^*$None means no algorithm is selected and corresponds to a empty set of hyper-parameters. $^\dag$ `d' and `c' represents discrete and continuous variables, respectively.}%
\end{tablenotes}
\end{threeparttable}
\end{adjustbox}
\end{table}
\paragraph{Note on parity between baselines.}
With a {\em fixed pipeline shape and order}, ADMM \& ASKL are optimizing over the same search space by making a single selection from each of the functional modules to generate a pipeline. In contrast, TPOT can use multiple methods from the same functional module within a single pipeline with {\em stacking} and {\em chaining} due to the nature of the splicing/crossover schemes in its underlying genetic algorithm. This gives TPOT {\em access to a larger search space of more complex pipelines featuring longer as well as parallel compositions}, rendering the comparison {\bf somewhat biased towards} TPOT. Notwithstanding this caveat, we consider TPOT as a baseline since it is a competitive open source AutoML alternative to ASKL, and is representative of the genetic programming based schemes for AutoML. We provide some examples of the complex pipelines found by TPOT in Appendix \ref{asec:TPOT-exs}.

\clearpage\pagebreak

\section{Learning ensembles with ADMM} \label{asec:ensembles}
We use the greedy selection based ensemble learning scheme proposed in \citet{caruana2004ensemble} and used in Auto-sklearn as a post-processing step \citep{feurer2015efficient}. We run ASKL and ADMM(BO, Ba) for $t_{\max}=300$ seconds and then utilize the following procedure to compare the ensemble learning capabilities of Auto-sklearn and our proposed ADMM based optimizer:
    \begin{itemize}
        \item We consider different ensemble sizes $e_1 = 1 < e_2 = 2 < e_3 = 4 \ldots < e_{\max} = 32$.
        \item We perform {\em library pruning} on the pipelines found during the optimization run for a maximum search time $t_{\max}$ by picking only the $e_{\max}$ best models (best relative to their validation score found during the optimization phase).
        \item Starting with the pipeline with the best $\hat{s}$ as the first member of the ensemble, for each ensemble size $e_j$, we greedily add the pipeline (with replacement) which results in the best performing bagged ensemble (best relative to the performance $\hat{s}'_j$ on the validation set $S_v$ after being trained on the training set $S_t$).
        \item Once the ensemble members (possibly with repetitions) are chosen for any ensemble size $e_j$, the ensemble members are retrained on the whole training set (the training + validation set) and the bagged ensemble is then evaluated on the unseen held-out test set $S_h$ to get $s'_j$. We follow this procedure since the ensemble learning uses the validation set and hence cannot be used to generate a fair estimate of the generalization performance of the ensemble.
        \item Plot the $(e_j, s'_j)$ pairs. 
        \item The whole process is repeated $R = 10$ times for the same $T$ and $e_j$s to get error bars for $s'_j$.
    \end{itemize}
For ADMM(BO,Ba), we implement the \citet{caruana2004ensemble} scheme ourselves. For ASKL:SMAC3, we use the post-processing ensemble-learning based on the example presented in their documentation at \url{https://automl.github.io/auto-sklearn/master/examples/example_sequential.html}.
\begin{figure*}[hb]
    \centering
    \begin{subfigure}{0.23\textwidth}
    \includegraphics[width=\textwidth]{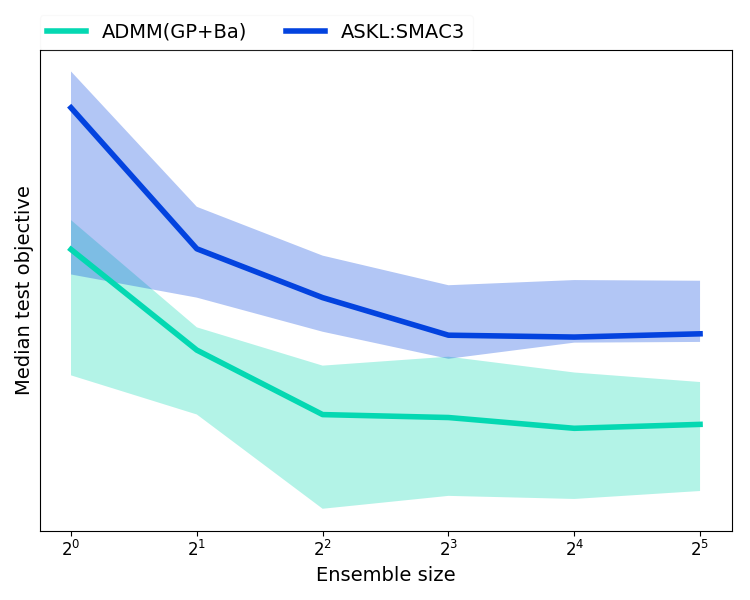}
    \caption{Bank8FM}
    \end{subfigure}
    ~ 
    \begin{subfigure}{0.23\textwidth}
    \includegraphics[width=\textwidth]{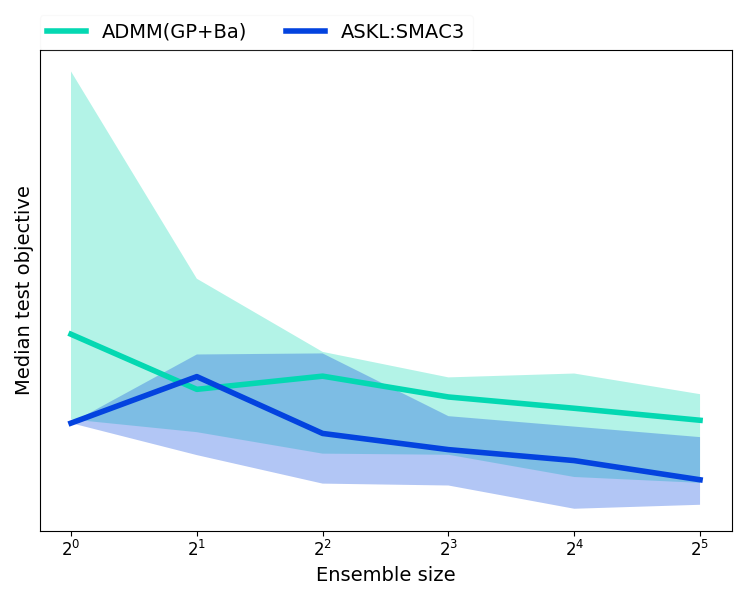}
    \caption{CPU small}
    \end{subfigure}
    ~ 
    \begin{subfigure}{0.23\textwidth}
    \includegraphics[width=\textwidth]{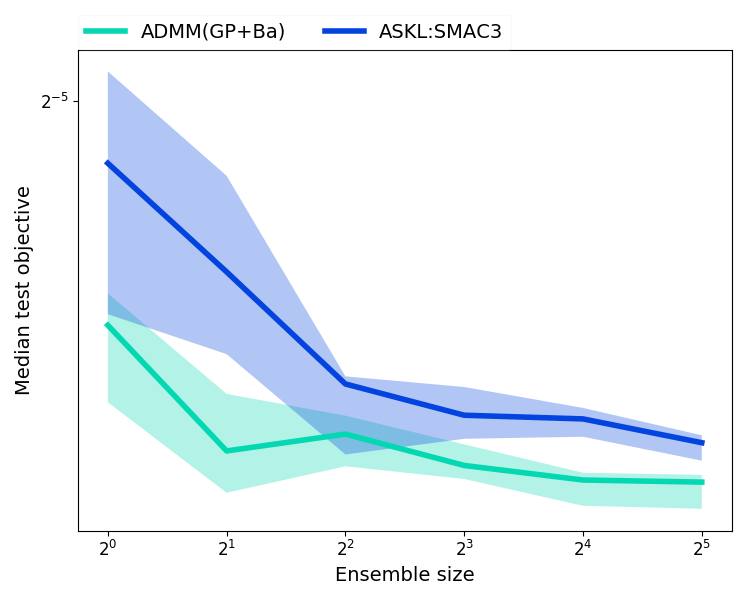}
    \caption{Delta Ailerons}
    \end{subfigure}
    ~ 
    \begin{subfigure}{0.23\textwidth}
    \includegraphics[width=\textwidth]{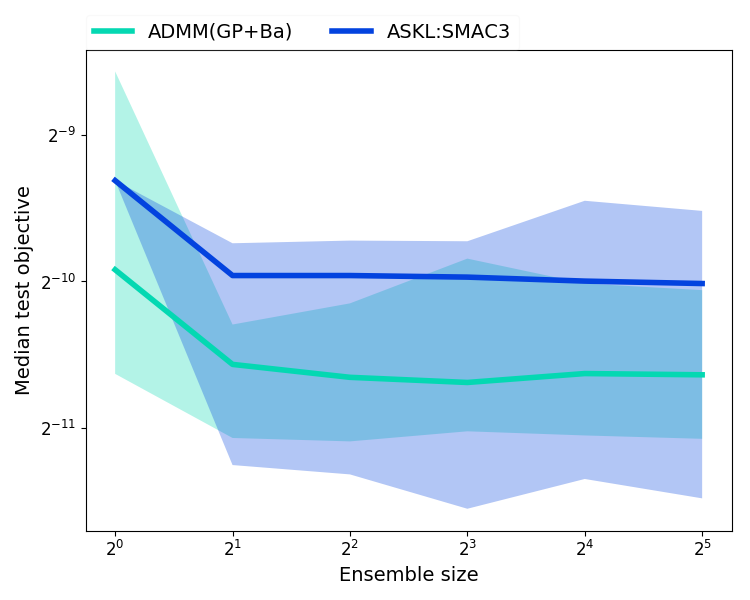}
    \caption{Japanese Vowels}
    \end{subfigure}
    ~ 
    \begin{subfigure}{0.23\textwidth}
    \includegraphics[width=\textwidth]{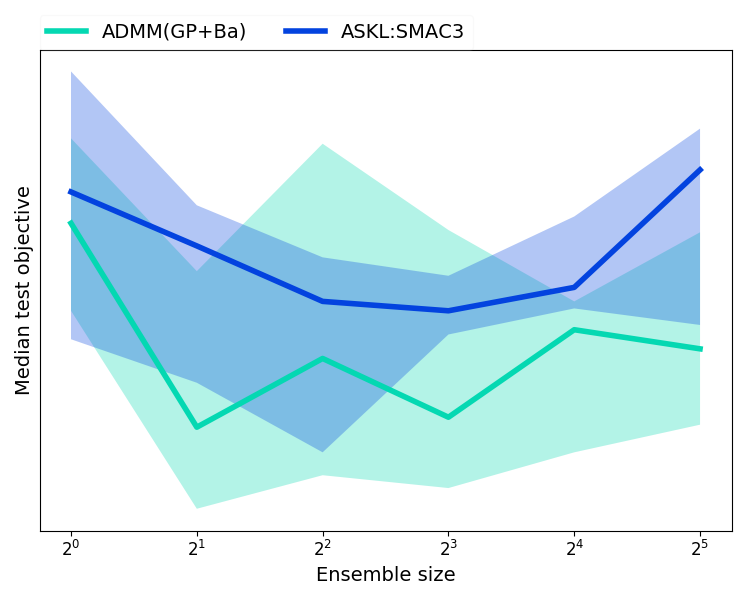}
    \caption{Page blocks}
    \end{subfigure}
    ~ 
    \begin{subfigure}{0.23\textwidth}
    \includegraphics[width=\textwidth]{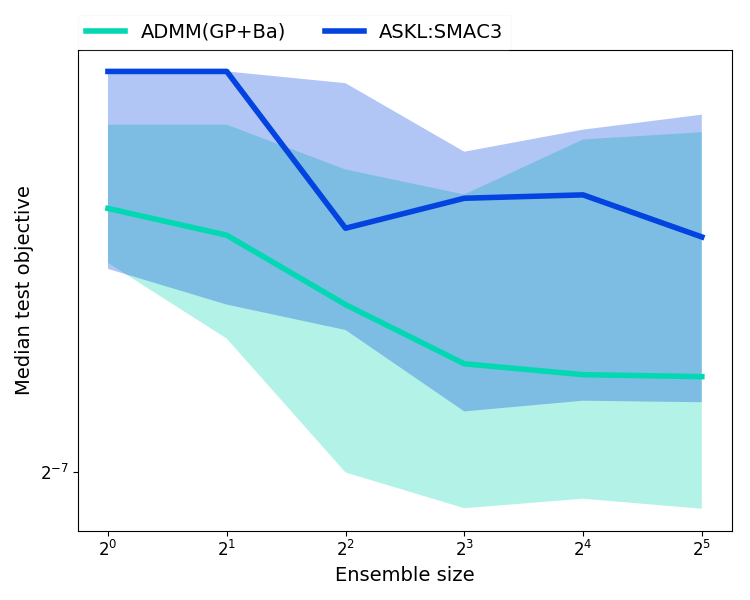}
    \caption{Sylvine}
    \end{subfigure}
    ~ 
    \begin{subfigure}{0.23\textwidth}
    \includegraphics[width=\textwidth]{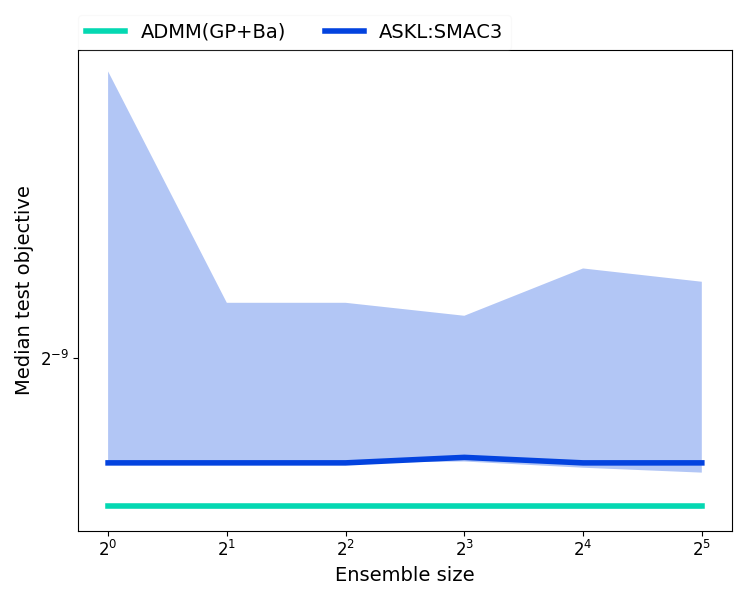}
    \caption{Twonorm}
    \end{subfigure}
    ~ 
    \begin{subfigure}{0.23\textwidth}
    \includegraphics[width=\textwidth]{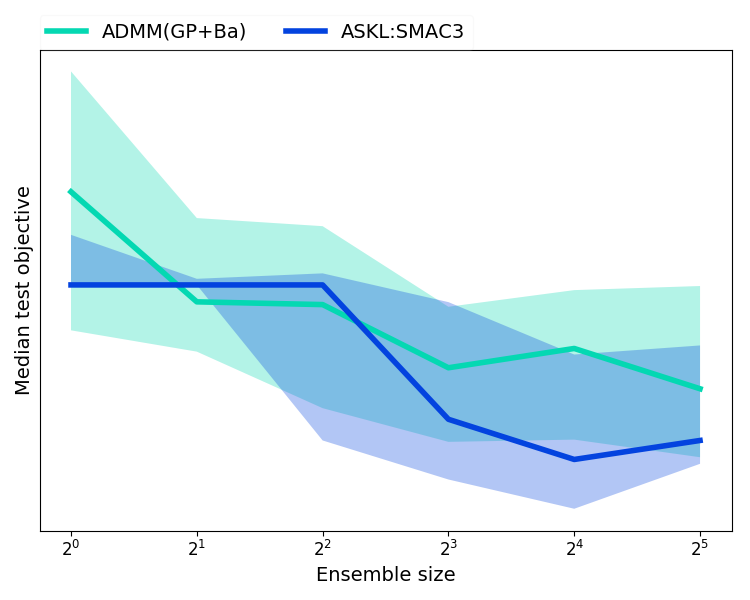}
    \caption{Wind}
    \end{subfigure}
    \caption{Ensemble size vs. median performance on the test set and the inter-quartile range ({\em please view in color}). The \textcolor{aquamarine}{Aquamarine} and \textcolor{blue}{Blue} curves correspond to ADMM(BO,Ba) and ASKL respectively.}
    \label{fig:esize-v-testperf-iqr}
\end{figure*}

The inter-quartile range (over 10 trials) of the test performance of the post-processing ensemble learning for a subset of the data sets in Table \ref{tab:datasets} is presented in Figure \ref{fig:esize-v-testperf-iqr}. The results indicate that the ensemble learning with ADMM is able to improve the performance similar to the ensemble learning in Auto-sklearn. The overall performance is driven by the starting point (the test error of the best single pipeline, corresponding to an ensemble of size 1) -- if ADMM and Auto-sklearn have test objective values that are close to each other (for example, in Page-blocks and Wind), their performance with increasing ensemble sizes are very similar as well.

\clearpage\pagebreak

\section{Parameter sensitivity check for ADMM}\label{asec: para_sensitivity} 
\begin{figure}[!!!htb]
\centering
\includegraphics[width=0.6\textwidth]{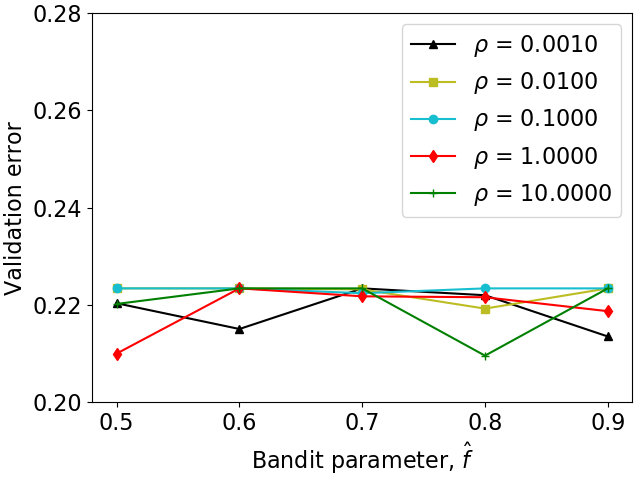}
\caption{Validation error of our proposed ADMM-based approach against ADMM parameter $\rho$ and CMAB parameter $\hat{f}$}
\label{fig: para_check}
\end{figure}

We investigate how sensitive our proposed approach is to the ADMM parameter $\rho$ and CMAB parameter $\hat{f}$. For each parameter combination of $\rho \in \{0.001, 0.01, 0.1, 1, 10 \}$ and $\hat{f} \in \{ 0.5, 0.6, 0.7, 0.8, 0.9\}$, in Figure \ref{fig: para_check} we present the validation error (averaged over $10$ trials) by running our approach 
on the HCDR dataset (see Appendix \ref{asec:data}). 
%
%

For this experiment, the results indicate that a large $\rho$ yields a slightly better performance. However, in general, our approach is not very sensitive to the choice of $\rho$ and $\hat{f}$ -- the range of the objectives achieved are in a very small range. Based on this observation, we set $\rho = 1$ and $\hat{f} = 0.7$ in all our empirical evaluations of ADMM(BO,Ba) unless otherwise specified.

\clearpage\pagebreak

\section{Details on the baselines and evaluation scheme} \label{asec:expt-details}
\paragraph{Evaluation scheme.} The optimization is run for some maximum runtime $T$ where each proposed configuration is trained on a set $S_t$ and evaluated on $S_v$ and the obtained score $\hat{s}$ is the objective that is being minimized by the optimizer. We ensure that all the optimizers use the same train-validation split. Once the search is over, the history of attempted configurations is used to generate a search time vs. holdout performance curve in the following manner for $N$ timestamps:
\begin{itemize}
\item For each timestamp $t_i, i = 1, \ldots, N, t_N = T$, we pick the best validation score $\hat{s}_i$ obtained by any configuration found by time $t_i$ from the start of the optimization (the incumbent best objective). 
\item Then we plot the $(t_i, \hat{s}_i)$ pairs. 
\item The whole above process is repeated $R$ times for the same $T, N$ and $t_i$s to get inter-quartile ranges for the curves.
\end{itemize}
For the presented results, $T=3600$ seconds, $N = 256$ and $R = 10$.

\paragraph{Parity with baselines.} First we ensure that the operations (such as model training) are done single-threaded (to the extent possible) to remove the effects of parallelism in the execution time. We set \texttt{OPENBLAS\_NUM\_THREADS} and \texttt{OMP\_NUM\_THREADS} to $1$ before the evaluation of ADMM and the other baselines. ADMM can take advantage of the parallel model-training much like the other systems, but we want to demonstrate the optimization capability of the proposed scheme independent of the underlying parallelization in model training. Beyond this, there are some details we note here regarding comparison of methods based on their internal implementation:
\begin{itemize}
    \item For any time $t_i$, if no predictive performance score (the objective being minimized) is available, we give that method the worst objective of $1.0$ for ranking (and plotting purposes). After the first score is available, all following time stamps report the best incumbent objective. So comparing the different baselines at the beginning of the optimization does not really give a good view of the relative optimization capabilities -- it just illustrates the effect of different starting heuristics.
    \item For ADMM, the first pipeline tried is Naive Bayes, which is why ADMM always has some reasonable solution even at the earliest timestamp.
    \item The per configuration run time and memory limits in Auto-sklearn are removed to allow Auto-sklearn to have access to the same search space as the ADMM variants.
    \item The ensembling and meta-learning capabilities of Auto-sklearn are disabled. The ensembling capability of Auto-sklearn is discussed further in Appendix \ref{asec:ensembles}.
    \item For ASKL, the first pipeline tried appears to be a Random Forest with 100 trees, which takes a while to be run. For this reason, there is no score (or an objective of $1.0$) for ASKL until its objective suddenly drops to a more competitive level since Random Forests are very competitive out of the box.
    \item For TPOT, the way the software is set up (to the best of our understanding and trials), scores are only available at the end of any generation of the genetic algorithm. Hence, as with ASKL, TPOT do not report any scores until the first generation is complete (which implies worst-case objective of $1.0$), and after that, the objective drops significantly. For the time limit considered ($T = 3600$ seconds), the default population size of 100 set in TPOT is unable to complete a multiple generations on most of the datasets. So we reduce the population size to 50 to complete a reasonable number of generations within the set time.
    \item As we have discussed earlier, TPOT has an advantage over ASKL and ADMM -- TPOT is allowed to use multiple estimators, transformers and preprocessors within a single pipeline via stacking and chaining due to the nature of the splicing and crossover schemes in its underlying genetic algorithm. This gives TPOT access to a larger search space of more complex pipelines featuring longer as well as parallel compositions; all the remaining baselines are allowed to only use a single estimator, transformers and preprocessor. Hence the comparison is somewhat biased towards TPOT, allowing TPOT to potentially find a better objective in our experimental set up. If TPOT is able to execute a significant number of generations, we have observed in many cases that TPOT is able to take advantage of this larger search space and produce the best performance.
    \item Barring the number of generations (which is guided by the maximum run time) and the population size (which is set to 50 to give us TPOT50), the remaining parameters of mutation rate, crossover rate, subsample fraction and number of parallel threads to the default values of 0.9, 0.1, 1.0 and 1 respectively.
\end{itemize}

Random search (RND) is implemented based on the Auto-sklearn example for random search at \url{https://automl.github.io/auto-sklearn/master/examples/example_random_search.html}.

\paragraph{Compute machines.} All evaluations were run single-threaded on a 8 core 8GB CentOS virtual machines.

\clearpage\pagebreak

%
%
\section{Convergence plots for all data sets for all AutoML baselines.} \label{asec:cvgplot}
\begin{figure}[!!!!h]
    \centering
    \begin{subfigure}{0.185\textwidth}
    \includegraphics[width=\textwidth]{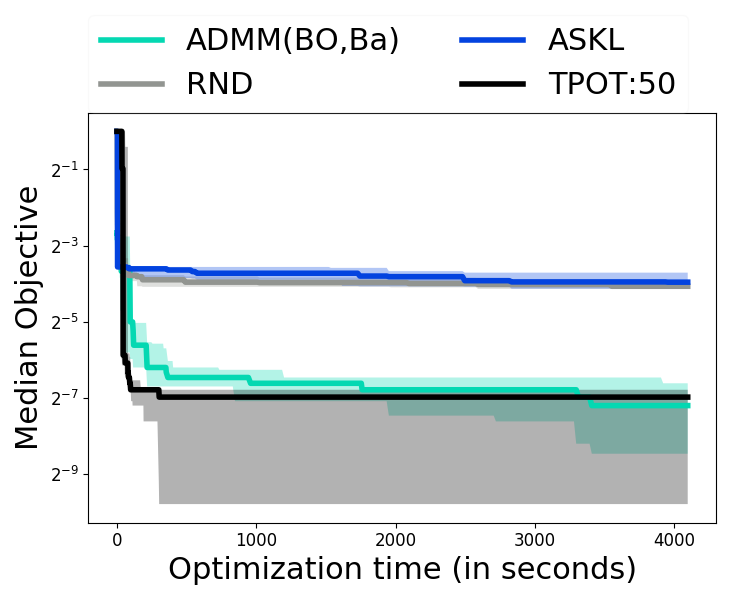}
    \caption{Sonar}
    \end{subfigure}
    ~ 
    \begin{subfigure}{0.185\textwidth}
    \includegraphics[width=\textwidth]{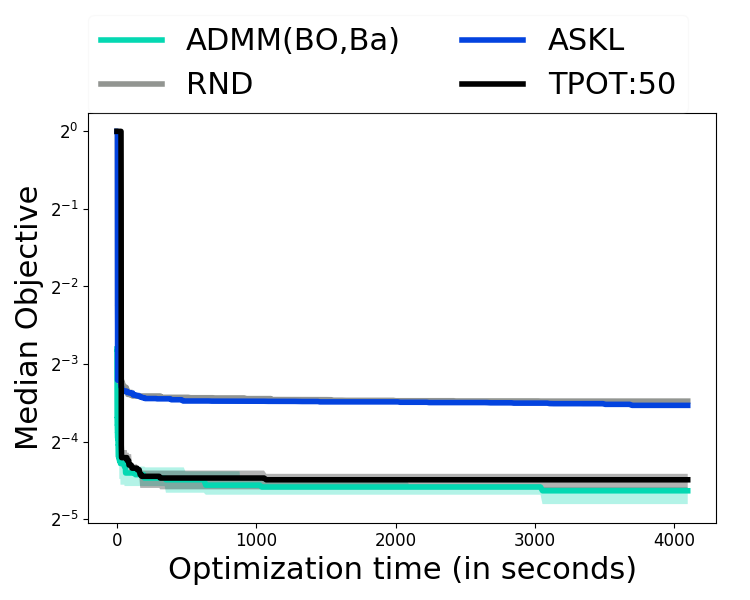}
    \caption{Heart-Statlog}
    \end{subfigure}
    ~ 
    \begin{subfigure}{0.185\textwidth}
    \includegraphics[width=\textwidth]{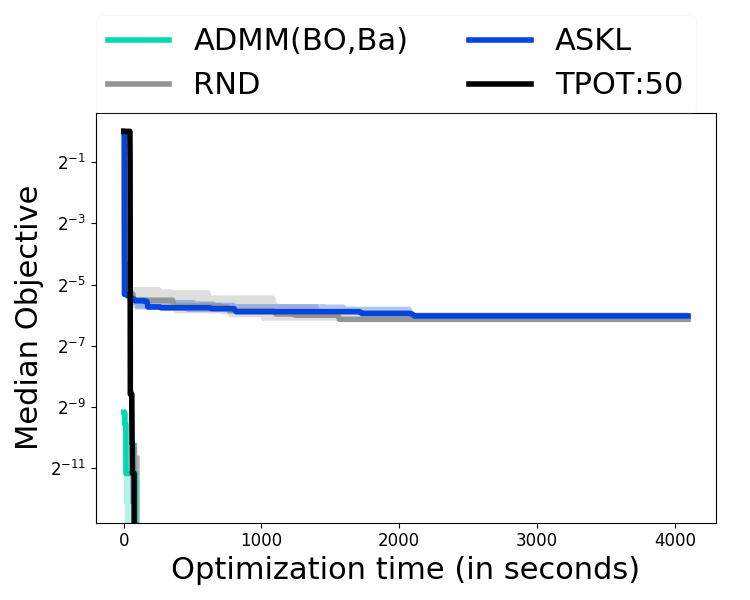}
    \caption{Ionosphere}
    \end{subfigure}
    ~ 
    \begin{subfigure}{0.185\textwidth}
    \includegraphics[width=\textwidth]{{app_images/oil_spill.iqr.search.val_perf.linx}.png}
    \caption{Oil spill}
    \end{subfigure}
    ~ 
    \begin{subfigure}{0.185\textwidth}
    \includegraphics[width=\textwidth]{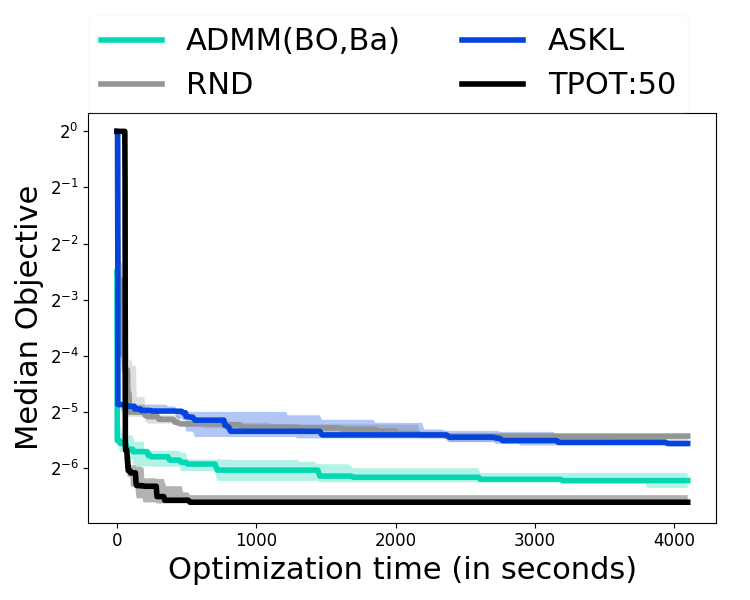}
    \caption{fri-c2}
    \end{subfigure}
    ~ 
    \begin{subfigure}{0.185\textwidth}
    \includegraphics[width=\textwidth]{{app_images/pc3.iqr.search.val_perf.linx}.png}
    \caption{PC3}
    \end{subfigure}
    ~ 
    \begin{subfigure}{0.185\textwidth}
    \includegraphics[width=\textwidth]{{app_images/pc4.iqr.search.val_perf.linx}.png}
    \caption{PC4}
    \end{subfigure}
    ~ 
    \begin{subfigure}{0.185\textwidth}
    \includegraphics[width=\textwidth]{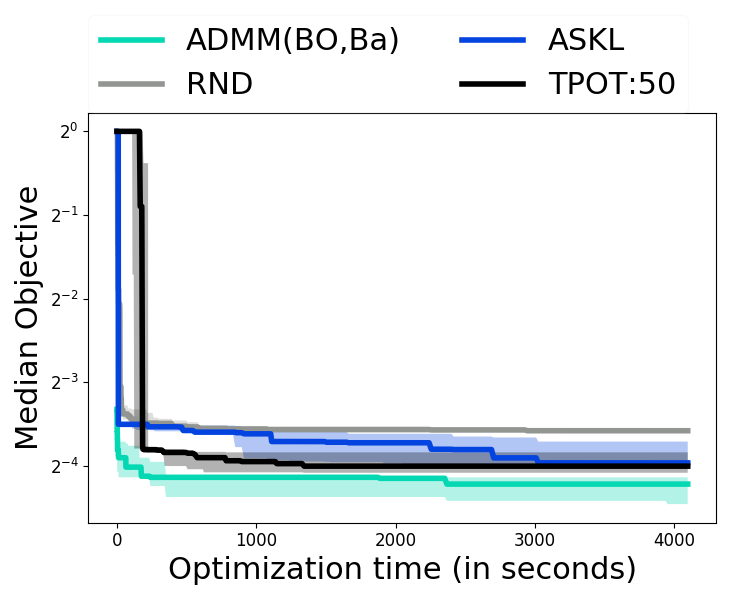}
    \caption{Space GA}
    \end{subfigure}
    ~ 
    \begin{subfigure}{0.185\textwidth}
    \includegraphics[width=\textwidth]{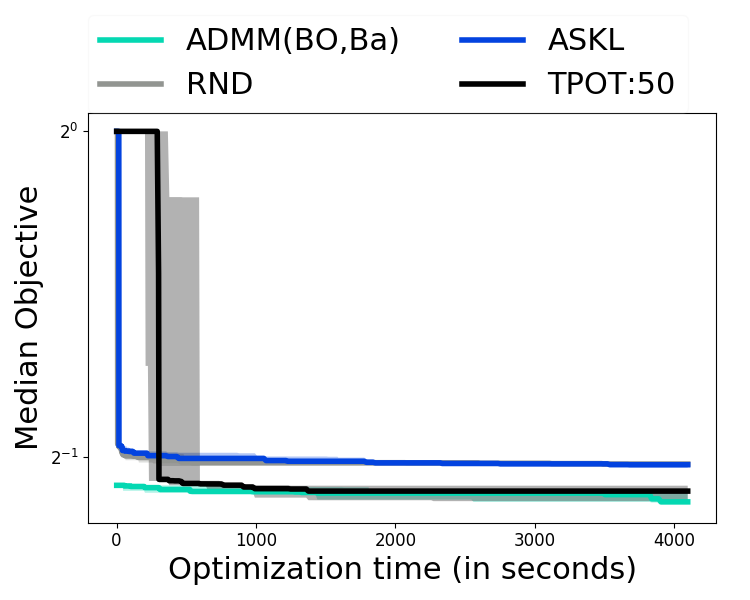}
    \caption{Pollen}
    \end{subfigure}
    ~ 
    \begin{subfigure}{0.185\textwidth}
    \includegraphics[width=\textwidth]{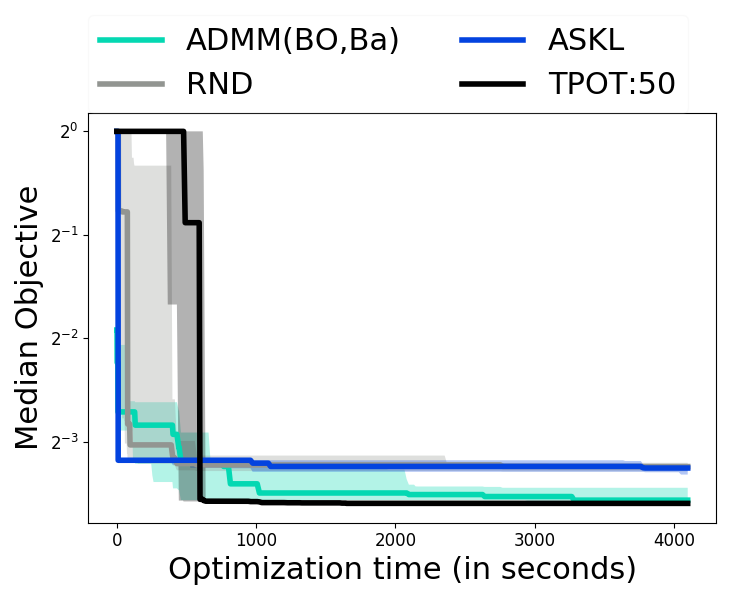}
    \caption{Ada-agnostic}
    \end{subfigure}
    ~ 
    \begin{subfigure}{0.185\textwidth}
    \includegraphics[width=\textwidth]{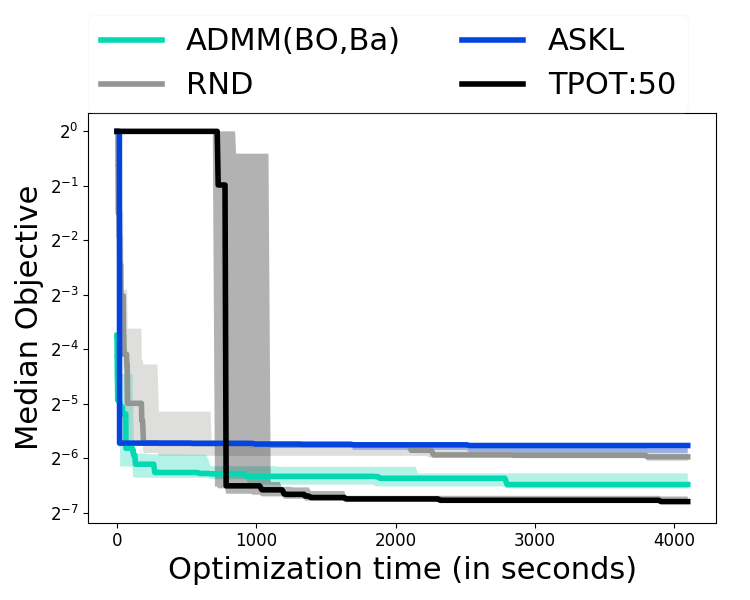}
    \caption{Sylvine}
    \end{subfigure}
    ~ 
    \begin{subfigure}{0.185\textwidth}
    \includegraphics[width=\textwidth]{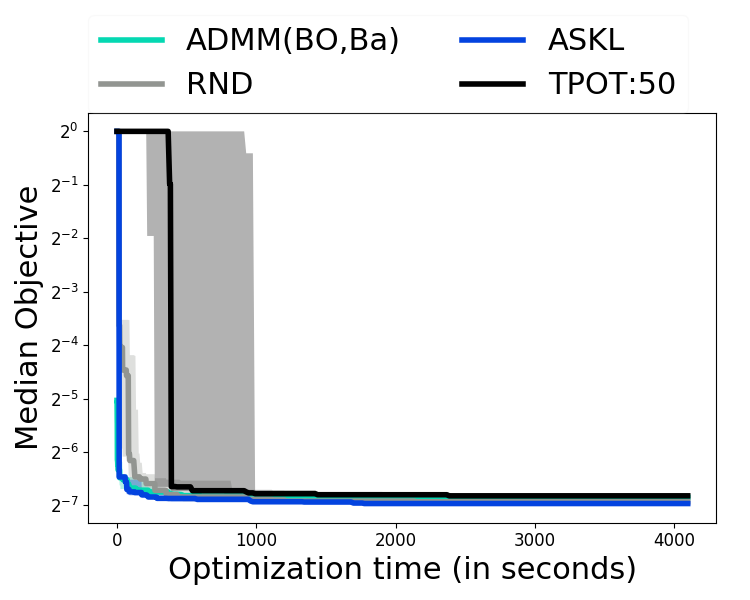}
    \caption{Page-blocks}
    \end{subfigure}
    ~ 
    \begin{subfigure}{0.185\textwidth}
    \includegraphics[width=\textwidth]{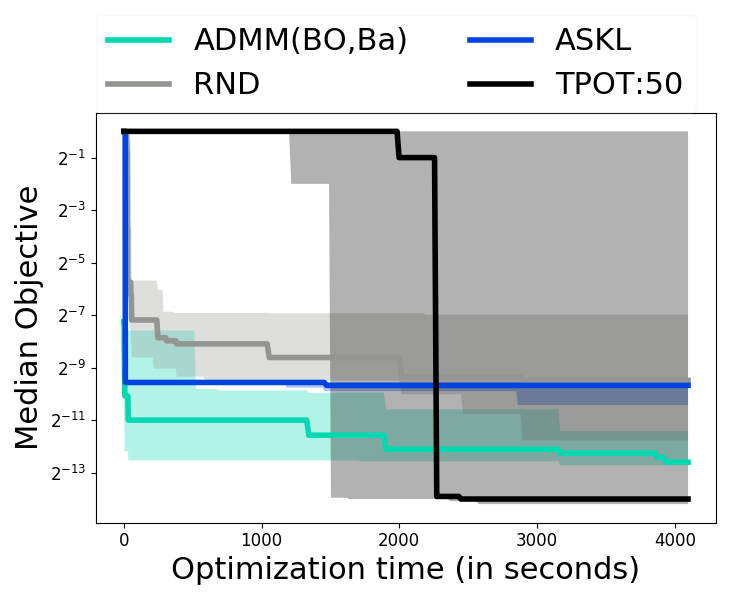}
    \caption{Optdigits}
    \end{subfigure}
    ~ 
    \begin{subfigure}{0.185\textwidth}
    \includegraphics[width=\textwidth]{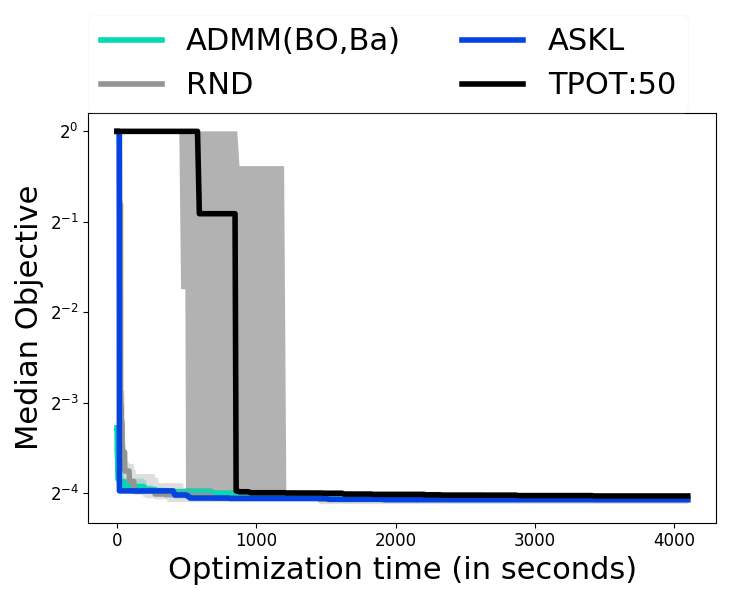}
    \caption{Wind}
    \end{subfigure}
    ~ 
    \begin{subfigure}{0.185\textwidth}
    \includegraphics[width=\textwidth]{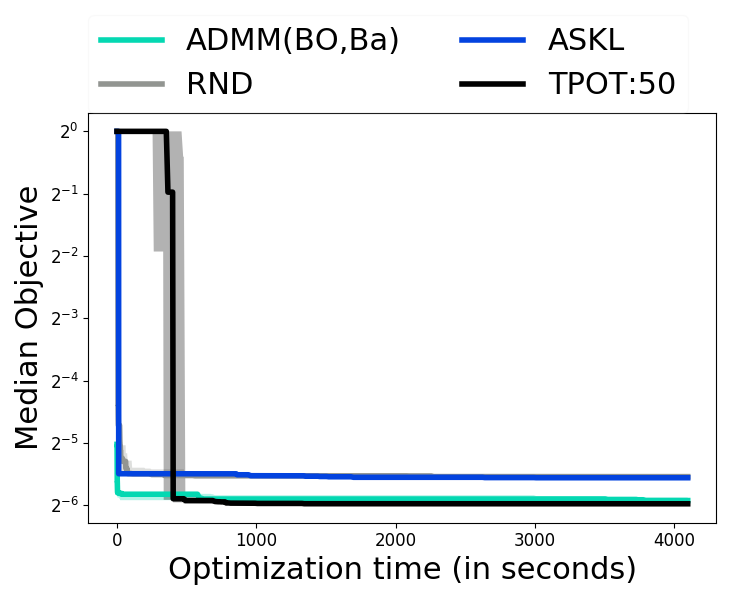}
    \caption{Delta-Ailerons}
    \end{subfigure}
    ~
%
    \begin{subfigure}{0.185\textwidth}
    \includegraphics[width=\textwidth]{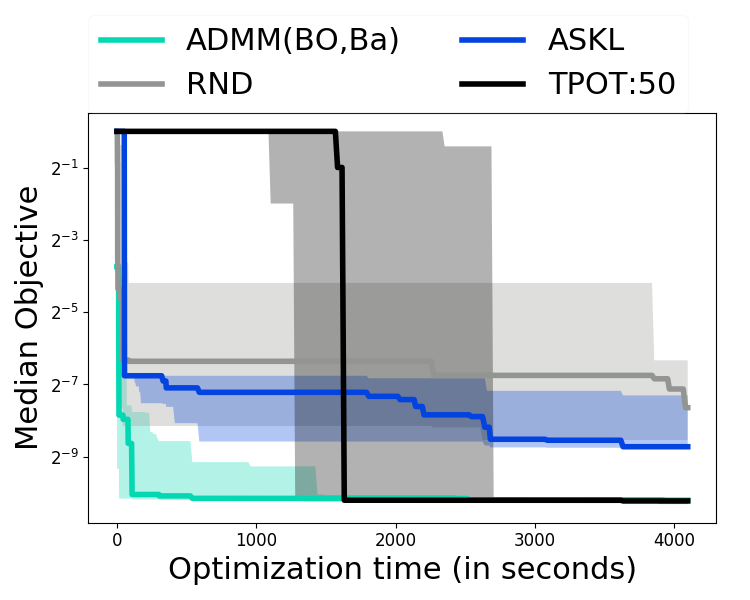}
    \caption{Ringnorm}
    \end{subfigure}
    ~
    \begin{subfigure}{0.185\textwidth}
    \includegraphics[width=\textwidth]{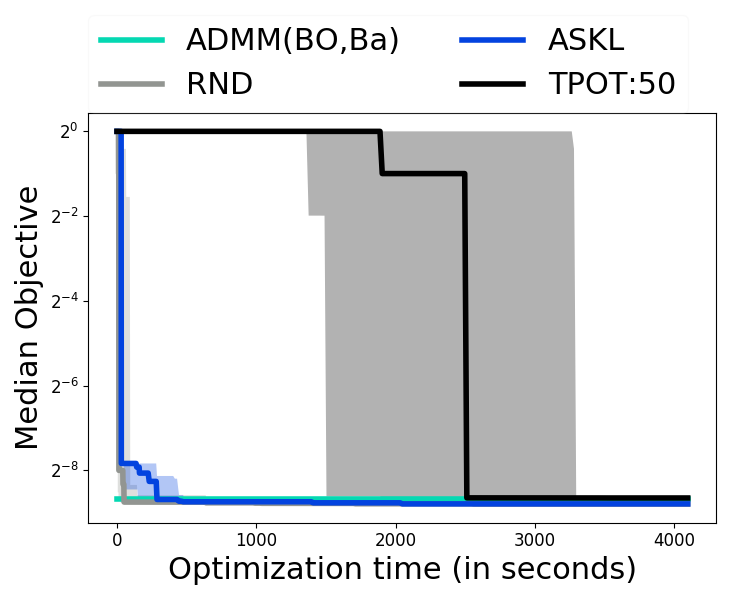}
    \caption{Twonorm}
    \end{subfigure}
    ~ 
    \begin{subfigure}{0.185\textwidth}
    \includegraphics[width=\textwidth]{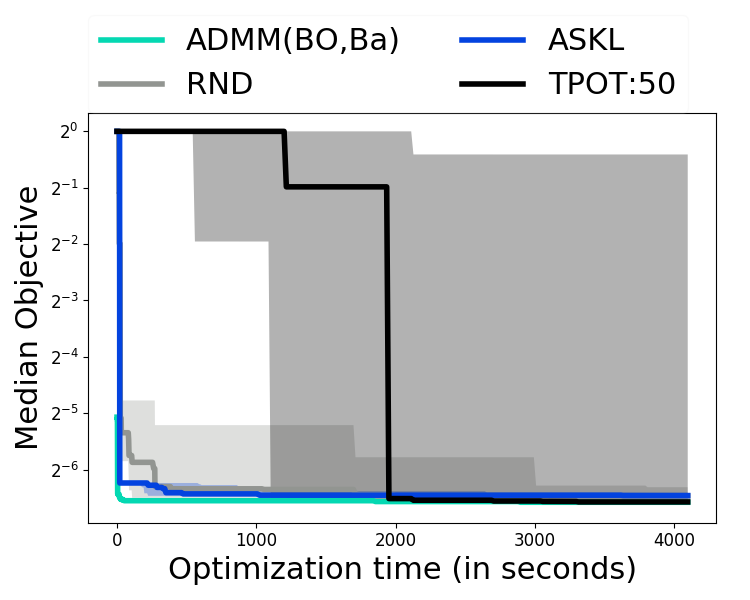}
    \caption{Bank8FM}
    \end{subfigure}
    ~ 
    \begin{subfigure}{0.185\textwidth}
    \includegraphics[width=\textwidth]{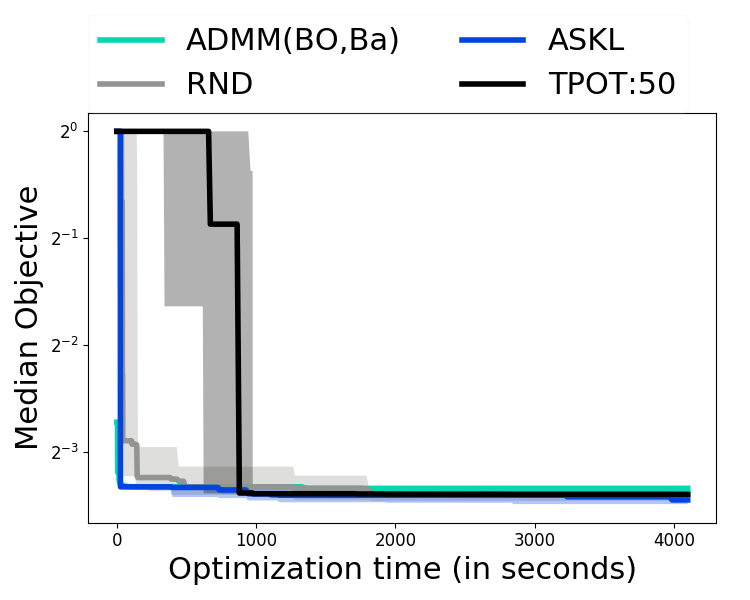}
    \caption{Puma8NH}
    \end{subfigure}
    ~ 
    \begin{subfigure}{0.185\textwidth}
    \includegraphics[width=\textwidth]{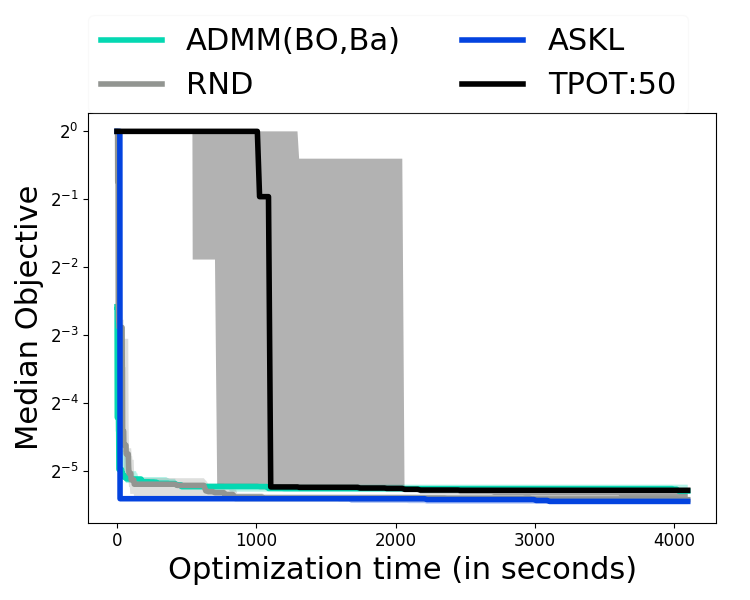}
    \caption{CPU small}
    \end{subfigure}
    ~ 
    \begin{subfigure}{0.185\textwidth}
    \includegraphics[width=\textwidth]{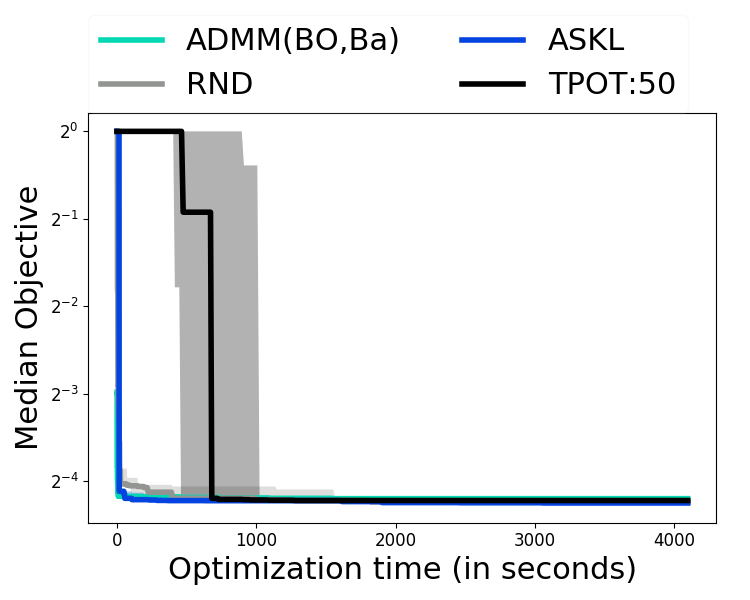}
    \caption{Delta elevators}
    \end{subfigure}
    ~ 
    \begin{subfigure}{0.185\textwidth}
    \includegraphics[width=\textwidth]{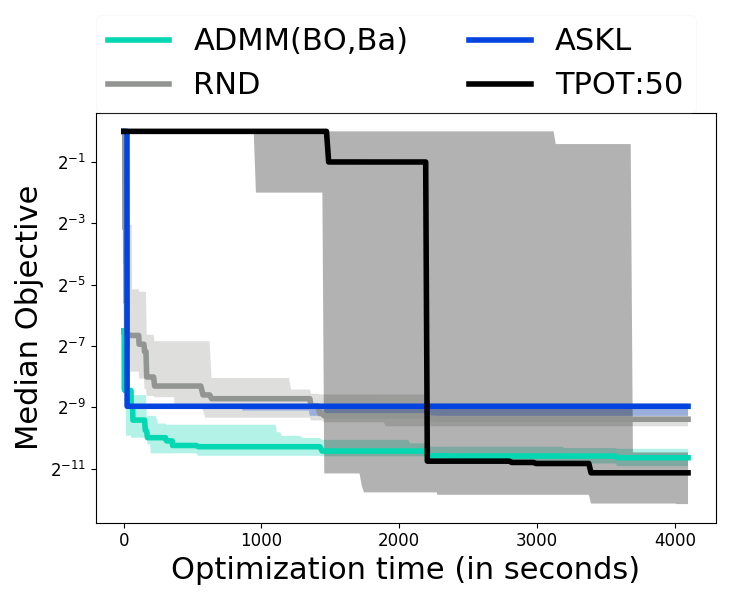}
    \caption{Japanese Vowels}
    \end{subfigure}
    ~ 
    \begin{subfigure}{0.185\textwidth}
    \includegraphics[width=\textwidth]{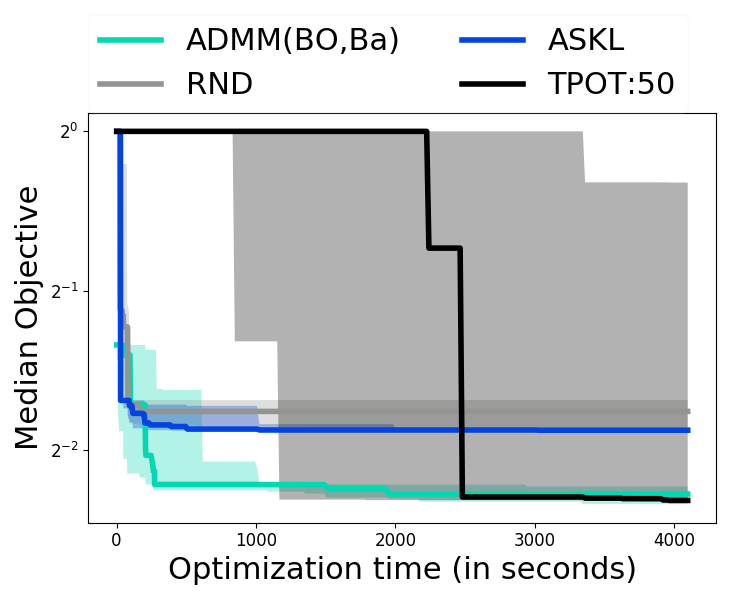}
    \caption{HCDR}
    \end{subfigure}
    ~ 
    \begin{subfigure}{0.185\textwidth}
    \includegraphics[width=\textwidth]{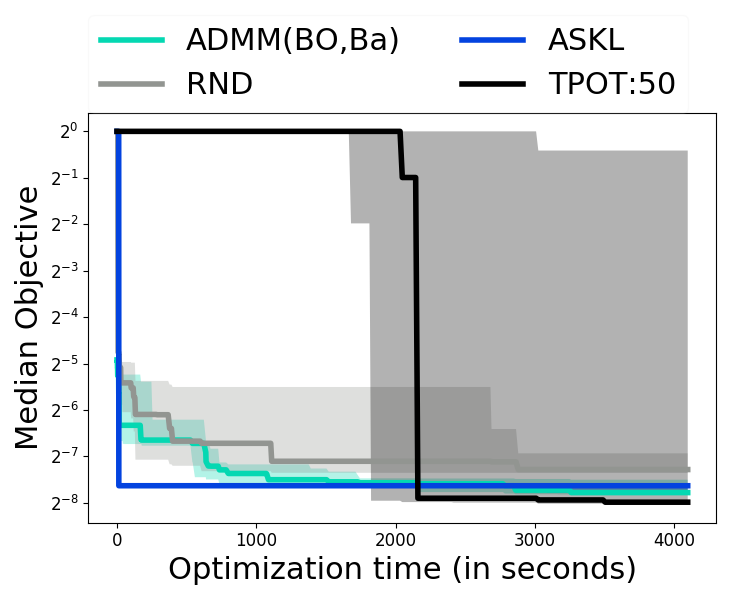}
    \caption{Phishing websites}
    \end{subfigure}
    ~
    \begin{subfigure}{0.185\textwidth}
    \includegraphics[width=\textwidth]{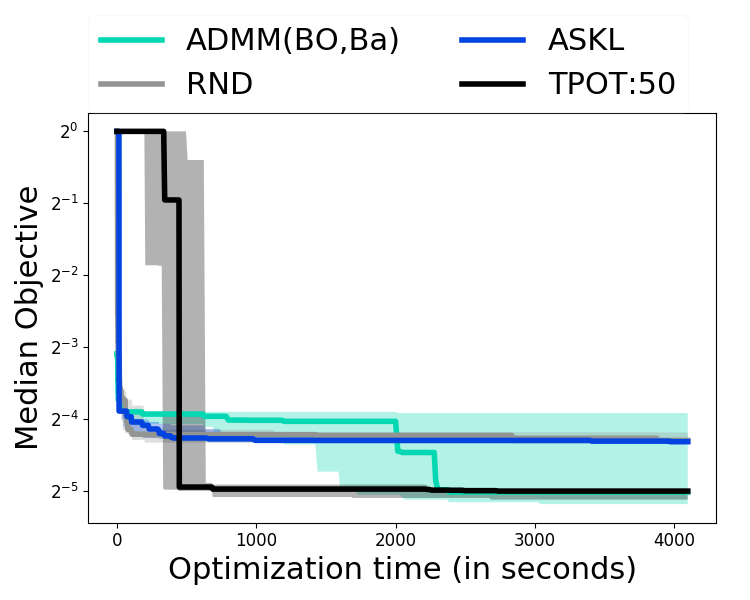}
    \caption{Mammography}
    \end{subfigure}
    ~ 
    \begin{subfigure}{0.185\textwidth}
    \includegraphics[width=\textwidth]{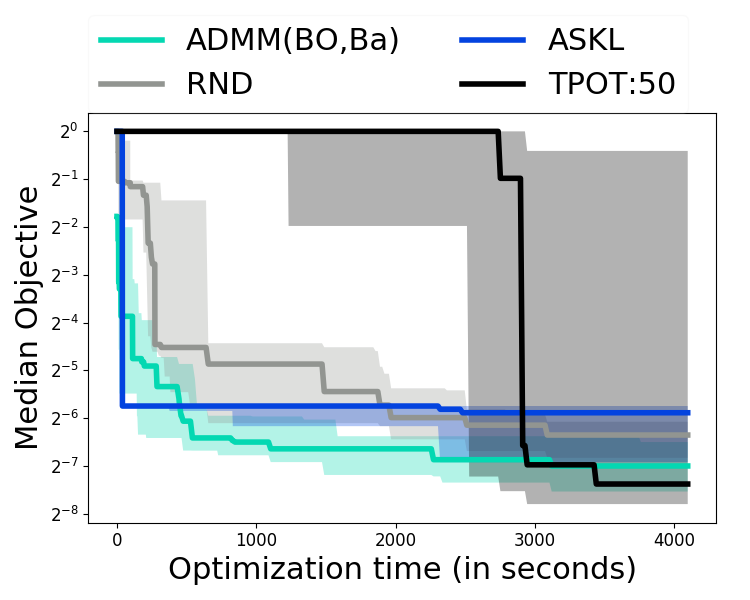}
    \caption{EEG-eye-state}
    \end{subfigure}
    ~ 
    \begin{subfigure}{0.185\textwidth}
    \includegraphics[width=\textwidth]{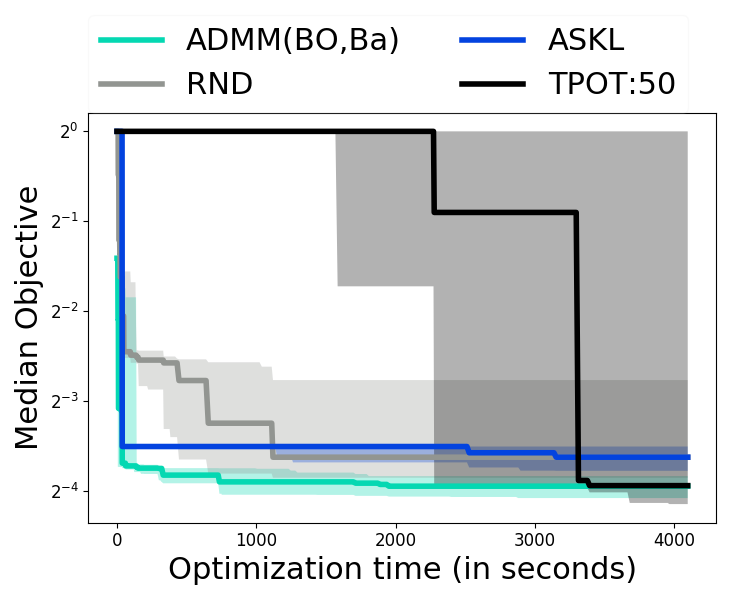}
    \caption{Elevators}
    \end{subfigure}
    ~ 
    \begin{subfigure}{0.185\textwidth}
    \includegraphics[width=\textwidth]{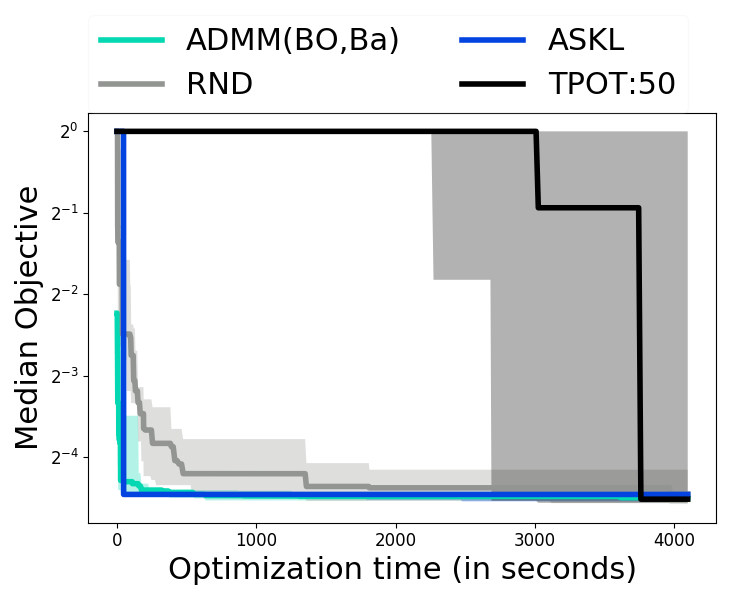}
    \caption{Cal housing}
    \end{subfigure}
    ~ 
    \begin{subfigure}{0.185\textwidth}
    \includegraphics[width=\textwidth]{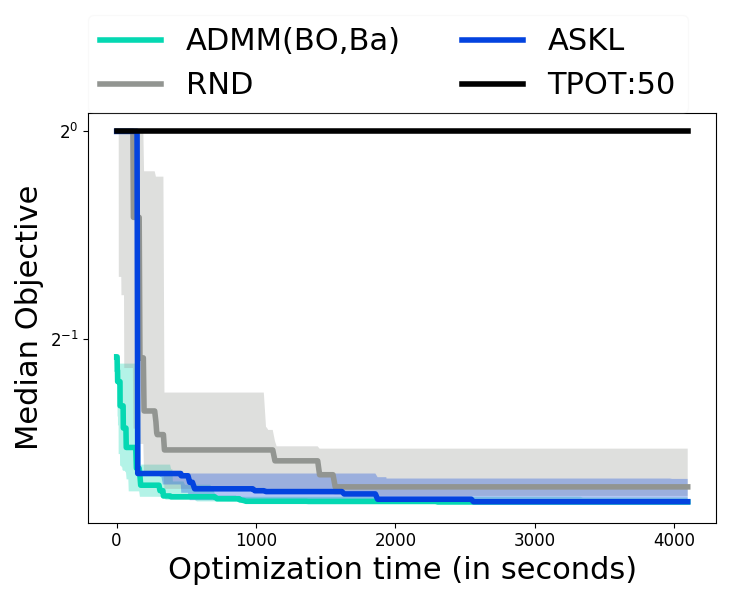}
    \caption{MLSS2017\#2}
    \end{subfigure}
    ~
    \begin{subfigure}{0.185\textwidth}
    \includegraphics[width=\textwidth]{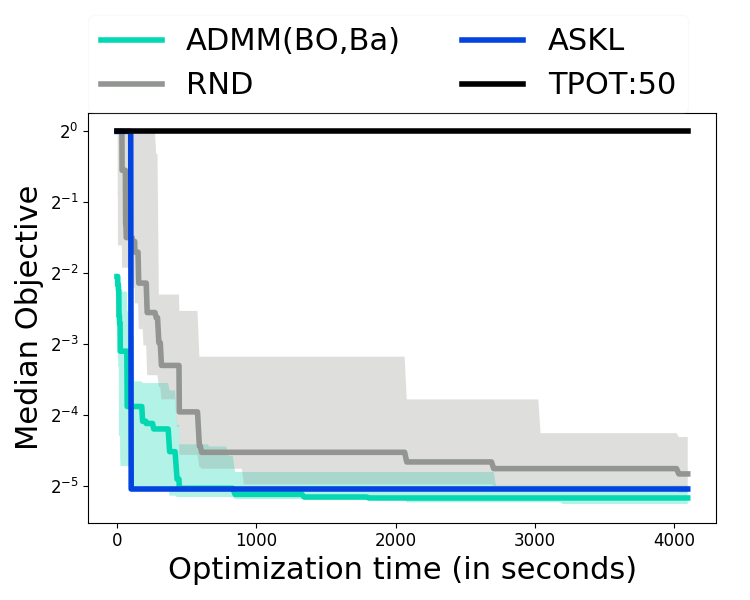}
    \caption{Electricity}
    \end{subfigure}
    \caption{Search/optimization time vs. median validation performance with the inter-quartile range over 10 trials ({\em please view in color}). The curves colored \textcolor{aquamarine}{Aquamarine}, \textcolor{gray}{Grey}, \textcolor{blue}{Blue} and \textcolor{black}{Black} correspond respectively to ADMM(BO,Ba), RND, ASKL and TPOT50.}
    \label{fig:time-v-valperf-iqr-all-double}
\end{figure}

\clearpage\pagebreak

\section{Computing the group-disparity fairness metric with respect to classification metric $\varepsilon$} \label{asec:group-disparity}
\paragraph{Computing the black-box function.}
The black-box objective $f(\bz, \btheta, \mathcal{A})$ is computed as follows for holdout-validation with some metric $\varepsilon$ (the metric $\varepsilon$ can be anything such as zero-one loss or area under the ROC curve):
\begin{itemize}
    \item Let $m$ be the pipeline specified by $(\bz, \btheta)$ 
    \item Split data set $\mathcal{A}$ into training set $\mathcal{A}_t$ and validation set $\mathcal{A}_v$
    \item Train the pipeline $m$ with training set $\mathcal{A}_t$ to get $m_{\mathcal{A}_t}$
    \item Evaluate the trained pipeline  $m_{\mathcal{A}_t}$ on the validation set $\mathcal{A}_v$ as follows:
    \begin{equation}
        \label{aeq:val_perf}
         \varepsilon\left( \mathcal{A}_t, \mathcal{A}_v\right) = \varepsilon\left( \left\{ \left(y, m_{\mathcal{A}_t} (x)\right) \forall (x,y) \in \mathcal{A}_v \right\} \right),
    \end{equation}
    where $m_{\mathcal{A}_t} (x)$ is the prediction of the trained pipeline $m_{\mathcal{A}_t}$ on any test point $x$ with label $y$ and 
    \begin{equation}
        \label{aeq:bb-eval}
        f(\bz, \btheta, \mathcal{A}) = \varepsilon\left( \mathcal{A}_t, \mathcal{A}_v\right).
    \end{equation}
\end{itemize}
For $k$-fold cross-validation, using the above notation, the objective is computed as follows:
\begin{itemize}
    \item Split data set $\mathcal{A}$ into training set $\mathcal{A}_{t_i}$ and validation set $\mathcal{A}_{v_i}$ for each of the $i = 1, \ldots, k$ folds
    \item For a pipeline $m$ specified with $(\bz, \btheta)$, the objective is computed as
    \begin{equation}
        \label{aeq:kf_perf}
        f(\bz, \btheta, \mathcal{A}) = \frac{1}{k} \sum_{i = 1}^k  \varepsilon\left( \mathcal{A}_{t_i}, \mathcal{A}_{v_i} \right).
    \end{equation}
\end{itemize}
%
%
\paragraph{NOTE.} The splitting of the data set $\mathcal{A}$ in training/validation pairs $(\mathcal{A}_t, \mathcal{A}_v)$ should be the same across all evaluations of $(\bz, \btheta)$. Similarly, the $k$-fold splits should be the same across all evaluations of $(\bz, \btheta)$.
\paragraph{Computing group disparate impact.}
Continuing with the notation defined in the previous subsection, for any given (test/validation) set $\mathcal{A}_v$, assume that we have a (probably user specified) ``protected'' feature $d$ and a grouping $G_d(\mathcal{A}_v) = \{ A_1, A_2, \ldots \}$ of the set $\mathcal{A}_v$ based on this feature (generally, $A_j \cap A_k = \emptyset \forall j \not= k$ and $\cup_{A_j \in G_d(A)} A_j = \mathcal{A}_v$). Then, given the objective function $f$ corresponding to the metric $\varepsilon$, the corresponding group disparate impact with holdout validation is given as
\begin{equation} \label{eq:group-parity-hv}
p(\bz, \btheta, \mathcal{A})  = \max_{A_j \in G_d(\mathcal{A}_v)} \varepsilon\left( \mathcal{A}_t, A_j \right) - \min_{A_j \in G_d(\mathcal{A}_v)} \varepsilon\left( \mathcal{A}_t, A_j \right)
\end{equation}
For $k$-fold cross-validated group disparate impact with the grouping per fold as $G_d(\mathcal{A}_{v_i}) = \left\{ A_{i,1},  A_{i, 2}, \ldots \right\}$, we use the following:
\begin{equation} \label{eq:group-parity-kcv}
p(\bz, \btheta, \mathcal{A})  = \frac{1}{k} \sum_{i=1}^k \left\{ \max_{A_{i,j} \in G_d(\mathcal{A}_{v_i})} \varepsilon\left( \mathcal{A}_{t_i}, A_{i,j} \right) - \min_{A_{i,j} \in G_d(\mathcal{A}_{v_i})} \varepsilon\left( \mathcal{A}_{t_i}, A_{i,j} \right) \right\}
\end{equation}
Example considered here:
\begin{itemize}
    \item Dataset $\mathcal{A}$: Home credit default risk Kaggle challenge
    \item Metric $\varepsilon$: Area under ROC curve
    \item Protected feature $d$: \texttt{DAYS\_BIRTH}
    \item Grouping $G_d$ based on $d$: Age groups 20-30, 30-40, 40-50, 50-60, 60-70
\end{itemize}

\clearpage\pagebreak

\section{Benchmarking Adaptive ADMM} \label{asec:adadmm}

It is common in ADMM to solve the sub-problems to higher level of approximation in the initial ADMM iterations and to an increasingly smaller levels of approximation as the ADMM progresses (instead of the same level of approximation for all ADMM iterations). We make use of this same {\em adaptive ADMM} and demonstrate that, empirically, the adaptive scheme produces expected gains in the AutoML problem as well. 

\begin{figure}[htb]
    \begin{subfigure}{0.35\textwidth}
    \includegraphics[width=\textwidth]{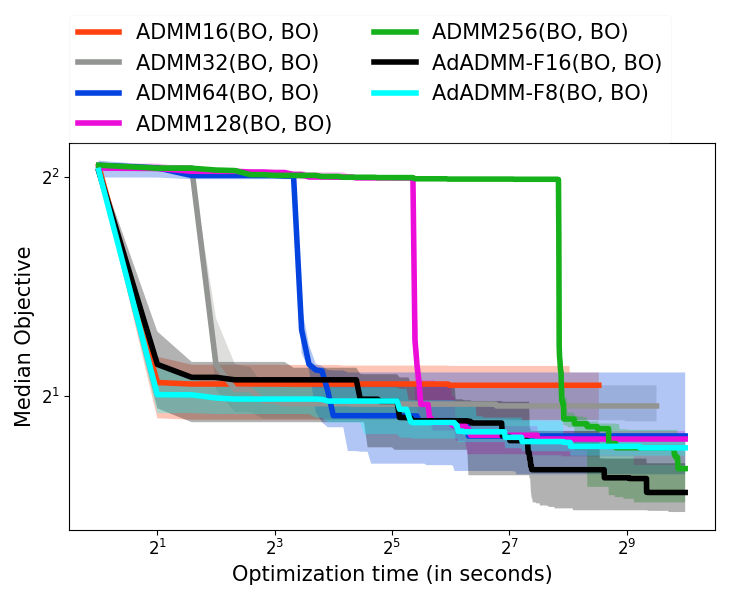}
    \caption{Artificial black-box objective}
    \end{subfigure}
    \\
    \begin{subfigure}{0.235\textwidth}
    \includegraphics[width=\textwidth]{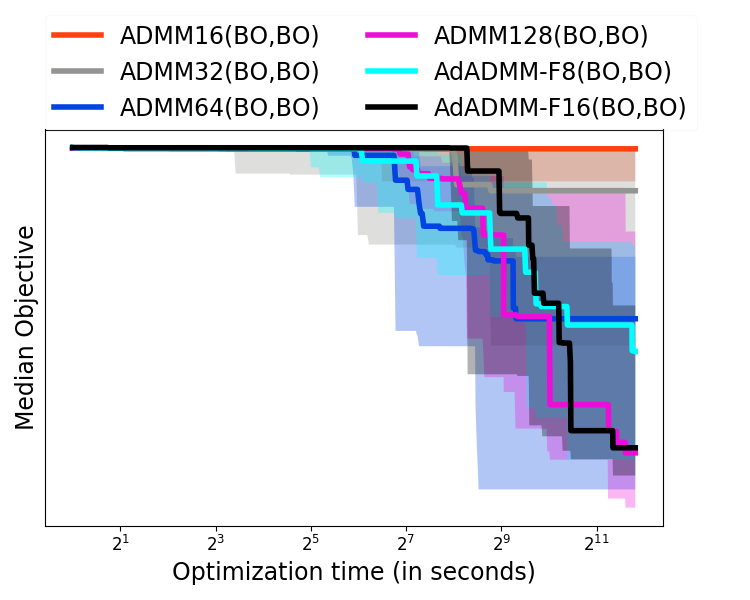}
    \caption{Bank8FM}
    \end{subfigure}
    ~ 
    \begin{subfigure}{0.235\textwidth}
    \includegraphics[width=\textwidth]{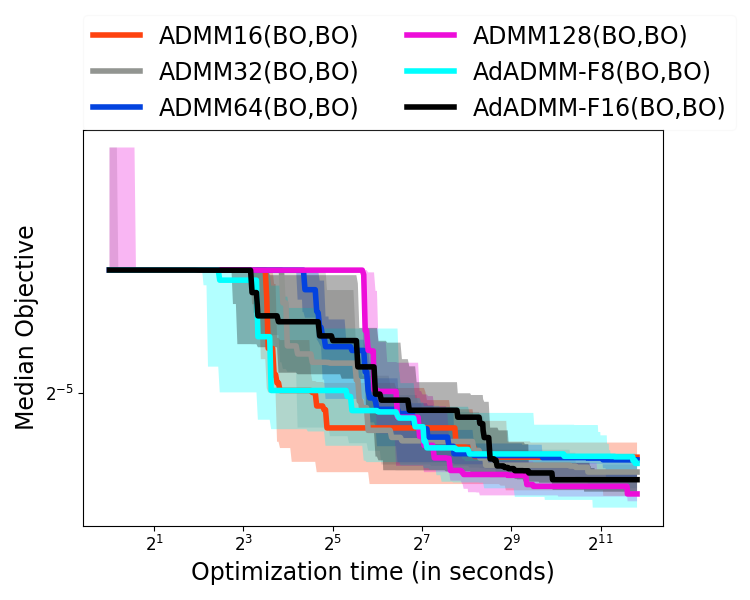}
    \caption{CPU small}
    \end{subfigure}
    ~ 
    \begin{subfigure}{0.235\textwidth}
    \includegraphics[width=\textwidth]{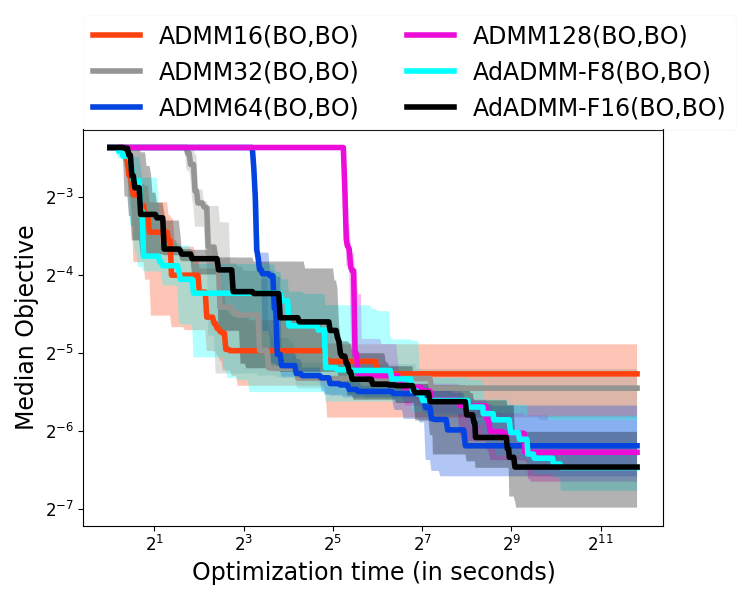}
    \caption{fri-c2}
    \end{subfigure}
    ~ 
    \begin{subfigure}{0.235\textwidth}
    \includegraphics[width=\textwidth]{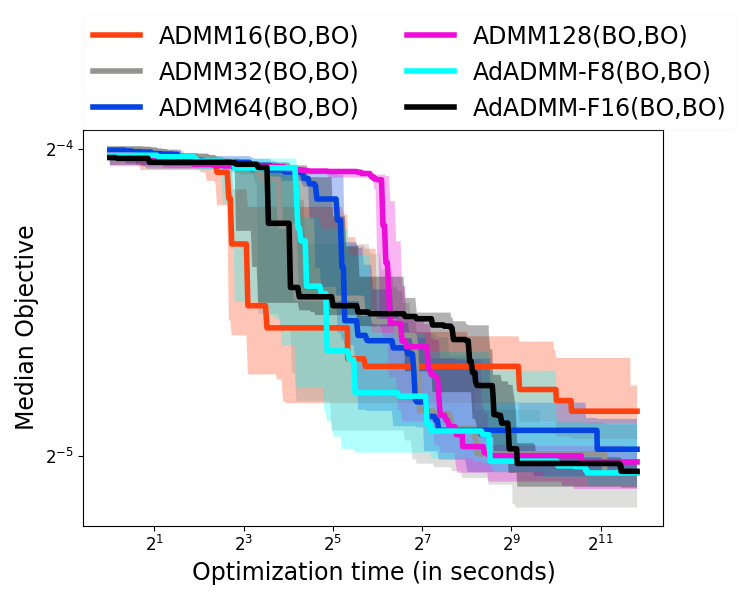}
    \caption{PC4}
    \end{subfigure}
    ~ 
    \begin{subfigure}{0.235\textwidth}
    \includegraphics[width=\textwidth]{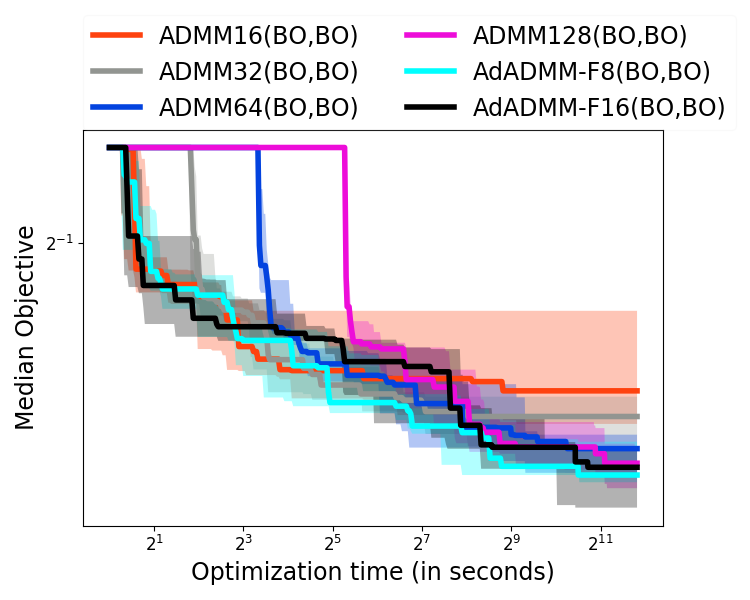}
    \caption{Pollen}
    \end{subfigure}
    ~ 
    \begin{subfigure}{0.235\textwidth}
    \includegraphics[width=\textwidth]{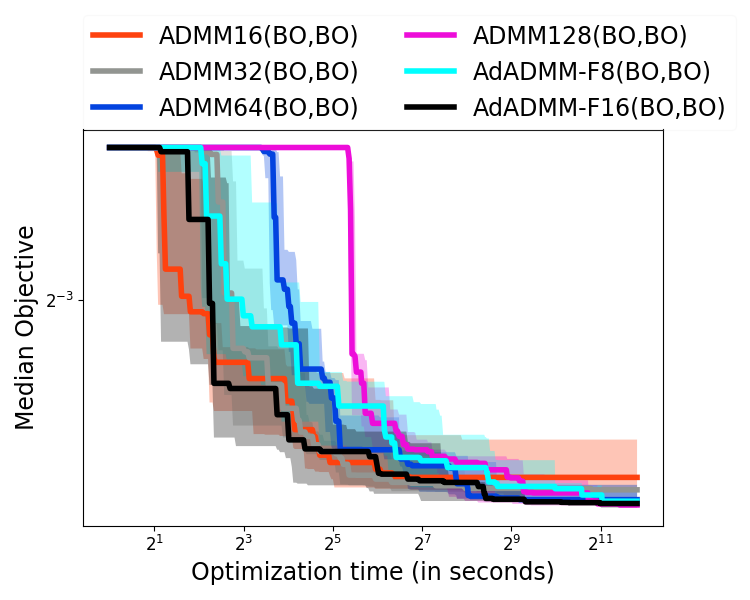}
    \caption{Puma8NH}
    \end{subfigure}
    ~ 
    \begin{subfigure}{0.235\textwidth}
    \includegraphics[width=\textwidth]{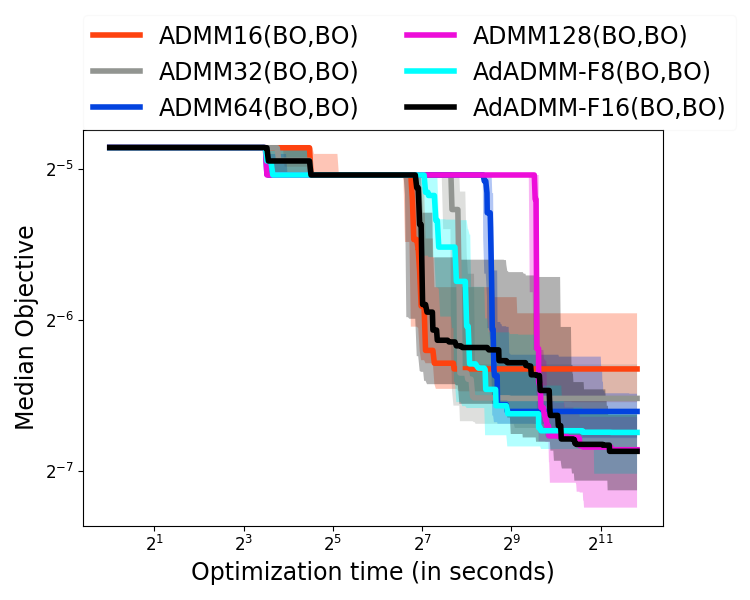}
    \caption{Sylvine}
    \end{subfigure}
    ~ 
    \begin{subfigure}{0.235\textwidth}
    \includegraphics[width=\textwidth]{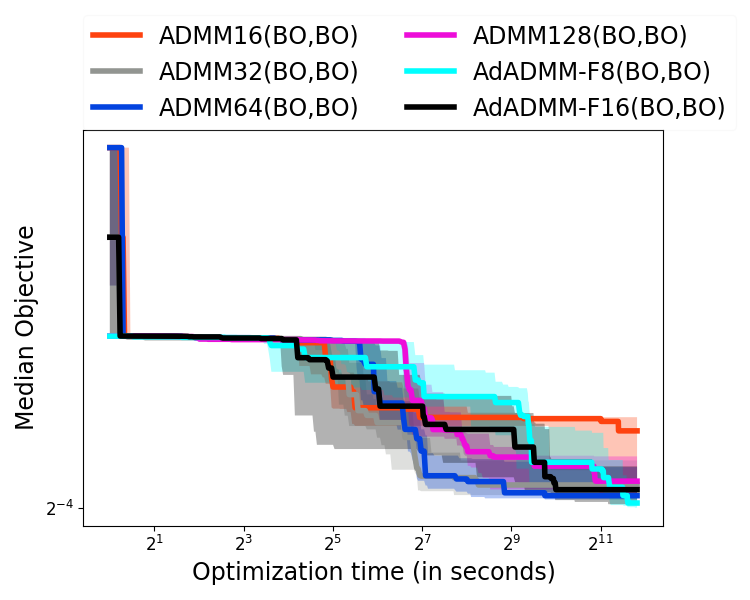}
    \caption{Wind}
    \end{subfigure}
    \caption{Search/optimization time (in seconds) vs. median validation performance with the inter-quartile range over 10 trials ({\em please view in color and note the log scale on both the horizontal and vertical axes}).}
    \label{fig:time-v-valperf-iqr-all-adadmm}
\end{figure}

In this empirical evaluation, we use BO for solving both the \eqref{eq: theta_min} and the \eqref{eq: z_step} problems. For ADMM with a fixed level of approximation (subsequently noted as {\em fixed ADMM}), we solve the sub-problems to a fixed number $I$ of BO iterations with $I = 16, 32, 64, 128$ (also $256$ for the artificial objective described in Appendix \ref{asec:artificial-obj})) denoted by ADMM$I$(BO,BO) (for example, ADMM16(BO,BO)). For ADMM with varying level of approximation, we start with $16$ BO iterations for the sub-problems and progressively increase it with an additive factor $F = 8$ or $16$ with every ADMM iteration until $128$ (until $256$ for the artificial objective) denoted by AdADMM-F8(BO,BO) and AdADMM-F16(BO,BO) respectively. The optimization is run for 3600 seconds for all the data sets and for 1024 seconds for the artificial objective function. The convergence plots are presented in Figure \ref{fig:time-v-valperf-iqr-all-adadmm}.

The figures indicate the expected behavior -- fixed ADMM with small $I$ dominate for small optimization time scale but saturate soon while fixed ADMM with large $I$ require a significant amount of startup time but then eventually lead to the best performance for the larger time scales. Adaptive ADMM (for both values of $F$) appears to somewhat match the performance of the best fixed ADMM for every time scale. This behavior is exemplified with the artificial black-box objective (described in Appendix \ref{asec:artificial-obj}) but is also present on the AutoML problem with real datasets.

\clearpage\pagebreak

\section{Evaluating the benefits of problem splitting in ADMM} \label{asec:opsplit}

\begin{figure}[htb]
    \begin{subfigure}{0.35\textwidth}
    \includegraphics[width=\textwidth]{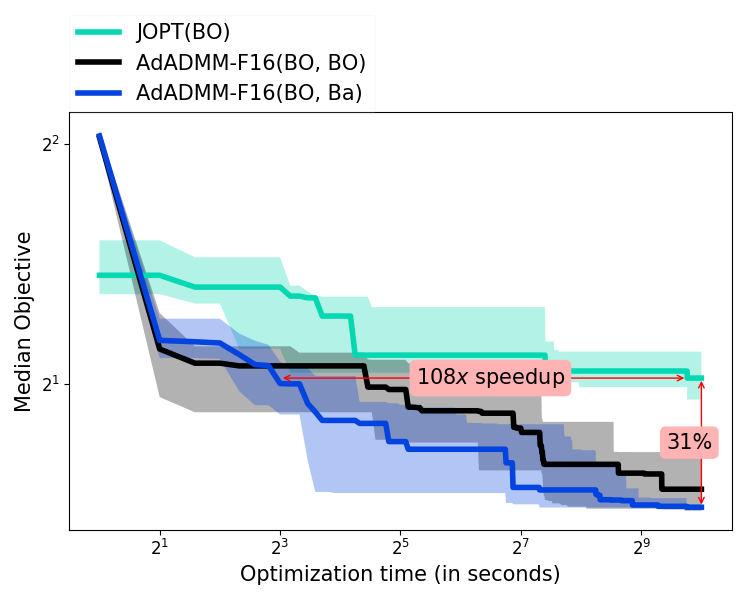}
    \caption{Artificial black-box objective}
    \end{subfigure}
    \\
    \begin{subfigure}{0.235\textwidth}
    \includegraphics[width=\textwidth]{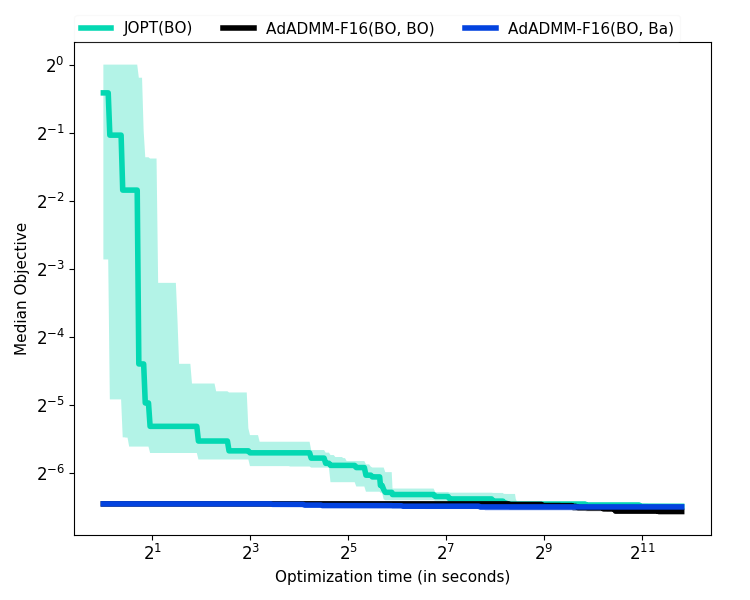}
    \caption{Bank8FM}
    \end{subfigure}
    ~ 
    \begin{subfigure}{0.235\textwidth}
    \includegraphics[width=\textwidth]{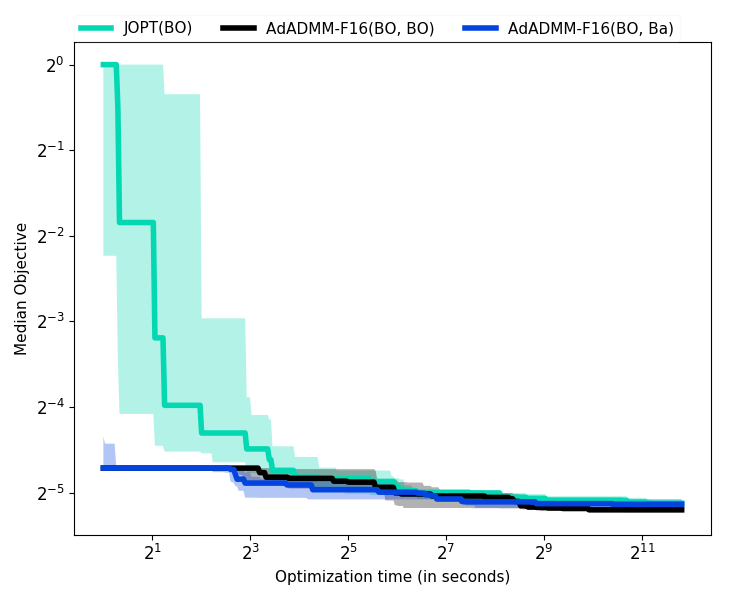}
    \caption{CPU small}
    \end{subfigure}
    ~ 
    \begin{subfigure}{0.235\textwidth}
    \includegraphics[width=\textwidth]{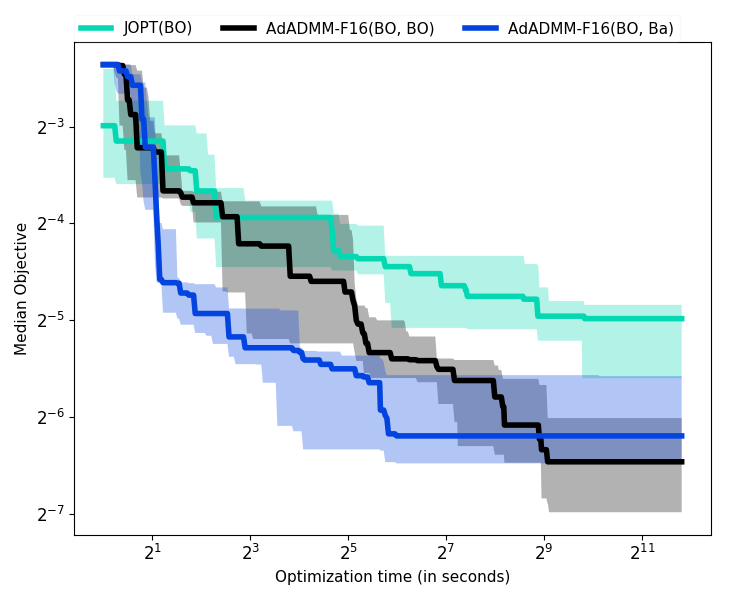}
    \caption{fri-c2}
    \end{subfigure}
    ~ 
    \begin{subfigure}{0.235\textwidth}
    \includegraphics[width=\textwidth]{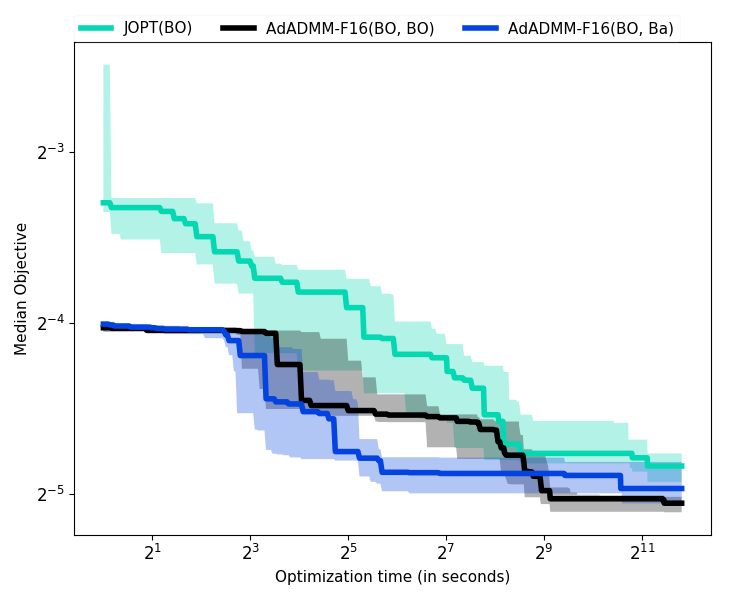}
    \caption{PC4}
    \end{subfigure}
    ~ 
    \begin{subfigure}{0.235\textwidth}
    \includegraphics[width=\textwidth]{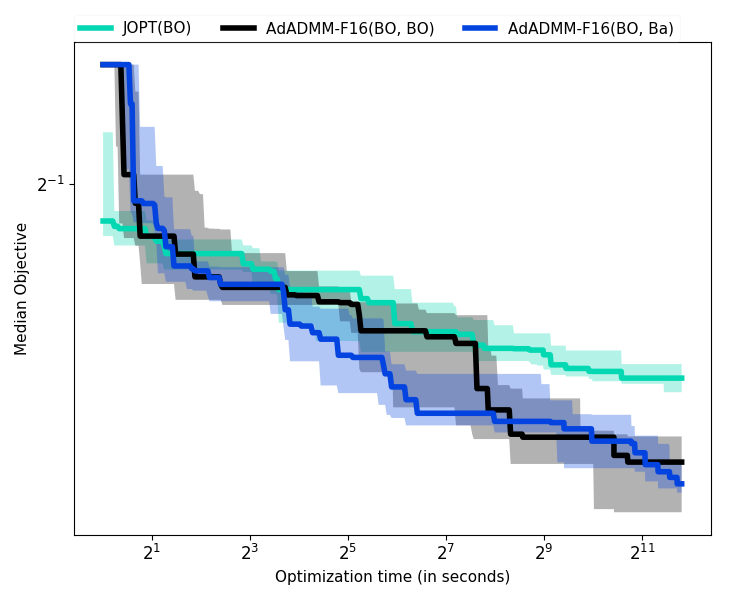}
    \caption{Pollen}
    \end{subfigure}
    ~ 
    \begin{subfigure}{0.235\textwidth}
    \includegraphics[width=\textwidth]{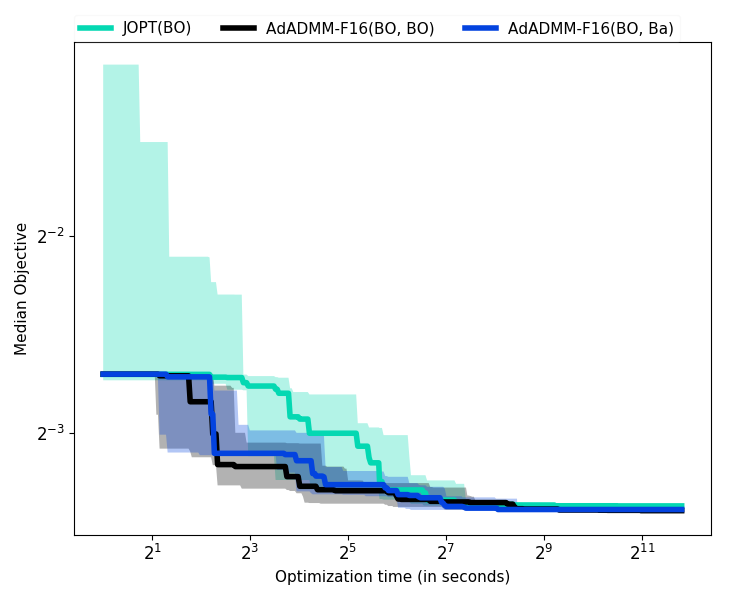}
    \caption{Puma8NH}
    \end{subfigure}
    ~ 
    \begin{subfigure}{0.235\textwidth}
    \includegraphics[width=\textwidth]{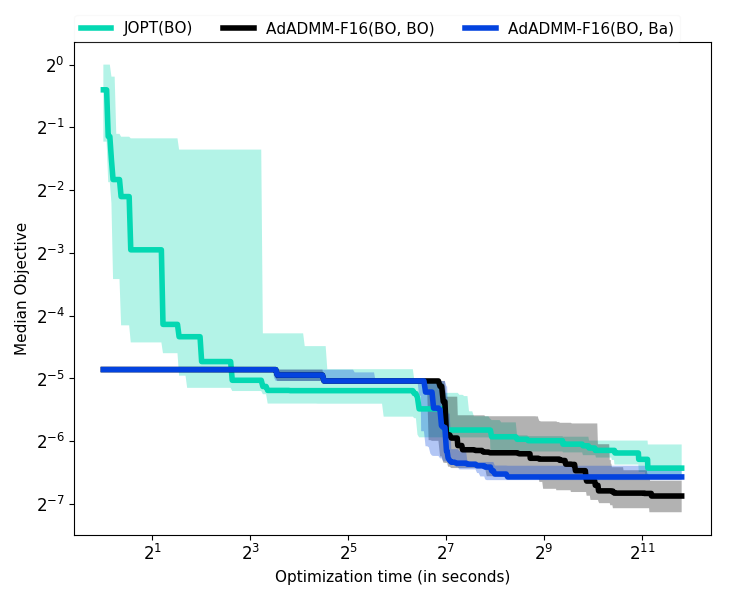}
    \caption{Sylvine}
    \end{subfigure}
    ~ 
    \begin{subfigure}{0.235\textwidth}
    \includegraphics[width=\textwidth]{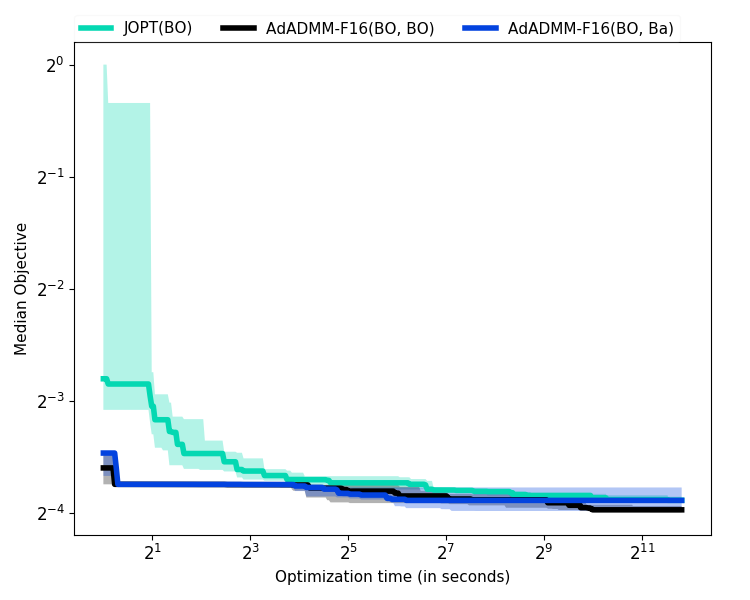}
    \caption{Wind}
    \end{subfigure}
    \caption{Search/optimization time vs. median validation performance with the inter-quartile range over 10 trials ({\em please view in color and note the log scale on both the horizontal and vertical axes}).}
    \label{fig:time-v-valperf-iqr-all-opsplit}
\end{figure}

In this empirical evaluation, we wish to demonstrate the gains from (i) splitting the AutoML problem \eqref{eq: prob0} into smaller sub-problems which are solved in an alternating fashion, and (ii) using different solvers for the differently structured \eqref{eq: theta_min} and \eqref{eq: z_step} problems. First, we attempt to solve the complete {\em joint optimization problem} \eqref{eq: prob0} with BO, leading to a Gaussian Process with a large number of variables. We denote this as JOPT(BO). Then we utilize {\em adaptive ADMM} where we use BO for each of the \eqref{eq: theta_min} and \eqref{eq: z_step} problems in each of the ADMM iteration, denoted as AdADMM-F16(BO,BO). Finally, we use adaptive ADMM where we use BO for each of the \eqref{eq: theta_min} problem and Combinatorial Multi-Armed Bandits (CMAB) for the \eqref{eq: z_step} problem, denoted as AdADMM-F16(BO,Ba). For the artificial black-box objective (described in Appendix \ref{asec:artificial-obj}), the optimization is run for 1024 seconds. For the AutoML problem with the actual data sets, the optimization is run for 3600 seconds. The convergence of the different optimizers are presented in Figure \ref{fig:time-v-valperf-iqr-all-opsplit}.

The results for the artificial objective is a case where the black-box optimization dominates the optimization time (since the black-box evaluation is cheap). In this case, both versions of the adaptive ADMM significantly outperforms the single BO (JOPT(BO)) for the whole problem 2 seconds onwards, demonstrating the advantage of the problem splitting in ADMM. Between the two versions of the adaptive ADMM, AdADMM(BO,Ba) (Bandits for \eqref{eq: z_step}) outperforms AdADMM(BO,BO) (BO for \eqref{eq: z_step}). This is potentially because BO is designed for continuous variables and is mostly suited for the \eqref{eq: theta_min} problem, whereas the Bandits interpretation is better suited for the \eqref{eq: z_step} problem.  By the end of the optimization time budget, AdADMM(BO,Ba) improves the objective by around $31\%$ over JOPT(BO) ($5\%$ over AdADMM(BO,BO)), and achieves the objective reached by JOPT(BO) with a $108 \times$ speedup (AdADMM(BO,BO) with a $4\times$ speedup). 

On the AutoML problem with real data sets, the optimization time is mostly dominated by the black-box evaluation, but even in this case, the problem splitting with ADMM demonstrates significant gains over JOPT(BO). For example, on the fri-c2 dataset, the results indicate that the operator splitting in ADMM allows it to reach the final objective achieved by JOPT with over $150\times$ speedup, and then further improves upon that final objective by over $50\%$. On the Pollen dataset, we observe a speedup of around $25\times$ with a further improvement of $4\%$. Table \ref{atab:opsplit-gains} \& \ref{atab:opsplit-gains-2} summarize the significant gains from the problem splitting in ADMM.
\begin{table}[htb]
    \centering
    \begin{tabular}{|l|r|r|}
        \hline
        Dataset & Speedup & Improvement \\
        \hline\hline
        Artificial & $108\times$ & $31\%$  \\
        Bank8FM    & $10 \times$ & $0 \%$  \\
        CPU small  & $4  \times$ & $0 \%$  \\
        fri-c2     & $153\times$ & $56\%$  \\
        PC4        & $42 \times$ & $8 \%$  \\
        Pollen     & $25 \times$ & $4 \%$  \\
        Puma8NH    & $11 \times$ & $1 \%$  \\
        Sylvine    & $9  \times$ & $9 \%$  \\
        Wind       & $40 \times$ & $0 \%$  \\
        \hline
    \end{tabular}
    \caption{Comparing AdADMM(BO,Ba) to JOPT(BO), we list the speedup achieved by AdADMM(BO,Ba) to reach the best objective of JOPT(BO), and any further improvement in the objective. These numbers are generated using the aggregate performance of JOPT and AdADMM over 10 trials.}
    \label{atab:opsplit-gains}
\end{table}

\begin{table}[htb]
    \centering
    \begin{tabular}{|l|r|r|}
        \hline
        Dataset & Speedup & Improvement \\
        \hline\hline
        Artificial & $39 \times$ & $27\%$  \\
        Bank8FM    & $2  \times$ & $5 \%$  \\
        CPU small  & $5  \times$ & $5 \%$  \\
        fri-c2     & $25 \times$ & $64\%$  \\
        PC4        & $5  \times$ & $13\%$  \\
        Pollen     & $7  \times$ & $3 \%$  \\
        Puma8NH    & $4  \times$ & $1 \%$  \\
        Sylvine    & $2  \times$ & $26\%$  \\
        Wind       & $5  \times$ & $5 \%$  \\
        \hline
    \end{tabular}
    \caption{Comparing AdADMM(BO,BO) to JOPT(BO), we list the speedup achieved by AdADMM(BO,BO) to reach the best objective of JOPT(BO), and any further improvement in the objective. These numbers are generated using the aggregate performance of JOPT and AdADMM over 10 trials.}
    \label{atab:opsplit-gains-2}
\end{table}


\clearpage\pagebreak

\section{Artificial black-box objective} \label{asec:artificial-obj}

We wanted to devise an artificial black-box objective to study the behaviour of the proposed scheme that matches the properties of the AutoML problem \eqref{eq: prob0} where 
\begin{enumerate}
    \item The same pipeline (the same algorithm choices $\bz$ and the same hyperparameters $\btheta$ always gets the same value.
    \item The objective is not convex and possibly non-continuous.
    \item The objective captures the conditional dependence between $\bz_i$ and $\btheta_{ij}$ -- the objective is only dependent on the hyper-parameters $\btheta_{ij}$ if the corresponding $z_{ij} = 1$.
    \item Minor changes in the hyper-parameters $\btheta_{ij}$ can cause only small changes in the objective.
    \item The output of module $i$ is dependent on its input from module $i-1$.
\end{enumerate}

\paragraph{Novel artificial black-box objective.}
To this end, we propose the following novel black-box objective:
\begin{itemize}
    \item For each $(i,j), i \in [N], j \in [K_i]$, we fix a weight vector $\mathbf{w}_{ij}$ (each entry is a sample from $\mathcal{N}(0,1)$) and a seed $s_{ij}$.
    \item We set $f_0 = 0$.
    \item For each module $i$, we generate a value 
    $$v_i = \sum_{j} z_{ij}  \left| \frac{\mathbf{w}_{ij}^\top \btheta_{ij}}{\mathbf{1}^T \btheta_{ij}} \right|$$ 
    which only depends on the $\btheta_{ij}$ corresponding to the $z_{ij} = 1$, and the denominator ensures that the number (or range) of the hyper-parameters does not bias the objective towards (or away from) any particular algorithm. 
    \item We generate $n$ samples $\{ f_{i,1}, \ldots, f_{i,n} \} \sim \mathcal{N}(f_{i-1}, v_i)$ with the fixed seed $s_{ij}$, ensuring that the same value will be produced for the same pipeline.
    \item $f_i = \max\limits_{m = 1, \ldots, n} | f_{i,m} |$.
\end{itemize}
The basic idea behind this objective is that, for each operator, we create a random (but fixed) weight vector $\mathbf{w}_{ij}$ and take a weighted normalized sum of the hyper-parameters $\btheta_{ij}$ and use this sum as the scale to sample from a normal distribution (with a fixed seed $s_{ij}$) and pick the maximum absolute of $n$ (say 10) samples. For the first module in the pipeline, the mean of the distribution is $f_0 = 0.0$. For the subsequent modules $i$ in the pipeline, the mean $f_{i-1}$ is the output of the previous module $i-1$. This function possesses all the aforementioned properties of the AutoML problem \eqref{eq: prob0}.

In black-box optimization with this objective, the black-box evaluations are very cheap in contrast to the actual AutoML problem where the black-box evaluation requires a significant computational effort (and hence time). However, we utilize this artificial objective to evaluate ADMM (and other baselines) when the computational costs are just limited to the actual derivative-free optimization.

\clearpage\pagebreak

\section{TPOT pipelines: Variable length, order and non-sequential} \label{asec:TPOT-exs}
The genetic algorithm in TPOT does stitches pipelines together to get longer length as well as non-sequential pipelines, using the same module multiple times and in different ordering. Given the abilities to 
\begin{enumerate}
\item[i.]   have variable length and variable ordering of modules, 
\item[ii.]  reuse modules, and 
\item[iii.] have non-sequential parallel pipelines,
\end{enumerate}
TPOT does have access to a much larger search space than Auto-sklearn and ADMM.  Here are some examples for our experiments:

\paragraph{Sequential, length 3 with 2 estimators}
\begin{verbatim}
Input --> PolynomialFeatures --> KNeighborsClassifier --> GaussianNB

GaussianNB(
  KNeighborsClassifier(
    PolynomialFeatures(
      input_matrix, 
      PolynomialFeatures__degree=2, 
      PolynomialFeatures__include_bias=False, 
      PolynomialFeatures__interaction_only=False
    ), 
    KNeighborsClassifier__n_neighbors=7, 
    KNeighborsClassifier__p=1, 
    KNeighborsClassifier__weights=uniform
  )
)
\end{verbatim}
\paragraph{Sequential, length 4 with 3 estimators}
\begin{verbatim}
Input 
  --> PolynomialFeatures 
    --> GaussianNB 
      --> KNeighborsClassifier 
        --> GaussianNB

GaussianNB(
  KNeighborsClassifier(
    GaussianNB(
      PolynomialFeatures(
        input_matrix, 
        PolynomialFeatures__degree=2, 
        PolynomialFeatures__include_bias=False, 
        PolynomialFeatures__interaction_only=False
      )
    ), 
    KNeighborsClassifier__n_neighbors=7, 
    KNeighborsClassifier__p=1, 
    KNeighborsClassifier__weights=uniform
  )
)
\end{verbatim}
\clearpage
\pagebreak
\paragraph{Sequential, length 5 with 4 estimators}
\begin{verbatim}
Input 
  --> RandomForestClassifier 
     --> RandomForestClassifier 
        --> GaussianNB 
           --> RobustScaler 
              --> RandomForestClassifier

RandomForestClassifier(
  RobustScaler(
    GaussianNB(
      RandomForestClassifier(
        RandomForestClassifier(
          input_matrix, 
          RandomForestClassifier__bootstrap=False, 
          RandomForestClassifier__criterion=gini, 
          RandomForestClassifier__max_features=0.68, 
          RandomForestClassifier__min_samples_leaf=16, 
          RandomForestClassifier__min_samples_split=13, 
          RandomForestClassifier__n_estimators=100
        ), 
        RandomForestClassifier__bootstrap=False, 
        RandomForestClassifier__criterion=entropy, 
        RandomForestClassifier__max_features=0.9500000000000001, 
        RandomForestClassifier__min_samples_leaf=2, 
        RandomForestClassifier__min_samples_split=18, 
        RandomForestClassifier__n_estimators=100
      )
    )
  ), 
  RandomForestClassifier__bootstrap=False, 
  RandomForestClassifier__criterion=entropy, 
  RandomForestClassifier__max_features=0.48, 
  RandomForestClassifier__min_samples_leaf=2, 
  RandomForestClassifier__min_samples_split=8, 
  RandomForestClassifier__n_estimators=100
)
\end{verbatim}
\clearpage
\pagebreak
\paragraph{Non-sequential}
\begin{verbatim}
Combine[ 
  Input, 
  Input --> GaussianNB --> PolynomialFeatures --> Normalizer 
] --> RandomForestClassifier

RandomForestClassifier(
  CombineDFs(
    input_matrix, 
    Normalizer(
      PolynomialFeatures(
        GaussianNB(
          input_matrix
        ), 
        PolynomialFeatures__degree=2, 
        PolynomialFeatures__include_bias=True, 
        PolynomialFeatures__interaction_only=False
      ), 
      Normalizer__copy=True, 
      Normalizer__norm=l2
    )
  ), 
  RandomForestClassifier__bootstrap=False, 
  RandomForestClassifier__criterion=entropy, 
  RandomForestClassifier__max_features=0.14, 
  RandomForestClassifier__min_samples_leaf=7, 
  RandomForestClassifier__min_samples_split=8, 
  RandomForestClassifier__n_estimators=100
)
\end{verbatim}

\clearpage\pagebreak

\end{document}